\documentclass{article}

\usepackage{silence}
\usepackage[T1]{fontenc}
\usepackage{textcomp}
\usepackage{xcolor}

\usepackage[main,final]{neurips_2026}

\usepackage{amsmath}
\usepackage{amssymb}
\usepackage{booktabs}
\usepackage{array}
\usepackage{graphicx}
\graphicspath{{figures/}}
\usepackage{colortbl}
\usepackage{multirow}
\usepackage{tikz}
\usetikzlibrary{shapes,arrows.meta,positioning,calc,fit,backgrounds,decorations.pathreplacing,shadings,3d}
\usepackage{fontawesome5}
\usepackage{pgfplots}
\pgfplotsset{compat=1.18}
\usepackage{subcaption}
\usepackage{placeins}
\usepackage{enumitem}
\usepackage{microtype}
\usepackage[hidelinks,linktoc=all]{hyperref}
\hypersetup{pdftitle={Weather-Robust Cross-View Geo-Localization via Prototype-Based Semantic Part Discovery},pdfauthor={Chi-Nguyen Tran, Duy-Minh Dao-Sy, Trung-Kiet Huynh, Phu-Hoa Pham, Phu-Quy Nguyen-Lam, Long Tran-Thanh}}
\usepackage{titletoc}

\definecolor{skyblue}{HTML}{0A84FF}
\definecolor{skygold}{HTML}{FFD60A}
\definecolor{skyorange}{HTML}{FF9F0A}
\definecolor{skygreen}{HTML}{30D158}
\definecolor{skyred}{HTML}{FF453A}
\definecolor{bestrow}{HTML}{E8F5E9}
\definecolor{skydark}{HTML}{1C1C1E}

\newcommand{\ours}{\textsc{SkyPart}}
\usepackage{pifont}
\newcommand{\cmark}{\textcolor{green!55!black}{\ding{51}}}
\newcommand{\xmark}{\textcolor{red!75!black}{\ding{55}}}
\newcommand{\pmv}[2]{#1\,{\scriptsize$\pm$\,#2}}
\newcommand{\gainv}[1]{{\scriptsize\textcolor{green!45!black}{(#1)}}}
\newcommand{\dup}[1]{{\scriptsize\textcolor{green!45!black}{(#1)}}}
\newcommand{\ddown}[1]{{\scriptsize\textcolor{red!70!black}{(#1)}}}
\newcommand{\gup}[1]{\textcolor{green!45!black}{#1}}

\newcommand{\reals}{\mathbb{R}}
\newcommand{\loss}{\mathcal{L}}
\newcommand{\etal}{et~al.}

\definecolor{hcgreen}{HTML}{C8E6C9}
\definecolor{hcyellow}{HTML}{FFF9C4}
\definecolor{hcred}{HTML}{FFCDD2}
\newcommand{\hc}[1]{%
  \ifdim #1 pt > 90 pt \cellcolor{hcgreen!70}%
  \else\ifdim #1 pt > 80 pt \cellcolor{hcgreen!35}%
  \else\ifdim #1 pt > 70 pt \cellcolor{hcyellow!60}%
  \else\ifdim #1 pt > 60 pt \cellcolor{hcyellow!90}%
  \else\ifdim #1 pt > 50 pt \cellcolor{skyorange!20}%
  \else\cellcolor{hcred!50}%
  \fi\fi\fi\fi\fi
  #1%
}
\newcommand{\hbest}[1]{\textcolor{skyred}{\textbf{\hc{#1}}}}
\newcommand{\hsecond}[1]{\textcolor{skyblue}{\textbf{\hc{#1}}}}
\newcommand{\sbest}[1]{\textcolor{skyred}{\textbf{#1}}}
\newcommand{\ssecond}[1]{\textcolor{skyblue}{\textbf{#1}}}

\newcommand{\dcat}[1]{\textcolor{skyred}{\textbf{#1}}~{\tiny\textcolor{skyred}{$\blacktriangleleft$}}}   %
\newcommand{\dlrg}[1]{\textcolor{skyblue}{\textbf{#1}}~{\tiny\textcolor{skyblue}{$\triangleleft$}}}      %
\newcommand{\dmin}[1]{\textcolor{gray}{#1}}                                                               %
\newcommand{\dnone}{---}                                                                                   %
\newcolumntype{\abconfig}{>{\raggedright\arraybackslash}p{2.6cm}}
\newcommand{\absecsep}{\midrule\addlinespace[0.1em]}
\newcommand{\ablationtabular}{%
  \begin{tabular}{@{} \abconfig *{11}{cc} r @{}}
  \toprule
  \multirow{2}{*}{\textbf{Configuration}} & \multicolumn{2}{c}{\textbf{Normal}} & \multicolumn{2}{c}{\textbf{Fog}} & \multicolumn{2}{c}{\textbf{Rain}} & \multicolumn{2}{c}{\textbf{Snow}} & \multicolumn{2}{c}{\textbf{F+R}} & \multicolumn{2}{c}{\textbf{F+S}} & \multicolumn{2}{c}{\textbf{R+S}} & \multicolumn{2}{c}{\textbf{Dark}} & \multicolumn{2}{c}{\textbf{Over-exp}} & \multicolumn{2}{c}{\textbf{Wind}} & \multicolumn{2}{c}{\textbf{Mean}} & \multirow{2}{*}{\textbf{$\Delta$Mean}} \\
  \cmidrule(lr){2-3} \cmidrule(lr){4-5} \cmidrule(lr){6-7} \cmidrule(lr){8-9} \cmidrule(lr){10-11} \cmidrule(lr){12-13} \cmidrule(lr){14-15} \cmidrule(lr){16-17} \cmidrule(lr){18-19} \cmidrule(lr){20-21} \cmidrule(lr){22-23}
  & \scriptsize R@1 & \scriptsize AP & \scriptsize R@1 & \scriptsize AP & \scriptsize R@1 & \scriptsize AP & \scriptsize R@1 & \scriptsize AP & \scriptsize R@1 & \scriptsize AP & \scriptsize R@1 & \scriptsize AP & \scriptsize R@1 & \scriptsize AP & \scriptsize R@1 & \scriptsize AP & \scriptsize R@1 & \scriptsize AP & \scriptsize R@1 & \scriptsize AP & \scriptsize R@1 & \scriptsize AP & \\
  \midrule}
\newcommand{\ablationcaptionsuffix}{%
  Training follows the matched weather-online recipe. Deltas in the text are computed against the corresponding full-model row. Weather-run implementation details are in Appendix~\ref{sec:appendix_weather}.}

\title{Weather-Robust Cross-View Geo-Localization via Prototype-Based Semantic Part Discovery}

\author{
  Chi-Nguyen Tran$^{1,2,4,5*\dagger}$\quad Duy-Minh Dao-Sy$^{1,2,4,5*\dagger}$\quad Trung-Kiet Huynh$^{1,2,4,5*\dagger}$ \\
  \textbf{Phu-Hoa Pham}$^{1,2,4,5\dagger}$\quad \textbf{Phu-Quy Nguyen-Lam}$^{1,2,4,5\dagger}$\quad \textbf{Long Tran-Thanh}$^{3}$ \\[0.4em]
  $^{1}$Faculty of Information Technology, University of Science, Ho Chi Minh City, Vietnam \\
  $^{2}$Vietnam National University, Ho Chi Minh City, Vietnam \\
  $^{3}$Department of Computer Science, University of Warwick, Coventry, United Kingdom \\
  $^{4}$CompactAI \\
  $^{5}$Realtime Robotics, Ho Chi Minh City, Vietnam \\[0.2em]
  {\small $^{*}$Equal contribution\quad $^{\dagger}$Work done during an internship at Realtime Robotics.}
}
\date{}

\begin{document}

\maketitle

\begin{abstract}
Low-altitude economy has been experiencing rapid growth in recent years, with significant contributions to the global economy. While common drone tasks such as delivery, inspection, and search-and-rescue typically use Global Navigation Satellite Systems (GNSS) to navigate, there is an increasing need for developing alternative solutions as GNSS signals can be easily jammed, spoofed, or unavailable over a prolonged operational time.
As such, cross-view geo-localization (CVGL), which matches an oblique drone view to a geo-referenced satellite tile, has emerged as a potent alternative that lets an autonomous drone localize itself when GNSS fails. 
Despite strong recent progress, three limitations persist in current CVGL methods: 1) global-descriptor designs compress the patch grid into a single vector without separating what is shared across the view gap (layout) from what is not (texture); 2) altitude-related scale variation is implicitly retained in the learned embedding rather than treated as a nuisance to be marginalized out; and 3) multi-objective training relies on hand-tuned scalars over losses that live on incompatible gradient scales. To address these limitations, we propose \ours{}, a lightweight swappable head for patch-based vision transformers (ViTs) that institutes explicit part grouping over the patch grid. \ours{} has four components grounded in established theory: (i) learnable prototypes that compete for patch tokens via a single-pass cosine assignment; (ii) altitude-conditioned linear modulation applied only during training so that the retrieval embedding is altitude-free at inference; (iii) a graph-attention readout over active prototypes, and (iv) a Kendall uncertainty-weighted multi-objective loss that learns the loss-group weights end-to-end rather than tuning them by hand. At 26.95\,M parameters and 22.14\,GFLOPs, \ours{} is the smallest among the top-performing methods in our comparison and sets a new state of the art on SUES-200, University-1652, and DenseUAV datasets under a single-pass, no-re-ranking, no test-time augmentation (TTA) protocol. Furthermore, its accuracy gap to the strongest baseline widens under the ten-condition WeatherPrompt corruption benchmark. %
Code is available at \url{https://github.com/chisngyen/SkyPart}.
\end{abstract}

\section{Introduction}
\label{sec:intro}

Autonomous drones and autonomous driving are two of the fastest maturing AI application areas in the last five years. Both are moving from research demonstration to large-scale deployment, such as commercial drone delivery and inspection fleets and self-driving vehicle pilots in dozens of cities. Both domains share the same hard requirement: the agent must know where it is on a prior map, in real time, without relying on a radio signal that an adversary or a dense urban canyon can take away. This requires efficient robust visual geo-localization against a previously captured overhead map, which has drawn increasing attention from the research community. %
In particular, an autonomous drone that cannot localize itself cannot complete a mission. Global Navigation Satellite System (GNSS) jamming and spoofing are now routine in contested airspace and dense urban canyons~\citep{kuusniemi2024gnss,couturier2021visual}, and the next generation of low-altitude services, such as last-mile delivery, precision agriculture, search-and-rescue swarms, and infrastructure inspection, is being designed on the premise that radio signals may be absent or adversarial. Cross-view geo-localization (CVGL), which matches an oblique drone image to a geo-referenced satellite tile, has become one of the most important techniques to address this challenge, as it has the potential to remain robust under altitude change, weather corruption, and tight onboard compute budgets, requirements that are typical of many modern-day visual navigation tasks.

The core difficulty of CVGL is that the visual statistics that modern recognition techniques rely on (e.g., texture, colour, local contrast) are precisely the statistics that change most frequently between views and under weather. This leads to three limitations that persist in the current CVGL literature: 1) global-descriptor designs~\citep{yang2024camp,deuser2023sample4geo,du2024ccr} compress the patch grid into a single vector with no mechanism to separate what is shared across the view gap from what is not; 2) altitude-related scale variation is implicitly retained in the learned embedding rather than marginalized as a nuisance; and 3) multi-objective training relies on hand-tuned scalars over losses on incompatible gradient scales. 

We argue that to address these issues, one needs to factor out texture and keep the spatial organization of recurring components such as roads, roofs, and vegetation patches. We hypothesize that layout cues are more stable than texture under viewpoint, lighting, and sensor changes, and build \ours{} around that assumption~\citep{marr1978representation,biederman1987rbc}. %
Following the invariance vs. sufficiency trade-off of the information bottleneck~\citep{achille2018invariance}, we adopt the working assumption that a good retrieval embedding should approximate the minimal sufficient statistic of the geolocation signal after nuisance variables have been marginalized out.
Given this, our idea is that CVGL should be formulated around explicit grouping rather than ever-larger global pooling, where self-supervised ViTs can be used to expose emergent object-centric structure at the patch level~\citep{caron2021dino,oquab2024dinov2}, and object-centric learning to identify a small set of competing slots to carve that structure into discrete, reusable components~\citep{locatello2020slot}. The layout of those components is what is shared across the drone/satellite gap.

This idea reformulates the aerial CVGL problem as a layout-centric factorization problem, from which four architectural choices follow directly: prototype-based part discovery, nuisance-aware training modulation, multi-objective training, and active prototypes identification.
To implement this pipeline, we develop \ours{}, which consists of:
\begin{enumerate}
    \item Single-pass cosine assignment module for prototype-based part discovery~\citep{locatello2020slot,luo2023segclip,sun2018beyond}; 
    \item Altitude-conditioned Feature-wise Linear Modulation (FiLM)~\citep{perez2018film,dumoulin2018featurewise}, which is applied only during training, so altitude is a nuisance to marginalize rather than an inference feature; 
    \item Kendall-weighted multi-objective training~\citep{kendall2018multi} over four loss groups (alignment, part quality, altitude, distillation), with the group weights learned end-to-end rather than hand-tuned; 
    \item A graph-attention readout~\citep{velickovic2018gat,kipf2017gcn} over active prototypes, which carries the pairwise compatibility of discovered components alongside the global image representation derived from the classification token (CLS) of the ViT. 
\end{enumerate}

We also apply a lightweight swappable head for patch-based ViTs that operationalizes the reformulation and trains end-to-end with a grouped multi-objective loss called \textsc{GeoPartLoss}. As such, the inference pipeline of \ours{} is a single forward pass with cosine similarity, no re-ranking, no test-time augmentation, and no query expansion.
Note that \ours{} ranks first on SUES-200, University-1652, and DenseUAV under our evaluation protocol (single-pass, no-re-ranking, no-TTA; see Sec.~\ref{sec:impl}), and is the most parameter-efficient among the top-accuracy-tier methods in the comparison. Under the reported WeatherPrompt protocol~\citep{wen2025weatherprompt}, \ours{} shows the strongest mean robustness, with especially large gains under fog$+$snow and darkness; SUES-200 and University-1652 baselines are taken from their published configurations and \ours{} is evaluated under a shared corruption pipeline, so resolution and training-budget differences (detailed in Sec.~\ref{sec:weather}) are reported explicitly rather than fully resolution-controlled.
Overall, \ours{} is among the most parameter- and compute-efficient methods in the top-accuracy tier (26.95\,M, 22.14\,GFLOPs), retains a strong relative robustness margin under the WeatherPrompt~\citep{wen2025weatherprompt} corruption benchmark, and we further include a lessons-learned catalogue of the methods we tried that underperformed (Appendix~\ref{sec:appendix_negative}).

We also apply \ours{} to the problem of zero-shot cross-dataset transfer (University-1652 $\to$ SUES-200). Here, \ours{} leads all Drone$\to$Satellite cells and remains competitive on Satellite$\to$Drone, where dedicated baselines retain advantages in some R@1/AP cells (see Appendix~\ref{sec:appendix_crossdata} for the full table).

\section{Related Work}
\label{sec:related}

\textbf{Cross-View Geo-Localization.} 
CVGL began as a ground$\to$aerial retrieval problem~\citep{workman2015widearea,lin2015crossview,hu2018cvmnet}, solved by NetVLAD-style aggregators~\citep{arandjelovic2016netvlad} and triplet objectives~\citep{schroff2015facenet}, and sharpened by geometry-aware priors such as polar warping~\citep{shi2020polar}. The UAV benchmarks~\citep{zheng2020university1652,zhu2023sues200,dai2023denseuav} removed most of that prior because a drone's oblique angle and continuous altitude variation replace a clean radial warp with a soft projective distortion. Recent UAV CVGL has responded on two fronts: a scale-up line stacking larger ConvNeXt-B backbones with hard-negative sampling and attention-based partitioning~\citep{dai2022fsra,shen2023mccg,deuser2023sample4geo,chen2024sdpl,du2024ccr,yang2024camp,chen2025multilevel,hou2025mcfa}, and an efficiency line that shrinks the model at a cost in accuracy~\citep{wang2022lpn,wang2024segcn,xu2025glqinet,ping2025dinomsra,gao2025scpnet,sun2024trisa}. Both lines retain a global-descriptor commitment, sometimes augmented by a fixed spatial partition; partial-overlap settings~\citep{ye2025anyvisloc} expose the low-altitude limit of both. \ours{} departs from this commitment and replaces the global pool with an explicit prototype grouping that the retrieval objective can reward directly. Closest in spirit among graph-based efficiency methods, SeGCN~\citep{wang2024segcn} relies on semantic categories from an auxiliary module to define graph nodes, while \ours{} discovers its prototype nodes end-to-end from patch-token competition without auxiliary semantic supervision. Classical vision-aided navigation can instead align video frames to views rendered from a geo-registered 3D model using geometric correspondences, inertial cues, and temporal optimization~\citep{chiu2014precise}. \ours{} ranks flat satellite tiles from a single image and may serve as a coarse front end for such a system.

\noindent
\textbf{Part-based Representations and Object-Centric Grouping.}
Part-based recognition has a long cognitive-vision history~\citep{marr1978representation,biederman1987rbc,sun2018beyond}, and its modern incarnation is slot attention~\citep{locatello2020slot}: a small set of latents competes for input tokens through iterative attention, producing object-centric groupings without supervision. SegCLIP~\citep{luo2023segclip} projects this onto ViT patch tokens~\citep{dosovitskiy2021vit,vaswani2017attention} for open-vocabulary segmentation, and masked autoencoders and feature-masking variants~\citep{he2022mae,zhou2022ibot,baevski2022data2vec} pursue analogous grouping through reconstruction. None was designed for cross-view retrieval, and the altitude nuisance specific to aerial imagery has no analogue in their benchmarks. \ours{} uses a single-pass cosine assignment (a fixed-point view of slot iteration that is sufficient when the downstream objective is retrieval rather than reconstruction), applies masking and distillation at the \emph{part} level rather than the patch level so that reconstruction pressure targets layout rather than texture, and conditions the clustering pathway on altitude via FiLM~\citep{perez2018film,dumoulin2018featurewise}.

\noindent
\textbf{Metric Learning and Multi-Task Optimization.}
Our retrieval objective groups Circle~\citep{sun2020circle}, proxy-anchor~\citep{kim2020proxy}, InfoNCE~\citep{oord2018infonce}, and patch-level NCE into a single \emph{alignment} task, and combines that alignment task with three further tasks (part quality, altitude regression, and distillation) via Kendall homoscedastic uncertainty weighting~\citep{kendall2018multi}, read through the alignment/uniformity decomposition of contrastive learning~\citep{wang2020alignment,chen2020simclr}: the decomposition predicts that alignment carries the signal and uniformity regularizes it, which the ablation in Sec.~\ref{sec:ablation} confirms. The general theorem of Sener \& Koltun~\citep{sener2018mtl} guarantees that any positive-weighted scalarization of the four task losses is Pareto-stationary in the vector objective, and Kendall weighting produces such positive weights $\exp(-s_g)$ automatically; what it buys us over a hand-tuned scalar combination (which is also Pareto-stationary) is that the four weights are learned end-to-end rather than swept manually. The relational readout over prototype nodes~\citep{velickovic2018gat,kipf2017gcn} complements the global CLS pathway; denoising~\citep{dai2024mcgf} and domain-adversarial~\citep{ganin2015dann} alternatives are orthogonal. Distillation~\citep{hinton2015distilling,gou2021knowledge,tian2020crd,jang2024whiten,park2019rkd} is used only to stabilize the teacher-space projector during training, so the deployed pipeline has no external detector, diffusion front-end, or inference-time adaptation.

\section{Method: \ours{}}
\label{sec:method}

\subsection{Architecture Overview}
\label{sec:overview}
We start with the shape of the head which follows from the framing we described in Sec.~\ref{sec:intro}. If layout is the invariant we want the embedding to preserve across the drone/satellite gap, then the representation should expose an identity axis (what the recurring components are) and an arrangement axis (how they are spatially related), and it should do so without letting altitude, which is a nuisance observed at training but unreliable at test, leak into the retrieval embedding. Four choices fall out of this framing. First, the part branch is built as a competition between a small bank of prototypes for a larger set of patch tokens, with the softmax running over prototype indices rather than patch positions, so that each patch commits to a prototype rather than each prototype averaging over all patches. 
This is the single-pass analogue of slot attention~\citep{locatello2020slot}, which we treat as sufficient in our retrieval setting because the downstream objective does not require a reconstruction from the grouping. Second, a graph readout over the active prototype nodes can produce a separate arrangement embedding, complementing the identity-heavy part descriptor and the global CLS pathway, which refers to the branch that uses the CLS token as a global image descriptor~\footnote{The CLS (classification) token is a learnable, special vector added to the input sequence of a Vision Transformer (ViT) to represent the entire image for classification tasks~\citep{lappe2025register}.}. Third, altitude conditions the clustering pathway through a feature-wise affine modulation~\citep{perez2018film,dumoulin2018featurewise} but is marginalized at inference through mean FiLM parameters. 
This yields an exact equivalence only at the modulation-operator level (linearity in $\gamma,\beta$), not a claim that training-conditioned features are globally unchanged (formal statement in Appendix~\ref{sec:appendix_theory}).
Fourth, the four training heads (alignment, part quality, altitude, and distillation) live on incompatible gradient scales. To combine them we use Kendall homoscedastic uncertainty weighting~\citep{kendall2018multi}. 
Sener \& Koltun's general theorem~\citep{sener2018mtl} guarantees Pareto-stationarity for any positive-weighted scalarization of the four task losses; Kendall weighting produces such positive weights $\exp(-s_g)$ automatically, so its advantage over a hand-tuned scalar combination (which is also Pareto-stationary) is that the four weights are learned end-to-end rather than swept manually. The theoretical justification of this idea can be found in 
Appendix~\ref{sec:appendix_theory}.

Figure~\ref{fig:architecture} depicts the resulting full pipeline. A shared patch-based encoder, DINOv2 ViT-S/14 in our experiments, chosen because its self-distillation pretext produces patch tokens with emergent object-centric structure~\citep{caron2021dino,oquab2024dinov2} that the prototype assignment can cluster on, processes both views to produce a CLS token and a set of patch tokens. The three branches read from these tokens: a global branch that keeps the CLS, a part branch that runs prototype competition and returns a part-pooled descriptor, and a graph branch that runs a two-layer Graph Attention Network (GAT)~\citep{velickovic2018gat} over per-view active prototype nodes (fully connected within each view; no explicit drone-satellite edges in the graph). A lightweight fusion gate combines the three into a single 768-D retrieval vector. Altitude-FiLM acts only as a metadata-conditioned branch during training; at inference it uses fixed mean FiLM parameters and requires no altitude input. The auxiliary signals (masked part reconstruction, teacher distillation, EMA self-ensembling) are discarded at inference. The deployed model is the student backbone plus the head, with cosine similarity and no re-ranking or test-time augmentation.

\definecolor{OvInputFill}{HTML}{E0F2FE}  \definecolor{OvInputEdge}{HTML}{0369A1}
\definecolor{OvBackFill}{HTML}{F1F5F9}   \definecolor{OvBackEdge}{HTML}{475569}
\definecolor{OvGlobalFill}{HTML}{DBEAFE} \definecolor{OvGlobalEdge}{HTML}{2563EB}
\definecolor{OvPartFill}{HTML}{FEF9C3}   \definecolor{OvPartEdge}{HTML}{B45309}
\definecolor{OvGraphFill}{HTML}{EDE9FE}  \definecolor{OvGraphEdge}{HTML}{7C3AED}
\definecolor{OvFuseFill}{HTML}{E5E7EB}   \definecolor{OvFuseEdge}{HTML}{374151}
\definecolor{OvLossFill}{HTML}{FCE7F3}   \definecolor{OvLossEdge}{HTML}{DB2777}
\definecolor{OvEmbA}{HTML}{3B82F6}       \definecolor{OvEmbB}{HTML}{EC4899}
\definecolor{ProtoAmber}{RGB}{255,140,30}   \definecolor{ProtoGreen}{RGB}{40,190,90}
\definecolor{ProtoPink}{RGB}{230,100,180}   \definecolor{ProtoPurple}{RGB}{180,60,220}
\definecolor{ProtoRed}{RGB}{220,55,55}

\begin{figure*}[t]
  \centering
  \includegraphics[width=\linewidth]{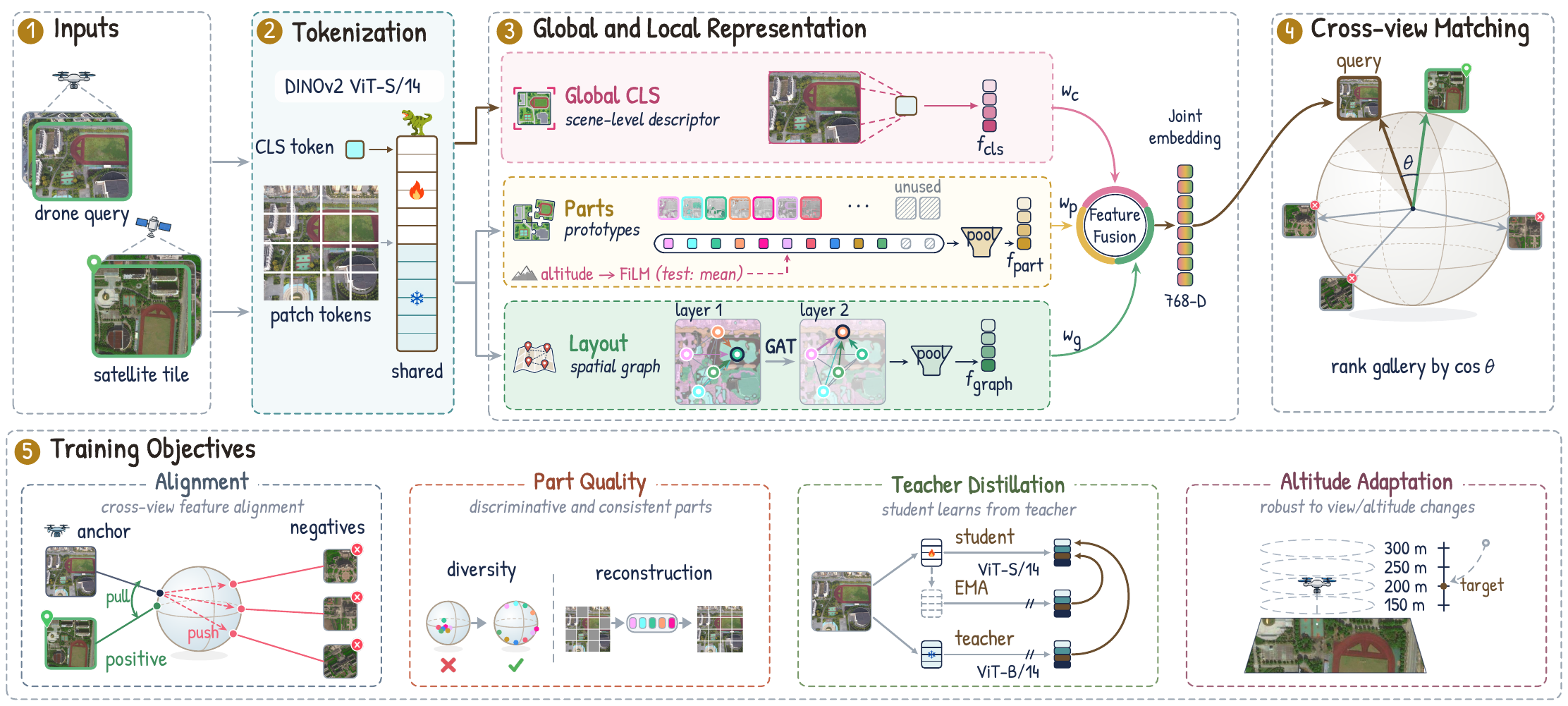}
  \caption{\textbf{\ours{} overview.} \textbf{(1--2)} A shared DINOv2 ViT-S/14 tokenizes the drone query and the satellite tile into a CLS token and patch tokens. \textbf{(3)} The head reads the tokens in three ways: the global CLS token; a part branch in which patch tokens are assigned by cosine similarity to $K_{\max}{=}12$ learnable prototypes and pooled into $\mathbf{f}_\text{part}$; and a two-layer GAT over the active prototypes that encodes their spatial layout. Altitude conditions the part branch through FiLM during training only; at test time FiLM uses its mean parameters and needs no altitude input. A fusion gate with weights $w_c, w_p, w_g$ combines the three readouts into one $\ell_2$-normalized 768-D embedding. \textbf{(4)} Retrieval is a single cosine ranking of the satellite gallery. \textbf{(5)} The four Kendall-weighted loss groups (alignment, part quality, teacher distillation, altitude) are used only during training.}
  \label{fig:architecture}
\end{figure*}

\subsection{Semantic Part Discovery and Three-Way Fusion}
\label{sec:parts}

\textbf{Prototype bank and assignment.} Our representation is organized around a small bank of $K_{\max}{=}12$ learnable prototypes $\{\mathbf{p}_k\}_{k=1}^{K_{\max}}$, each acting as a soft detector for a recurring aerial region (buildings, vegetation, roads, water); the per-image salience gate (next paragraph) selects an active subset of typical size $K{=}10$, so $K_{\max}$ is the bank capacity and $K$ the typical active count throughout the paper. These prototypes are latent functional parts rather than supervised semantic categories; the region names are illustrative. Each projected, altitude-modulated patch token $\mathbf{z}_i$ commits to one prototype through a single-pass temperature-scaled cosine assignment
\begin{equation}
\mathbf{A}_{i,k} = \frac{\exp(\langle \bar{\mathbf{p}}_k,\bar{\mathbf{z}}_i\rangle/\tau)}{\sum_j \exp(\langle \bar{\mathbf{p}}_j,\bar{\mathbf{z}}_i\rangle/\tau)},
\label{eq:assignment}
\end{equation}
where $\bar{\mathbf{p}}_k$ and $\bar{\mathbf{z}}_i$ denote the $\ell_2$-normalized prototype and patch token (so $\langle\bar{\mathbf{p}}_k,\bar{\mathbf{z}}_i\rangle$ is a cosine similarity); the softmax runs \emph{over prototype indices} rather than over patch positions, which is what turns the bank into a competing set of slots~\citep{locatello2020slot} rather than a linear reweighting. Prototype count, dimensionality, and temperature are given in Appendix~\ref{sec:appendix_hyperparam}.

\textbf{Per-image salience gating.} Because different scenes activate different subsets of the bank (an urban tile rarely needs a ``water'' prototype), we learn a per-image Gumbel-sigmoid salience gate over the prototype indices that zeroes out inactive prototypes at the image level while keeping a small active set. The gate, the residual MLP that refines each active part, and the spatial-centroid bookkeeping are detailed in Appendix~\ref{sec:appendix_mar}.

\textbf{Three-way fusion.} The retrieval embedding combines three readouts. A global branch keeps the CLS token ($\mathbf{f}_\text{cls}$), a part branch pools the salience-weighted part descriptors into $\mathbf{f}_\text{part}$, and a graph branch passes the active prototype nodes through a two-layer GAT~\citep{velickovic2018gat} over a fully-connected adjacency (feasible for this small prototype bank) to yield a layout readout $\mathbf{f}_\text{graph}$. The part and CLS pathways carry identity-heavy information (what the components are), and the graph readout carries arrangement-heavy information (how they are related), consistent with the layout-invariance framing of Sec.~\ref{sec:intro} and formalized as an information-bottleneck factorization in Appendix~\ref{sec:appendix_theory}. A lightweight gating MLP learns non-negative weights summing to one, with
\begin{equation}
[w_p, w_c, w_g] = \mathrm{softmax}\bigl(\mathbf{W}_2\,\mathrm{ReLU}(\mathbf{W}_1 [\mathbf{f}_\text{part};\mathbf{f}_\text{cls};\mathbf{f}_\text{graph}])\bigr),
\label{eq:gate}
\end{equation}
initialized to start biased toward the part branch and re-balance during training. The output is the L2-normalized sum $\mathbf{f} = \text{L2Norm}(w_p\mathbf{f}_\text{part} + w_c\mathbf{f}_\text{cls} + w_g\mathbf{f}_\text{graph})$.

\begin{figure*}[t]
  \centering
  \includegraphics[width=\linewidth,height=0.34\textheight,keepaspectratio]{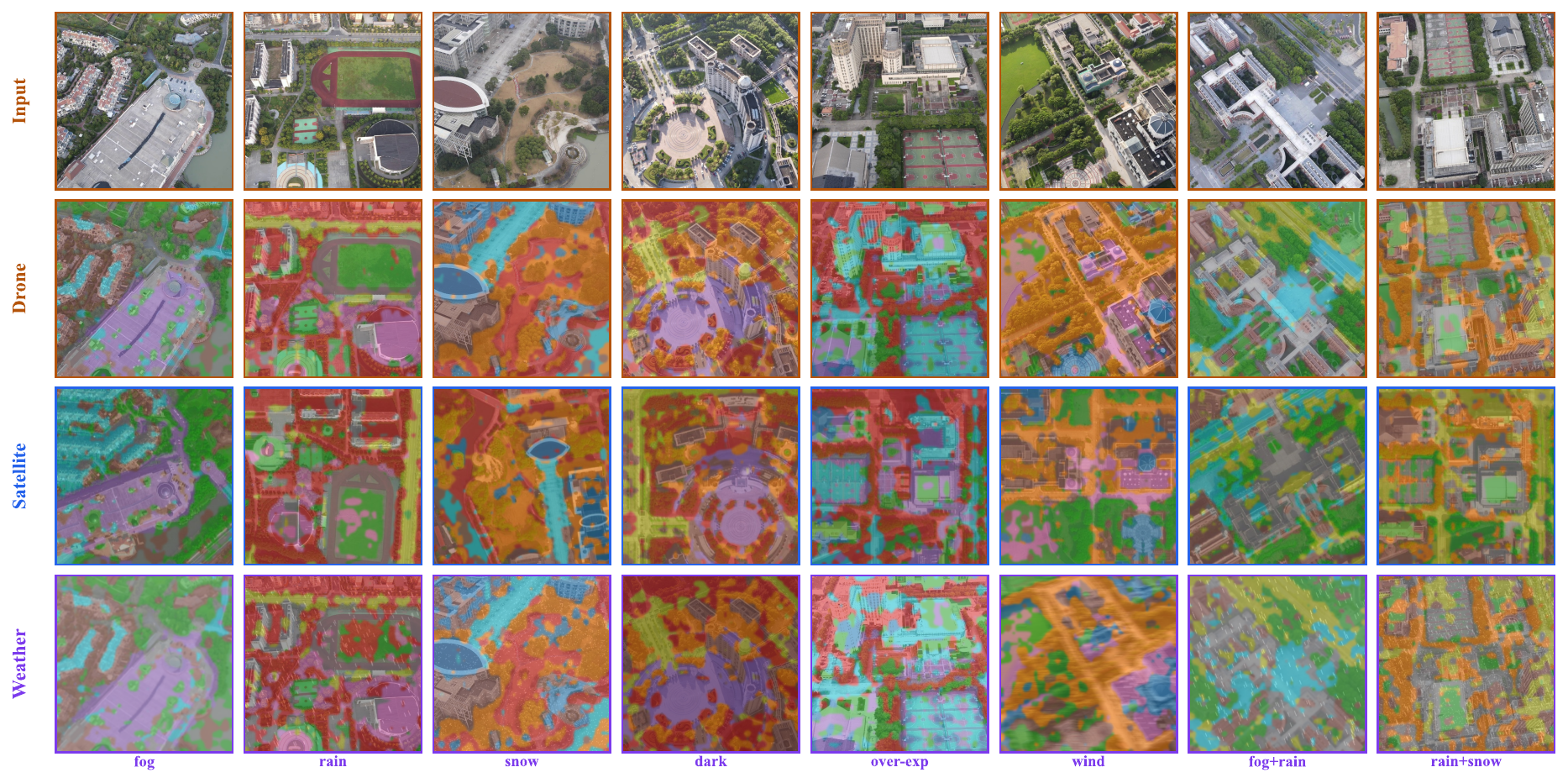}
  \caption{\textbf{Part-level evidence under weather shifts.}
    Rows show clean drone inputs, their part-level activations, paired satellite
    views, satellite part activations, weather-corrupted drone queries, and the
    corresponding part activations. Columns cover different corruptions and mixed
    weather conditions. Across substantial appearance changes, the part-discovery
    head continues to produce spatially structured activations over buildings,
    roads, fields, and open regions, rather than collapsing to a single dominant
    texture cue. These visualizations support prototype-based layout as a robust
    cross-view abstraction learned from the retrieval objective alone, without
    segmentation supervision.}
  \label{fig:part_attention}
\end{figure*}

\subsection{Altitude as a Training-Time Nuisance (FiLM)}
\label{sec:modulation}

Altitude changes scale and perspective but is not part of the location identity we want the retrieval embedding to encode, so we treat it as a nuisance to marginalize rather than a feature to keep. During training, FiLM conditions projected patch tokens on the available altitude bin before prototype assignment. At inference, no altitude input is used: altitude-specific FiLM parameters are replaced by their marginalized mean and this fixed modulation is applied to both views. Thus the deployed embedding has no dependence on test-time altitude metadata, while still benefiting from altitude-aware regularization during training. A paired three-seed control with the recipe, checkpoint rule, and evaluator fixed shows that active altitude-FiLM adds only $+0.14$\,pp clean D$\to$S R@1 but $+0.62$\,pp on the ten-condition mean (Appendix~\ref{sec:app_film}), so we treat FiLM as a robustness regularizer rather than a main accuracy driver. The formal mean-FiLM argument is given in Appendix~\ref{sec:appendix_theory}. Masked Part Reconstruction and distillation further shape the part pathway during training and are detailed in Appendix~\ref{sec:appendix_mar}.

\subsection{Training Objective: \textsc{GeoPartLoss}}
\label{sec:objective}

\textbf{GeoPartLoss}. We use a multi-objective training loss that jointly optimizes four complementary objectives via Kendall homoscedastic uncertainty weighting~\citep{kendall2018multi}:
$\mathcal{L}_{\textsc{GeoPartLoss}} = \sum_{g \in \mathcal{G}} \bigl[\exp(-s_g)\,\mathcal{L}_g + s_g\bigr]$,
where $\mathcal{G} = \{\text{align},\,\text{part},\,\text{distill},\,\text{alt}\}$ are the four loss groups, and $s_g = \log\sigma_g^2$ is a per-group learnable log-variance. The weight $\exp(-s_g)$ is always positive, so a larger $s_g$ down-weights that group; the additive $s_g$ term penalizes collapse to zero. By construction, any stationary point of this loss is a Pareto-stationary point of the four-objective vector $(\mathcal{L}_\text{align}, \mathcal{L}_\text{part}, \mathcal{L}_\text{distill}, \mathcal{L}_\text{alt})$~\citep{sener2018mtl}, removing the need for hand-tuned loss scalars.

\textbf{Four loss groups.}
\textit{(i) Alignment} ($\mathcal{L}_\text{align}$): Circle loss~\citep{sun2020circle}, proxy-anchor~\citep{kim2020proxy}, and InfoNCE~\citep{oord2018infonce} applied to the fused 768-D embedding, pulling same-location drone--satellite pairs together and pushing different-location pairs apart, plus a patch-level NCE term on part-assigned patches (Appendix~\ref{sec:appendix_loss}).
\textit{(ii) Part quality} ($\mathcal{L}_\text{part}$): masked part reconstruction (MAR) and prototype diversity, encouraging prototypes to cover distinct and spatially coherent scene regions.
\textit{(iii) Distillation} ($\mathcal{L}_\text{distill}$): feature-level alignment toward a frozen DINOv2 ViT-B/14 teacher and an exponential moving average (EMA) self-ensemble~\citep{tarvainen2017ema}, providing a stable training signal for the smaller student backbone.
\textit{(iv) Altitude} ($\mathcal{L}_\text{alt}$): SmoothL1 regression on normalized altitude, active only when metadata is available and omitted otherwise.
More details of this loss can be found in Appendix~\ref{sec:appendix_loss}.

\textbf{Uncertainty-Aware Proxy Alignment (UAPA).} Within the alignment group, we modulate the distillation temperature using the relative entropy gap between the drone and satellite proxy distributions. More uncertain drone queries receive softer teacher targets while confident queries retain the default temperature, with gradient magnitude preserved. Full formulation and ablation are in Appendix~\ref{sec:appendix_uapa}.

\textbf{Inference.} \ours{} reduces at test time to a single forward pass through the backbone and head. The teacher, EMA self-ensemble, altitude-supervision branch, and all \textsc{GeoPartLoss} terms are discarded; altitude-specific FiLM is replaced by fixed mean-FiLM parameters requiring no altitude metadata. The deployed model is the 26.95\,M student and its 768-D embedding. Retrieval uses cosine similarity with no re-ranking, TTA, query expansion, multi-crop inference, or post-processing.

\section{Experiments}
\label{sec:exp}

\subsection{Datasets and Protocol}
\label{sec:datasets}
We evaluate \ours{} on SUES-200~\citep{zhu2023sues200}, University-1652~\citep{zheng2020university1652}, and DenseUAV~\citep{dai2023denseuav} under a strict single-pass protocol: a single-scale $448{\times}448$ crop, the full confusion gallery when defined by the benchmark, and no test-time augmentation, re-ranking, query expansion, multi-crop inference, or other post-processing. Published baselines in the SOTA tables are reported from their original papers and may follow method-specific protocols, as detailed in Sec.~\ref{sec:sota}. Altitude is treated as an optional training-time nuisance signal, not a required input feature. We use FiLM bins when metadata is available (150/200/250/300\,m on SUES-200; dataset-provided altitude bins on DenseUAV), and mean-FiLM otherwise, omitting the altitude-supervision term on datasets without altitude labels. Thus the same head supports datasets with and without altitude annotations without pseudo-labeling or folder-derived heuristics. For weather robustness we adopt the WeatherPrompt protocol~\citep{wen2025weatherprompt} verbatim. Split sizes, per-method weather fine-tuning recipes, and dataset quirks appear in Appendix~\ref{sec:appendix_protocol} and~\ref{sec:appendix_weather}. All results use official splits and full benchmark galleries.  %
WeatherPrompt corrupts drone queries only. 

\subsection{Implementation Details}
\label{sec:impl}

The deployed model is a DINOv2 ViT-S/14 student with the \ours{} head, totaling 26.95\,M parameters and 22.14\,GFLOPs for a single $448{\times}448$ forward pass. Training uses AdamW, identity-balanced mini-batches, standard crop/colour augmentation, and the \textsc{GeoPartLoss} groups described in Sec.~\ref{sec:objective}. DenseUAV additionally uses rotation augmentation during training because its satellite crops have arbitrary basemap orientation; SUES-200 and University-1652 use canonical orientations. At inference, the teacher, EMA, FiLM-altitude auxiliaries, and all training objectives are discarded; retrieval is cosine similarity between 768-D embeddings, with no re-ranking, TTA, query expansion, multi-crop inference, or post-processing. Results for \ours{} follow this deployed protocol. Across three independently trained seeds (42, 123, 2024; final-epoch checkpoints, no test-set selection), clean D$\to$S R@1 is $98.74\pm0.11$ on SUES-200, $96.47\pm0.25$ on University-1652, and $91.85\pm0.26$ on DenseUAV (mean $\pm$ sample SD). Full hyperparameters and dataset-specific training details are in Appendix~\ref{sec:appendix_hyperparam} and Appendix~\ref{sec:appendix_protocol}.

\subsection{Comparison with State of the Art}
\label{sec:sota}

\textbf{SUES-200.}
Table~\ref{tab:sues200} reports the SUES-200 comparison. Under the single-pass protocol (a single $448{\times}448$ crop, one forward pass, the full 200-tile confusion gallery, and \emph{no} re-ranking, TTA, query expansion, or multi-crop inference, while baselines retain whatever inference-time wrappers their original papers use), \ours{} sets a new state of the art across every altitude band on D$\to$S with the smallest model in the table. The margin grows at the lowest altitude, where the field of view is narrow and only partial facades are visible: global descriptors lose traction precisely in the regime where part-level layout becomes the deciding cue, exactly as the framing in Sec.~\ref{sec:intro} predicts.

\begin{table*}[!t]
\caption{\textbf{SUES-200 per-altitude comparison.} R@1/AP (\%) at four UAV altitudes for D$\to$S and S$\to$D, with mean R@1 per direction in the rightmost column of each block. \ours{} is evaluated with a single-scale $448{\times}448$ crop, the full 200-satellite confusion gallery, and no post-processing; the \ours{} row is one checkpoint (three-seed mean D$\to$S R@1: $98.74\pm0.11$, Sec.~\ref{sec:impl}); non-\ours{} numbers are reported from the original publications. Cell shading scales with R@1/AP (green $=$ high, red $=$ low). \textcolor{skyred}{\textbf{Red}}: best; \textcolor{skyblue}{\textbf{blue}}: second; \ours{} has the lowest Params/GFLOPs.}
\label{tab:sues200}
\centering
\scriptsize
\setlength{\tabcolsep}{2.4pt}
\renewcommand{\arraystretch}{0.95}
\resizebox{\textwidth}{!}{%
\begin{tabular}{@{}l l c c cccccccc c cccccccc c@{}}
\toprule
& & & & \multicolumn{9}{c}{\textbf{Drone${\to}$Satellite}} & \multicolumn{9}{c}{\textbf{Satellite${\to}$Drone}} \\
\cmidrule(lr){5-13} \cmidrule(lr){14-22}
\textbf{Method} & \textbf{Venue} & \textbf{Params$\downarrow$} & \textbf{GFLOPs$\downarrow$} &
\multicolumn{2}{c}{150m} & \multicolumn{2}{c}{200m} & \multicolumn{2}{c}{250m} & \multicolumn{2}{c}{300m} & \textbf{Mean} &
\multicolumn{2}{c}{150m} & \multicolumn{2}{c}{200m} & \multicolumn{2}{c}{250m} & \multicolumn{2}{c}{300m} & \textbf{Mean} \\
\cmidrule(lr){5-6}\cmidrule(lr){7-8}\cmidrule(lr){9-10}\cmidrule(lr){11-12}%
\cmidrule(lr){14-15}\cmidrule(lr){16-17}\cmidrule(lr){18-19}\cmidrule(lr){20-21}
& & & & R@1 & AP & R@1 & AP & R@1 & AP & R@1 & AP & R@1 & R@1 & AP & R@1 & AP & R@1 & AP & R@1 & AP & R@1 \\
\midrule
LPN~\citep{wang2022lpn}               & TCSVT'22  & 62.4\,M & 36.78 & \hc{61.58} & \hc{67.23} & \hc{70.85} & \hc{75.96} & \hc{80.38} & \hc{83.80} & \hc{81.47} & \hc{84.53} & \hc{73.57} & \hc{83.75} & \hc{66.78} & \hc{88.75} & \hc{75.01} & \hc{92.50} & \hc{81.34} & \hc{92.50} & \hc{85.72} & \hc{89.38} \\
Baseline~\citep{zhu2023sues200}          & TCSVT'23  & --      & --    & \hc{55.65} & \hc{61.92} & \hc{66.78} & \hc{71.55} & \hc{72.00} & \hc{76.43} & \hc{74.05} & \hc{78.26} & \hc{67.12} & \hc{75.00} & \hc{55.46} & \hc{85.00} & \hc{66.05} & \hc{86.25} & \hc{69.94} & \hc{88.75} & \hc{74.46} & \hc{83.75} \\
Sample4Geo~\citep{deuser2023sample4geo}& ICCV'23   & 87.6\,M & 90.24 & \hc{92.60} & \hc{94.00} & \hc{97.38} & \hc{97.81} & \hc{98.28} & \hc{98.64} & \hc{99.18} & \hc{99.36} & \hc{96.86} & \hc{92.50} & \hc{94.00} & \hc{97.38} & \hsecond{97.81} & \hc{98.28} & \hsecond{98.64} & \hsecond{99.18} & \hsecond{99.36} & \hc{96.84} \\
MCCG~\citep{shen2023mccg}              & TCSVT'24 & 56.6\,M & 51.04 & \hc{82.22} & \hc{85.47} & \hc{89.38} & \hc{91.41} & \hc{93.82} & \hc{95.04} & \hc{95.07} & \hc{96.20} & \hc{90.12} & \hc{93.75} & \hc{89.72} & \hc{93.75} & \hc{92.21} & \hc{96.25} & \hc{96.14} & \hc{98.75} & \hc{96.64} & \hc{95.62} \\
SeGCN~\citep{wang2024segcn}             & JSTARS'24 & \ssecond{28.0\,M} & \ssecond{24.58} & \hc{90.80} & \hc{92.32} & \hc{91.93} & \hc{93.41} & \hc{92.53} & \hc{93.90} & \hc{93.33} & \hc{94.61} & \hc{92.15} & \hc{93.75} & \hc{92.45} & \hc{95.00} & \hc{93.65} & \hc{96.25} & \hc{94.39} & \hc{97.50} & \hc{94.55} & \hc{95.62} \\
CCR~\citep{du2024ccr}                   & TCSVT'24  & 156.6\,M& 160.6 & \hc{87.08} & \hc{89.55} & \hc{93.57} & \hc{94.90} & \hc{95.42} & \hc{96.28} & \hc{96.82} & \hc{97.39} & \hc{93.22} & \hc{92.50} & \hc{88.54} & \hc{97.50} & \hc{95.22} & \hc{97.50} & \hc{97.10} & \hc{97.50} & \hc{97.49} & \hc{96.25} \\
CAMP~\citep{yang2024camp}
& TGRS'24
& 91.4\,M
& 90.24
& \hc{95.40} & \hc{96.38}
& \hc{97.63} & \hc{98.16}
& \hc{98.05} & \hc{98.45}
& \hc{99.33} & \hc{99.46}
& \hc{97.60}
& \hc{96.25} & \hc{93.69}
& \hc{97.50} & \hc{96.76}
& \hsecond{98.75} & \hc{98.10}
& \hbest{100.00} & \hc{98.85}
& \hc{98.12} \\

DAC~\citep{xia2024dac}
& TCSVT'24
& 96.5\,M
& 90.24
& \hsecond{96.80} & \hsecond{97.54}
& \hc{97.48} & \hc{97.97}
& \hc{98.20} & \hc{98.62}
& \hc{97.58} & \hc{98.14}
& \hc{97.52}
& \hsecond{97.50} & \hc{94.06}
& \hsecond{98.75} & \hc{96.66}
& \hsecond{98.75} & \hc{98.09}
& \hc{98.75} & \hc{97.87}
& \hc{98.44} \\

MEAN~\citep{chen2025multilevel}
& TGRS'25
& 36.5\,M
& 26.18
& \hc{95.50} & \hc{96.46}
& \hsecond{98.38} & \hsecond{98.72}
& \hsecond{98.95} & \hsecond{99.17}
& \hsecond{99.52} & \hsecond{99.63}
& \hsecond{98.09}
& \hsecond{97.50} & \hsecond{94.75}
& \hbest{100.00} & \hc{97.09}
& \hbest{100.00} & \hc{98.28}
& \hbest{100.00} & \hc{99.21}
& \hsecond{99.38} \\
\midrule
\textbf{\ours{} (Ours)} & \textbf{--} & \sbest{26.95\,M} & \sbest{22.14} &
\hbest{97.25} & \hbest{97.90} &
\hbest{98.75} & \hbest{99.33} &
\hbest{99.30} & \hbest{99.64} &
\hbest{99.60} & \hbest{99.77} &
\hbest{98.725} &
\hbest{100.00} & \hbest{97.36} &
\hbest{100.00} & \hbest{97.95} &
\hbest{100.00} & \hbest{99.98} &
\hbest{100.00} & \hbest{99.95} &
\hbest{100.00} \\
\bottomrule
\end{tabular}%
}
\end{table*}

\textbf{University-1652 and DenseUAV.}
Table~\ref{tab:u1652_denseuav} extends the same picture to the other two benchmarks. On University-1652, \ours{} leads R@1 and AP at the lowest parameter budget in the comparison while running the inference path with mean-FiLM (no altitude metadata), evidence that the architecture itself, rather than altitude conditioning, carries the advantage. DenseUAV is the harder benchmark: its dense urban gallery makes neighbouring locations visually ambiguous and three flight heights stress the altitude module-yet \ours{} still has the highest reported R@1 and the highest reported SDM@1; some high-R@1 baselines do not report SDM@1.

\textbf{Protocol note for Tables~\ref{tab:sues200} and~\ref{tab:u1652_denseuav}.}
\ours{} is evaluated under the strict single-pass protocol of Sec.~\ref{sec:datasets}; the baseline numbers are taken directly from the original publications, where some methods use inference-time wrappers such as $k$-reciprocal re-ranking, multi-scale TTA, or query expansion. We do not apply these wrappers to the headline \ours{} results; when combined on \ours{}, they increase D$\to$S R@1 by $+0.66$, $+1.25$, and $+1.83$\,pp on SUES-200, University-1652, and DenseUAV, respectively (three seeds; Appendix~\ref{sec:app_ladder}). Conversely, \ours{} uses a $448{\times}448$ input, larger than the $384{\times}384$ or $256{\times}256$ inputs of several baselines. Tables~\ref{tab:sues200} and~\ref{tab:u1652_denseuav} should therefore be read as comparisons of published operating points rather than fully protocol-normalized results.

\begin{table*}[!t]
\caption{\textbf{University-1652 (left) and DenseUAV (right).} Side-by-side state-of-the-art comparison. \ours{} leads both benchmarks while remaining among the smallest models in the comparison. Same cell shading as Table~\ref{tab:sues200} (green $=$ high R@1/AP). \textcolor{skyred}{\textbf{Red}}: best; \textcolor{skyblue}{\textbf{blue}}: second. $^\dagger$DenseUAV: $256{\times}256$ input, as reported in the comparison of~\citep{xu2025precise}; $^\ast$param count estimated from backbone (ViT-S$\approx$22\,M, ConvNeXt-T$\approx$28\,M, ConvNeXt-B$\approx$88\,M, DINOv2-B$\approx$86\,M).}
\label{tab:u1652_denseuav}
\centering
\begin{subtable}[t]{0.56\textwidth}
  \centering
  \caption*{\textbf{(a) University-1652}}
  \vspace{2pt}
  \resizebox{\linewidth}{!}{%
  \footnotesize
  \setlength{\tabcolsep}{3pt}
  \renewcommand{\arraystretch}{0.95}
  \begin{tabular}{@{}llrcccc@{}}
  \toprule
  & & & \multicolumn{2}{c}{\textbf{D${\to}$S}} & \multicolumn{2}{c}{\textbf{S${\to}$D}} \\
  \cmidrule(lr){4-5} \cmidrule(lr){6-7}
  \textbf{Method} & \textbf{Venue} & \textbf{Params} & \textbf{R@1} & \textbf{AP} & \textbf{R@1} & \textbf{AP} \\
  \midrule
  LPN~\citep{wang2022lpn}                   & TCSVT'22 & 62.4\,M  & \hc{77.71} & \hc{80.80} & \hc{90.30} & \hc{78.78} \\
  Sample4Geo~\citep{deuser2023sample4geo}   & ICCV'23  & 87.6\,M  & \hc{92.65} & \hc{93.81} & \hc{95.14} & \hc{91.39} \\
  MCCG~\citep{shen2023mccg}                 & TCSVT'24 & 56.6\,M  & \hc{89.28} & \hc{91.01} & \hc{94.29} & \hc{89.29} \\
  CCR~\citep{du2024ccr}                     & TCSVT'24 & 156.6\,M & \hc{92.54} & \hc{93.78} & \hc{95.15} & \hc{91.80} \\
  CAMP~\citep{yang2024camp}                 & TGRS'24  & 91.4\,M  & \hsecond{94.46} & \hsecond{95.38} & \hsecond{96.15} & \hsecond{92.72} \\
  MEAN~\citep{chen2025multilevel}           & TGRS'25  & \ssecond{36.5\,M}  & \hc{93.55} & \hc{94.53} & \hc{96.01} & \hc{92.08} \\
  \midrule
  \textbf{\ours{} (Ours)} & \textbf{--} & \sbest{26.95\,M} & \hbest{96.47} & \hbest{97.77} & \hbest{98.43} & \hbest{98.24} \\
  \bottomrule
  \end{tabular}%
  }
  \label{tab:u1652}
\end{subtable}
\hfill
\begin{subtable}[t]{0.43\textwidth}
  \centering
  \caption*{\textbf{(b) DenseUAV}}
  \vspace{2pt}
  \resizebox{\linewidth}{!}{%
  \footnotesize
  \setlength{\tabcolsep}{3pt}
  \renewcommand{\arraystretch}{0.95}
  \begin{tabular}{@{}llrccc@{}}
  \toprule
  \textbf{Method} & \textbf{Venue} & \textbf{Params} & \textbf{R@1} & \textbf{R@5} & \textbf{SDM@1} \\
  \midrule
  DenseUAV baseline$^\dagger$~\citep{dai2023denseuav} & TIP'24 & $\sim$22\,M$^\ast$ & \hc{83.01} & \hc{95.58} & \hc{86.50} \\
  Sample4Geo$^\dagger$~\citep{deuser2023sample4geo} & ICCV'23 & 87.6\,M & \hc{49.38} & \hc{78.29} & \hc{61.72} \\
  MCCG$^\dagger$~\citep{shen2023mccg}           & TCSVT'24 & $\sim$28\,M$^\ast$ & \hc{83.14} & \hc{93.39} & \hc{85.94} \\
  CAMP~\citep{yang2024camp}           & TGRS'24 & 91.4\,M & \hc{88.72} & -- & -- \\
  Yang \etal$^\dagger$~\citep{yang2025dinov2uav}& RA-L'25 & $\sim$86\,M$^\ast$ & \hc{86.27} & \hsecond{96.83} & \hc{88.87} \\
  CEUSP$^\dagger$~\citep{xu2025precise}         & PRCV'25 & $\sim$28\,M$^\ast$ & \hc{89.45} & \hc{96.05} & \hsecond{91.01} \\
  MEAN~\citep{chen2025multilevel}     & TGRS'25 & 36.5\,M & \hsecond{90.18} & -- & -- \\
  \midrule
  \textbf{\ours{} (Ours)} & \textbf{--} & 26.95\,M & \hbest{91.85} & \hbest{97.81} & \hbest{93.59} \\
  \bottomrule
  \end{tabular}%
  }
  \label{tab:denseuav}
\end{subtable}
\end{table*}

\subsection{Weather Robustness (WeatherPrompt)}
\label{sec:weather}

Clean-data accuracy is a weak proxy for a drone flying through fog at dusk. WeatherPrompt~\citep{wen2025weatherprompt} stress-tests this setting by applying nine weather and visibility corruptions to the drone query (ten conditions with the clean Normal case) while keeping the satellite gallery clean. Since the deployment-critical direction is drone-query localization, Table~\ref{tab:weather} reports D$\to$S R@1 on SUES-200, University-1652, and DenseUAV; full per-condition R@1/AP results for both D$\to$S and S$\to$D are deferred to Appendix~\ref{sec:appendix_weather}.

On SUES-200 and University-1652, baselines are reported from WeatherPrompt at $384{\times}384$, while \ours{} is retrained at $392{\times}392$ (yielding a $28{\times}28$ patch grid for DINOv2 ViT-S/14) under the same corruption pipeline, so these two blocks should be read as reported-baseline rather than fully resolution-controlled comparisons. Two controls bound the resolution difference (Appendix~\ref{sec:app_resolution}): re-running WeatherPrompt at $392{\times}392$ with its own recipe changes the ten-condition mean by only $+0.18$, $+0.10$, and $+0.12$\,pp on SUES-200, University-1652, and DenseUAV; retraining \ours{} at $378{\times}378$, below WeatherPrompt's 384-px input, lowers its own mean by $0.51$, $0.61$, and $1.63$\,pp, respectively. On DenseUAV, we retrain every method under the same 60-epoch budget, batch size, online corruption seeds, and official gallery with the WeatherPrompt pipeline, while retaining method-specific input resolutions and learning rates; \ours{} also ranks first on this denser urban robustness benchmark. Across all datasets, \ours{} preserves the highest mean robustness and degrades most under low-visibility conditions (fog$+$snow, darkness, and over-exposure), showing evidence that layout survives texture corruption better than global descriptors.

\begin{table*}[!t]
\caption{\textbf{Weather robustness summary.} Drone$\to$Satellite R@1 (\%) under the 10-condition WeatherPrompt corruption pipeline with clean satellite galleries. SUES-200 and University-1652 baselines are reported from WeatherPrompt~\citep{wen2025weatherprompt} at $384{\times}384$; \ours{} is retrained at $392{\times}392$. DenseUAV uses the same corruption pipeline with reproduced baselines, yielding a third weather-robustness benchmark. Full results for D$\to$S and S$\to$D are in Appendix~\ref{sec:appendix_weather}. $^*$: official pretrained weights; $^\dagger$: official weights pretrained on University-1652 (markers as in WeatherPrompt). \textcolor{skyred}{\textbf{Red}}: best; \textcolor{skyblue}{\textbf{blue}}: second.}
\label{tab:weather}
\centering
\small
\setlength{\tabcolsep}{2.4pt}
\renewcommand{\arraystretch}{1.04}
\resizebox{\textwidth}{!}{%
\begin{tabular}{@{}llc ccccccccccc@{}}
\toprule
\textbf{Method} & \textbf{Venue} & \textbf{Params$\downarrow$}
& \textbf{Normal} & \textbf{Fog} & \textbf{Rain} & \textbf{Snow}
& \textbf{F+R} & \textbf{F+S} & \textbf{R+S}
& \textbf{Dark} & \textbf{Over-exp} & \textbf{Wind} & \textbf{Mean} \\
\midrule

\multicolumn{14}{c}{\cellcolor{gray!10}\textbf{SUES-200}} \\
Sample4Geo$^\dagger$~\citep{deuser2023sample4geo} & ICCV'23 & 87.6\,M
& \hc{74.93} & \hc{72.58} & \hc{34.60} & \hc{28.95}
& \hc{35.10} & \hc{12.95} & \hc{20.05}
& \hc{34.18} & \hc{38.40} & \hsecond{67.80} & \hc{41.95} \\
Safe-Net$^*$~\citep{lin2025safe} & TIP'25 & 25.6\,M
& \hc{76.31} & \hsecond{73.53} & \hc{54.15} & \hc{48.94}
& \hc{45.12} & \hc{40.05} & \hc{25.95}
& \hc{29.74} & \hc{54.86} & \hc{58.10} & \hc{50.68} \\
CCR$^\dagger$~\citep{du2024ccr} & TCSVT'24 & 156.6\,M
& \hc{73.22} & \hc{70.95} & \hc{60.14} & \hc{50.31}
& \hc{45.87} & \hc{45.80} & \hc{31.25}
& \hc{31.03} & \hc{59.97} & \hc{52.02} & \hc{52.06} \\
WeatherPrompt~\citep{wen2025weatherprompt} & NeurIPS'25 & 87.6\,M
& \hsecond{76.72} & \hc{68.49} & \hsecond{71.77} & \hsecond{59.95}
& \hsecond{58.24} & \hsecond{64.36} & \hsecond{58.49}
& \hsecond{40.42} & \hsecond{61.57} & \hc{65.19} & \hsecond{62.52} \\
\textbf{\ours{} (Ours)} & \textbf{--} & 26.95\,M
& \hbest{97.21} & \hbest{96.27} & \hbest{96.06} & \hbest{95.33}
& \hbest{94.71} & \hbest{90.44} & \hbest{95.17}
& \hbest{92.06} & \hbest{93.34} & \hbest{96.54} & \hbest{94.71} \\

\midrule
\multicolumn{14}{c}{\cellcolor{gray!10}\textbf{University-1652}} \\
Sample4Geo$^*$~\citep{deuser2023sample4geo} & ICCV'23 & 87.6\,M
& \hsecond{92.70} & \hsecond{88.70} & \hc{62.44} & \hc{52.76}
& \hc{52.70} & \hc{19.79} & \hc{38.19}
& \hc{46.34} & \hc{75.77} & \hsecond{81.54} & \hc{61.10} \\
Safe-Net$^*$~\citep{lin2025safe} & TIP'25 & 25.6\,M
& \hc{86.98} & \hc{82.12} & \hc{67.13} & \hc{60.50}
& \hc{54.80} & \hc{32.12} & \hc{25.83}
& \hc{41.10} & \hc{69.87} & \hc{74.32} & \hc{59.48} \\
CCR$^*$~\citep{du2024ccr} & TCSVT'24 & 156.6\,M
& \hc{92.54} & \hc{85.57} & \hc{67.46} & \hc{55.16}
& \hc{63.11} & \hc{27.74} & \hc{23.06}
& \hc{51.10} & \hsecond{75.90} & \hc{81.31} & \hc{62.30} \\
WeatherPrompt~\citep{wen2025weatherprompt} & NeurIPS'25 & 87.6\,M
& \hc{82.78} & \hc{81.46} & \hsecond{80.34} & \hsecond{77.60}
& \hsecond{78.75} & \hsecond{73.38} & \hsecond{78.41}
& \hsecond{67.22} & \hc{74.20} & \hc{77.26} & \hsecond{77.14} \\
\textbf{\ours{} (Ours)} & \textbf{--} & 26.95\,M
& \hbest{95.15} & \hbest{93.78} & \hbest{93.44} & \hbest{92.05}
& \hbest{90.05} & \hbest{88.51} & \hbest{89.47}
& \hbest{83.26} & \hbest{88.23} & \hbest{89.68} & \hbest{90.36} \\

\midrule
\multicolumn{14}{c}{\cellcolor{gray!10}\textbf{DenseUAV}} \\
Sample4Geo~\citep{deuser2023sample4geo} & ICCV'23 & 87.6\,M
& \hsecond{72.37} & \hsecond{72.42} & \hsecond{73.66} & \hsecond{72.80}
& \hsecond{73.14} & \hsecond{70.44} & \hsecond{72.59}
& \hsecond{66.50} & \hsecond{69.76} & \hsecond{72.29} & \hsecond{71.60} \\
Safe-Net~\citep{lin2025safe} & TIP'25 & 25.6\,M
& \hc{41.14} & \hc{39.47} & \hc{40.11} & \hc{37.11}
& \hc{37.54} & \hc{31.87} & \hc{39.30}
& \hc{33.20} & \hc{37.07} & \hc{40.03} & \hc{37.68} \\
CCR~\citep{du2024ccr} & TCSVT'24 & 156.6\,M
& \hc{56.76} & \hc{55.21} & \hc{55.94} & \hc{54.70}
& \hc{54.10} & \hc{51.48} & \hc{55.30}
& \hc{49.85} & \hc{54.44} & \hc{54.87} & \hc{54.26} \\
WeatherPrompt~\citep{wen2025weatherprompt} & NeurIPS'25 & 87.6\,M
& \hc{26.25} & \hc{24.15} & \hc{26.68} & \hc{24.28}
& \hc{25.35} & \hc{20.72} & \hc{25.10}
& \hc{20.08} & \hc{22.48} & \hc{25.78} & \hc{24.09} \\
\textbf{\ours{} (Ours)} & \textbf{--} & 26.95\,M
& \hbest{91.25} & \hbest{90.13} & \hbest{91.38} & \hbest{90.35}
& \hbest{89.79} & \hbest{87.69} & \hbest{91.12}
& \hbest{84.64} & \hbest{90.18} & \hbest{91.76} & \hbest{89.83} \\

\bottomrule
\end{tabular}}
\end{table*}

\subsection{Ablation Study}
\label{sec:ablation}

\begin{table}[!htbp]
\centering
\footnotesize
\setlength{\tabcolsep}{2.4pt}
\renewcommand{\arraystretch}{1.08}
\caption{\textbf{Weather-aligned ablation on SUES-200} (Drone$\to$Satellite). \textsuperscript{\textdagger}Full (Ours) is the single reference row; all $\Delta$Mean values in blocks A--C are computed against it. Block~D is a paired run and its $\Delta$ is within-block only. Blocks A--C share one ablation checkpoint (Normal 98.69, Mean 96.03), trained separately from the deployed checkpoint of Table~\ref{tab:weather} (Normal 97.21, Mean 94.71); the 1.5\,pp normal gap reflects the modest clean-accuracy cost of weather-online training. \textcolor{skyred}{\textbf{Red}} $\blacktriangleleft$: catastrophic drop ($>$10\,pp); \textcolor{skyblue}{\textbf{blue}} $\triangleleft$: large drop (2--10\,pp); \textcolor{gray}{gray}: minor ($<$2\,pp). \ablationcaptionsuffix}
\label{tab:ablation}
\resizebox{\linewidth}{!}{%
\ablationtabular

\textbf{Full (Ours)} \textsuperscript{\textdagger}
& \textbf{98.69} & \textbf{99.23} & \textbf{97.84} & \textbf{98.75} & \textbf{97.19} & \textbf{98.43} & \textbf{96.41} & \textbf{97.94} & \textbf{95.90} & \textbf{97.69} & \textbf{92.31} & \textbf{95.27} & \textbf{95.53} & \textbf{97.43} & \textbf{93.09} & \textbf{95.37} & \textbf{95.91} & \textbf{97.60} & \textbf{97.46} & \textbf{98.54} & \textbf{96.03} & \textbf{97.62} & \dnone \\
\midrule
\multicolumn{24}{@{}l}{\cellcolor{skyred!18}\scriptsize\textbf{A.}\ \textit{GeoPartLoss group removal}\strut} \\
$-\mathcal{L}_{\mathrm{align}}$
& 54.49 & 66.02 & 34.49 & 47.21 & 21.56 & 33.75 & 16.76 & 29.27 & 17.58 & 28.20 & 7.24 & 16.36 & 13.24 & 23.98 & 29.52 & 41.73 & 45.01 & 57.62 & 36.83 & 50.04 & 27.67 & 39.42 & \dcat{$-$68.4} \\
$-\mathcal{L}_{\mathrm{part}}$
& 97.28 & 98.28 & 96.39 & 97.78 & 96.51 & 97.92 & 95.58 & 97.25 & 95.14 & 97.00 & 90.89 & 94.12 & 95.36 & 97.11 & 91.30 & 94.04 & 94.31 & 96.45 & 96.48 & 97.85 & 94.92 & 96.78 & \dmin{$-$1.1} \\
$-\mathcal{L}_{\mathrm{alt}}$
& 97.47 & 98.45 & 96.96 & 98.17 & 96.84 & 98.15 & 95.59 & 97.38 & 95.46 & 97.25 & 90.98 & 94.12 & 95.63 & 97.35 & 92.41 & 94.88 & 94.19 & 96.49 & 96.68 & 98.07 & 95.22 & 97.03 & \dmin{$-$0.8} \\
$-\mathcal{L}_{\mathrm{distill}}$
& 97.63 & 98.56 & 96.70 & 98.03 & 96.16 & 97.72 & 95.36 & 97.17 & 94.16 & 96.44 & 89.42 & 92.95 & 94.77 & 96.77 & 91.79 & 94.40 & 94.32 & 96.53 & 96.24 & 97.77 & 94.65 & 96.64 & \dmin{$-$1.4} \\
\absecsep
\multicolumn{24}{@{}l}{\cellcolor{skyorange!22}\scriptsize\textbf{B.}\ \textit{Representation branch}\strut} \\
CLS readout only
& 93.56 & 96.10 & 90.43 & 94.15 & 87.55 & 92.29 & 87.39 & 92.09 & 84.01 & 89.56 & 73.19 & 81.26 & 84.78 & 90.15 & 81.91 & 87.56 & 86.20 & 91.09 & 87.34 & 91.95 & 85.64 & 90.62 & \dcat{$-$10.4} \\
Part branch only
& 97.51 & 98.58 & 96.19 & 97.81 & 95.54 & 97.35 & 94.69 & 96.87 & 93.13 & 95.90 & 88.10 & 92.34 & 93.58 & 96.06 & 91.98 & 94.58 & 94.40 & 96.66 & 95.73 & 97.41 & 94.09 & 96.36 & \dmin{$-$1.9} \\
w/o Graph branch
& 97.56 & 98.28 & 96.88 & 97.90 & 96.06 & 97.46 & 95.33 & 97.06 & 94.35 & 96.46 & 90.51 & 93.87 & 95.24 & 96.99 & 91.61 & 94.12 & 94.98 & 96.67 & 96.10 & 97.48 & 94.86 & 96.63 & \dmin{$-$1.2} \\
\absecsep
\multicolumn{24}{@{}l}{\cellcolor{skygreen!18}\scriptsize\textbf{C.}\ \textit{Prototype salience gate}\strut} \\
All prototypes active
& 97.79 & 98.62 & 96.81 & 98.09 & 95.44 & 97.24 & 95.34 & 97.08 & 93.08 & 95.81 & 86.44 & 91.24 & 93.32 & 95.81 & 90.55 & 93.57 & 92.86 & 95.60 & 95.54 & 97.25 & 93.72 & 96.03 & \dlrg{$-$2.3} \\
\absecsep
\multicolumn{24}{@{}l}{\cellcolor{skyblue!14}\scriptsize\textbf{D.}\ \textit{UAPA (paired run; $\Delta$ within-block only)}\strut} \\
w/o UAPA
& 96.92 & 98.10 & 95.64 & 97.39 & 94.91 & 97.04 & 94.74 & 96.93 & 93.22 & 95.97 & 89.23 & 93.23 & 93.43 & 96.07 & 91.71 & 94.43 & 93.37 & 95.96 & 95.68 & 97.37 & 93.88 & 96.25 & \dmin{$-$1.1} \\
w/ UAPA \textsuperscript{(paired)}
& 98.06 & 98.88 & 96.85 & 98.15 & 96.16 & 97.68 & 95.84 & 97.50 & 94.34 & 96.62 & 89.60 & 93.37 & 94.73 & 96.74 & 92.93 & 95.22 & 94.42 & 96.73 & 96.48 & 97.95 & 94.94 & 96.88 & \dnone \\

\bottomrule
\end{tabular}}
\end{table}

Table~\ref{tab:ablation} sweeps four ablation axes (loss-group removal, representation branch, prototype salience, paired UAPA) under the same D$\to$S weather-online evaluation as Table~\ref{tab:weather}. Blocks~A--C share an ablation-baseline checkpoint (Mean R@1 96.03\%) trained with the matched weather-online recipe, which is a separate training run from the deployed weather-benchmark checkpoint of Table~\ref{tab:weather} (Mean 94.71\%). The two share architecture and recipe family but are separate runs at different input resolutions ($448$ vs.\ $392$), so within-table deltas, rather than absolute scores, are the comparable quantity; Block~D is paired separately and read by its within-block delta.

\textit{Block~A (\textnormal{\textsc{GeoPartLoss}} groups).} Cross-view alignment is the dominant training signal: removing the alignment group is catastrophic (Mean R@1 collapses by tens of points), while removing part quality, altitude, or distillation each costs only about a point of Mean R@1. The asymmetry shows that \textsc{GeoPartLoss} factorizes cleanly: alignment is the signal-carrying head, and the three auxiliary heads regularize and calibrate it, paying off under stress conditions rather than on Normal weather.

\emph{Block~B (representation branch).} The \emph{CLS readout only} variant still trains the full head end-to-end (gradients still shape the backbone) and only drops the part and graph readouts at retrieval; even with this advantage, mean R@1 falls by roughly ten points relative to Full, with the largest gap on the hardest combined corruptions where layout is the only signal that survives texture loss. Reintroducing the part branch recovers most of the gap and the graph branch supplies the rest, supporting the identity/arrangement factorization motivated in Sec.~\ref{sec:intro}. These Block~B numbers differ from the \emph{CLS-only} backbone-ablation baselines in Appendix~\ref{sec:appendix_backbone}, which strip the head from training entirely (no part/graph gradients during training) and therefore sit several points lower.

\emph{Block~C (prototype salience).} Disabling the salience gate while keeping the same prototype bank and graph readout (\emph{All prototypes active}) costs about two points of Mean R@1, isolating the gain that comes from learning \emph{which} prototypes matter per image rather than merely from prototype capacity. \emph{Block~D (paired UAPA).} On the paired seed/checkpoint run, uncertainty-aware temperature modulation lifts within-block Mean R@1 by roughly a point and Mean AP by half a point; we read this delta as the contribution of UAPA in isolation, without conflating it with the canonical benchmark score. Backbone-substrate controls (DINOv2 / iBOT / MoCo v3 ViT-S) are deferred to Appendix~\ref{sec:appendix_backbone}.

\section{Conclusion}
\label{sec:conclusion}

We built \ours{} around the principle that aerial cross-view matching should factor texture out and preserve layout, and instantiated it as a lightweight part-discovery head for patch-based ViTs. Under a strict single-pass protocol (no re-ranking, no TTA), \ours{} sets the new state of the art on SUES-200, University-1652, and DenseUAV while remaining among the most parameter-efficient methods in the comparison, and its margin grows under the ten WeatherPrompt conditions~\citep{wen2025weatherprompt}. The ablations (Sec.~\ref{sec:ablation}) confirm the predicted hierarchy: alignment carries the retrieval signal, the three-way fusion absorbs most of the weather-induced drop, and altitude-FiLM acts as a training-time regularizer with no inference dependency. Zero-shot University-1652$\to$SUES-200 transfer (Appendix~\ref{sec:appendix_crossdata}) leads on D$\to$S and remains competitive on S$\to$D, supporting explicit grouping as a complement to global pooling. Broader impacts are in Appendix~\ref{sec:appendix_limitations}.

\begin{ack}
We thank the anonymous reviewers and the area chair for their constructive feedback, which improved this paper. All experiments were run on the free GPU quota provided by Kaggle. This work received no external funding, and the authors declare no competing interests.
\end{ack}

{\sloppy
\emergencystretch=4em
\hbadness=10000
\vbadness=10000
\bibliographystyle{plainnat}
\bibliography{main}
}

\newpage
\appendix
\makeatletter
\setlength{\@fptop}{0pt}
\setlength{\@fpsep}{12pt plus 2pt}
\setlength{\@fpbot}{0pt plus 1fil}
\makeatother
\setcounter{section}{0}
\renewcommand{\thesection}{A\arabic{section}}

\startcontents[appendix]
\printcontents[appendix]{}{0}{\section*{Appendix Contents}\setcounter{tocdepth}{2}}
\vspace{0.5em}
\hrule
\vspace{1em}
\newpage 

\noindent Unless explicitly stated otherwise (Appendices~\ref{sec:app_ladder} and~\ref{sec:app_repair}), \ours{} results use the single-pass evaluation protocol of the main paper; reported baselines follow the source protocols stated in the corresponding tables.

\section{Method Details}
\label{sec:app_method}
\subsection{Symbol Reference}
\label{sec:appendix_symbols}
\begin{table}[ht]
\centering\small
\caption{Key symbols and their corresponding modules.}
\setlength{\tabcolsep}{4pt}
\begin{tabular}{@{}l p{0.62\linewidth} l@{}}
\toprule
\textbf{Symbol} & \textbf{Module} & \textbf{Section} \\
\midrule
$\mathbf{P}, \mathbf{p}_k$ & Prototype bank: $K_{\max}{=}12$ slots (architectural ceiling), $d_p{=}256$; the salience gate selects a per-image active subset of typical size $K{=}10$ & \ref{sec:parts} \\
$\mathbf{A}_{i,k}$ & Soft assignment matrix & \ref{sec:parts} \\
$\hat{\mathbf{p}}_k$ & Part-aggregated features & \ref{sec:parts} \\
$g_k$ & Per-prototype salience gate (Eq.~\ref{eq:gate_k}) & \ref{sec:appendix_mar} \\
$w_p, w_c, w_g$ & Three-way fusion weights (Eq.~\ref{eq:gate}) & \ref{sec:parts} \\
$\gamma_a, \beta_a$ & FiLM scale \& shift & \ref{sec:modulation} \\
$\loss_\text{mar}$ & Masked Part Reconstruction & \ref{sec:appendix_mar} \\
$\loss_\text{proxy}$ & Proxy Anchor loss & \ref{sec:objective} \\
$\loss_\text{uapa}$ & Uncertainty-aware proxy alignment & \ref{sec:objective} \\
$\loss_\text{div}$ & Prototype diversity regularizer & \ref{sec:objective} \\
$s_g$ & GeoPartLoss learnable log-variance & \ref{sec:objective} \\
\bottomrule
\end{tabular}
\end{table}

\subsection{Theoretical Motivation: Formal Arguments}
\label{sec:appendix_theory}

The three design choices of Sec.~\ref{sec:overview} (an identity/arrangement factorization of the retrieval embedding, a softmax-over-prototype-indices assignment, and altitude as a training-time nuisance), together with the uncertainty-weighted multi-task training of Sec.~\ref{sec:objective}, are stated there only briefly. This section expands each into a formal argument. None of the results below are claims of mathematical novelty; they are restatements of established theorems~\citep{achille2018invariance,locatello2020slot,perez2018film,kendall2018multi,sener2018mtl} applied to the CVGL setting. For convenience, we label the four arguments C1 (layout as sufficient statistic), C2 (slot-competition assignment), C3 (altitude as marginalized nuisance), and C4 (Pareto-stationary MTL).

\paragraph{C1 as an information-bottleneck factorization.}
Let $X_d,X_s$ be drone and satellite views of the same scene with geolocation identity $Y$, and let $N=(N_\text{view},N_\text{alt},N_\text{weather})$ be the jointly observed nuisances. Write the CVGL retrieval embedding as $T = f_\theta(X)$. The information bottleneck objective~\citep{achille2018invariance} asks for
\begin{equation}
\min_\theta \; I(T;X) - \beta\, I(T;Y),
\label{eq:ib}
\end{equation}
with a separate \emph{invariance} term $I(T;N)\to 0$. Factorize $T = (T_\text{id},T_\text{arr})$ where $T_\text{id}$ is the part-identity descriptor (drone's part bank encoding) and $T_\text{arr}$ is the part-arrangement descriptor (graph readout over prototype nodes). Two elementary observations motivate the split: (i) the part-arrangement statistic is approximately invariant to $N_\text{view}$ (spatial layout of roads/roofs is preserved under oblique reprojection of a single planar scene) and to $N_\text{weather}$ (weather destroys local texture but preserves spatial structure), so $I(T_\text{arr};N) \ll I(T_\text{id};N)$; (ii) $T_\text{arr}$ alone is insufficient-two disjoint locations can share the same layout, so $I(T_\text{arr};Y) < I(T;Y)$ and $T_\text{id}$ must remain in the embedding. The three-branch fusion $\mathbf{f} = w_p \mathbf{f}_\text{part} + w_c \mathbf{f}_\text{cls} + w_g \mathbf{f}_\text{graph}$ is then exactly a learned linear combination of an identity-heavy component ($\mathbf{f}_\text{part}$, $\mathbf{f}_\text{cls}$) and an arrangement-heavy component ($\mathbf{f}_\text{graph}$), with the weights $w_p,w_c,w_g$ learned end-to-end under the alignment objective. Empirically, the \emph{No-Graph} ablation (Table~\ref{tab:ablation}, Block~B) costs 1.17\,pp Mean R@1 but 1.80\,pp on Fog$+$Snow: the gap widens exactly in the regime where $T_\text{id}$'s texture component is corrupted and the factorization is forced to rely on $T_\text{arr}$.

\paragraph{C2 as a single-pass slot assignment.}
Slot attention~\citep{locatello2020slot} iterates an attention-then-GRU update for $K$ slots competing over $L$ input tokens. The competition comes from a softmax \emph{across slots} (not across tokens), producing mutually-exclusive assignments that concentrate each token on one slot. Our single-pass assignment (Eq.~\ref{eq:assignment}) is the fixed point of one such iteration when (a) the slot-to-input similarity is parametrized as cosine with temperature $\tau$, (b) the GRU update reduces to identity, and (c) the slot initialization is the learned prototype bank rather than a sampled noise. In return for dropping the iterative refinement we lose a reconstruction-consistent slot assignment, but we do not need one. The downstream loss is not reconstruction; it is cross-view retrieval, for which a single competitive assignment is a sufficient statistic. The cost we pay is architectural: prototypes must be \emph{learnable and directly trained}, rather than emerging from reconstruction pressure.

\paragraph{C3 as nuisance marginalization via FiLM mean.}
FiLM applies an affine modulation $\text{FiLM}(\mathbf{z}; \gamma_a, \beta_a) = \gamma_a \odot \mathbf{z} + \beta_a$ conditioned on an altitude bin $a \in A$. At inference, when altitude is unknown we replace $(\gamma_a, \beta_a)$ with $(\bar{\gamma}, \bar{\beta}) = \tfrac{1}{|A|} \sum_a (\gamma_a, \beta_a)$. Under a uniform prior $p(a)$ this is exactly the expectation $\mathbb{E}_{a \sim p(a)}[\text{FiLM}(\mathbf{z}; \gamma_a, \beta_a)]$, the marginalized modulation. Because FiLM is linear in $(\gamma_a, \beta_a)$, the marginal is exactly the modulation produced by the marginal parameters \emph{at the FiLM-operator output}. This equality is exact only at the affine modulation step; we do not claim the final fused embedding equals the marginal of the altitude-conditioned embedding, since downstream nonlinear layers (assignment softmax, GAT, fusion gate) break linearity. With this scope, the mean-FiLM fallback is a theoretically grounded substitution at the modulation operator rather than an ad-hoc hack. Conditioning the retrieval embedding directly on $a$ (rather than conditioning only the clustering pathway and letting the final embedding collapse the altitude dimension through fusion) would instead make the embedding altitude-dependent at inference, which fails the deployment constraint motivated in Sec.~\ref{sec:overview}.

\paragraph{C4 as Pareto-stationary MTL.}
Let $\{\loss_g\}_{g=1}^G$ be the four group losses. Kendall homoscedastic uncertainty weighting~\citep{kendall2018multi} models each task as a Gaussian likelihood with learned variance $\sigma_g^2$: the log-likelihood contributes $\tfrac{1}{2\sigma_g^2}\loss_g + \tfrac{1}{2}\log\sigma_g^2$, which after the reparameterization $s_g = \log\sigma_g^2$ becomes $\tfrac{1}{2}\exp(-s_g)\loss_g + \tfrac{1}{2}s_g$. We adopt the equivalent form that drops the factor of $\tfrac{1}{2}$ (absorbed into the learning rate),
\begin{equation}
\loss_\text{total} = \sum_g \bigl[\exp(-s_g)\,\loss_g + s_g\bigr],
\label{eq:theory_total}
\end{equation}
which is the training objective introduced in Sec.~\ref{sec:objective}. Sener \& Koltun~\citep{sener2018mtl} prove that any stationary point of a non-negative weighted sum $\sum_g \alpha_g \loss_g$ with $\alpha_g \geq 0$ and $\sum_g \alpha_g > 0$ is a Pareto-stationary point of the vector objective $(\loss_1, \dots, \loss_G)$: there is no descent direction that improves all $\loss_g$ simultaneously. Eq.~\ref{eq:theory_total} produces positive weights $\alpha_g = \exp(-s_g)$ by construction, so its stationary points inherit Pareto-stationarity. The full $\alpha_g\loss_g - \log\alpha_g$ objective prevents degenerate weights; the $+s_g$ term specifically prevents $\alpha_g \to 0$, and the learned values (alignment 0.94, distillation 1.05, part 1.12, altitude 1.18; Appendix~\ref{sec:appendix_loss}) land in a tight range, a diagnostic that no single task is dominating. We do not claim that Pareto-stationarity is equivalent to Pareto-optimality; we claim only that Eq.~\ref{eq:theory_total} removes a four-dimensional weight sweep and replaces it with a stationarity property that follows from~\citep{sener2018mtl}.

\paragraph{Why these four, and why not more.}
A reader may ask why \ours{} does not additionally commit to, e.g., an equivariant backbone or an explicit nuisance adversary. Both are in the prior-art space (\citep{cohen2016groupequivariant,ganin2015dann}); we omitted both by design. Equivariance would bake a fixed rotation/scale prior into the backbone, but CVGL altitude/rotation distributions are dataset-specific (SUES-200 canonically oriented, DenseUAV arbitrary)-a fixed prior helps one and hurts the other. A domain-adversarial branch would couple training stability to a minimax game we explicitly want to avoid, given the Pareto-stationarity argument above. C1--C4 are the minimal set that determined our design; we do not claim they are the only reasonable set.

\subsection{Part-Head Mechanics, Masked Part Reconstruction, and Multi-Level Distillation}
\label{sec:appendix_mar}

\paragraph{Salience gate and active-set selection.} The per-image salience gate of Sec.~\ref{sec:parts} selects which prototypes are active for a given image. It is a Gumbel-sigmoid relaxation over per-prototype logits $\mathbf{s}\in\reals^{K_{\max}}$ computed from the part-aggregated descriptors $\hat{\mathbf{p}}_k$ (post-refine; the same tensor used by the graph readout and MAR) via a lightweight 2-layer MLP, $s_k=\mathrm{MLP}_g(\hat{\mathbf{p}}_k)$ with hidden width $d_p/4{=}64$, GELU activation, and final-bias initialized to $b_2{=}2.0$ so the gate starts biased open ($\sigma(2.0)\!\approx\!0.88$) and learns to prune:
\begin{equation}
g_k = \sigma\!\bigl((s_k+\epsilon_k)/\tau_g\bigr),\qquad \epsilon_k\sim\mathrm{Gumbel}(0,1),
\label{eq:gate_k}
\end{equation}
with $\tau_g{=}0.5$. A floor $K_{\min}{=}4$ is enforced at training time to prevent the gate from collapsing to a single prototype early. Each active prototype aggregates its assigned patch tokens through the soft assignment $\mathbf{A}$, passes through a small residual MLP, and its coordinate-normalized centroid in $[0,1]^2$ is retained as the node position for the graph readout; this 2-D position is fed to the GAT as a node feature alongside the part descriptor, letting the layout readout depend on spatial relations between prototypes, not just their identities.

\paragraph{Per-branch projections to the fusion space.} The three branches in Eq.~\ref{eq:gate} are projected to a shared 768-D space before the fusion gate combines them. (i) The CLS branch maps the 384-D DINOv2 ViT-S/14 CLS token through $\mathbf{f}_\text{cls}=\mathrm{ReLU}(\mathrm{BN}(W_c\,\mathrm{CLS}))$ with $W_c\in\reals^{768\times 384}$. (ii) The part branch passes each of the $K$ active 256-D part descriptors $\hat{\mathbf{p}}_k$ (post-refine, salience-weighted) through PartAwarePooling: it concatenates the top-3 part descriptors into a $3\!\times\!256{=}768$ vector and applies a 2-layer MLP $\reals^{768}\!\to\!\reals^{768}$ (Linear, LayerNorm, GELU, Linear) to produce $\mathbf{f}_\text{part}$. (iii) The graph branch projects the 256-D GAT readout $\mathbf{h}_\text{GAT}$ through $\mathbf{f}_\text{graph}=\mathrm{LN}(W_g\,\mathbf{h}_\text{GAT})$ with $W_g\in\reals^{768\times 256}$. The fusion gate's MLP is $\reals^{3\cdot 768}\!\to\!\reals^{384}\!\to\!\reals^{3}$, followed by softmax. All three branch outputs are L2-normalized prior to the weighted sum.

\paragraph{Masked Part Reconstruction (MAR).} 30\% of patch tokens are masked; each masked position $i$ is reconstructed through the part-aggregated features:
\begin{equation}
\loss_\text{mar} = \frac{1}{|\mathcal{M}|} \sum_{i \in \mathcal{M}} \bigl(1 - \cos(\hat{\mathbf{z}}_i, \mathbf{z}_i)\bigr),
\qquad \hat{\mathbf{z}}_i = \text{Decoder}\bigl(\sum_k \mathbf{A}_{i,k}\,\hat{\mathbf{p}}_k\bigr),
\end{equation}
where the decoder is a 2-layer MLP (GELU, LayerNorm, hidden size 512). Prototype aggregation $\hat{\mathbf{p}}_k$ uses only visible (unmasked) tokens, preventing target leakage. Soft assignments $\mathbf{A}_{i,k}$ at masked positions are detached (\texttt{stop\_gradient}): without this, gradients flow back through assignment weights and one dominant prototype monopolizes all masked positions to trivially minimize $\loss_\text{mar}$, bypassing diversity. The 30\% mask ratio and 10-epoch warm-up balance reconstruction difficulty with optimization stability.

\paragraph{Multi-level distillation.} We distill from a frozen DINOv2-B/14 teacher and an EMA self-ensemble, and regress the normalized ground-truth altitude $\tilde{a} = (a - 150)/150$, where $a$ is the raw altitude in meters and $\hat{a}$ denotes the prediction of the altitude regression head:
\begin{align}
\loss_\text{cross} &= \text{MSE}(\phi\mathbf{f}, \mathbf{f}^\mathcal{T}) + (1 - \cos(\phi\mathbf{f}, \mathbf{f}^\mathcal{T})), \\
\loss_\text{ema} &= 1 - \cos(\mathbf{f}, \mathbf{f}^\text{ema}) \quad (\alpha{=}0.996~\text{\citep{tarvainen2017ema}}), \\
\loss_\text{alt} &= \text{SmoothL1}(\hat{a}, \tilde{a}).
\end{align}
The teacher mapping $\phi$ is a lightweight projector $\phi(\mathbf{f})=\text{LayerNorm}(\mathbf{W}\mathbf{f})$ with $\mathbf{W}\!\in\!\reals^{768\times768}$ matching student to teacher dimensionality.

\subsection{UAPA: Formulation and Forward-KL Ablation}
\label{sec:appendix_uapa}

\paragraph{UAPA formulation.}
Let $\mathbf{p}_d,\mathbf{p}_s$ be the softmax probabilities over the $|C|$ proxy classes, with one proxy per training location ($|C|{=}120$ for SUES-200, 701 for University-1652, and 2{,}256 for DenseUAV). The proxy-classification head reuses the $|C|$ proxy embeddings from metric learning: each logit is the cosine similarity between the normalized retrieval embedding and a learned class proxy, followed by a softmax over the $|C|$ classes. Entropy is computed over these proxy-class probabilities.

UAPA modulates only the KL temperature using the positive relative entropy gap between the drone and satellite proxy distributions:
\begin{equation}
T = T_0\bigl(1 + \Delta H_+ / H_{\max}\bigr), \qquad
\Delta H_+ = \max\{H(\mathbf{p}_d) - H(\mathbf{p}_s), 0\},
\label{eq:uapa}
\end{equation}
where $H(\cdot)$ is Shannon entropy, $T_0{=}4$, and $H_{\max}=\log|C|$ normalizes the entropy gap. Since $0\leq \Delta H_+ \leq H_{\max}$, the effective UAPA temperature lies in $[T_0,2T_0]=[4,8]$. When the drone view is more uncertain than the satellite view, $T$ increases and softens the satellite-teacher target; otherwise the temperature remains at the base value. The loss itself is the standard temperature-softened reverse KL evaluated at $T$, $\loss_\text{uapa} = T^2\,\mathrm{KL}\!\left(\mathrm{softmax}(\mathbf{z}_d/T)\;\big\|\;\mathrm{softmax}(\mathbf{z}_s/T)\right)$, where $\mathbf{z}_d,\mathbf{z}_s$ are the proxy-class logits and the $T^2$ scaling preserves gradient magnitude across temperatures, so only the temperature (not a whole-sample re-weighting) is modulated.

\paragraph{KL direction.}
We use reverse KL (drone$\|$satellite, mode-seeking) rather than forward KL (mode-covering), so the drone branch is encouraged to concentrate on the most confident satellite peak. A forward-KL variant plateaus at 96.52\% R@1 versus 98.725\% for reverse KL, the deployed model of Table~\ref{tab:sues200} ($-2.205$\,pp). GeoPartLoss group weights converge to similar ratios under either choice (alignment 0.97, distillation 1.01, part 1.03, altitude 1.04 for forward KL), suggesting that self-calibration is robust to the KL direction.

\subsection{GeoPartLoss Grouping and Training Details}
\label{sec:appendix_loss}

\begin{table}[ht]
\caption{\textbf{\textsc{GeoPartLoss} groups} (aligned with implementation). Converged weights are end-of-training averages from the SUES-200 canonical run (60 epochs); higher $\exp(-s_g)$ = group receives stronger gradient.}
\label{tab:app_loss}
\centering
\scriptsize
\setlength{\tabcolsep}{5pt}
\renewcommand{\arraystretch}{1.15}
\begin{tabular}{@{}llcc@{}}
\toprule
\textbf{Group} & \textbf{Constituent losses} & \textbf{Weighting} & \textbf{Converged $\exp(-s_g)$} \\
\midrule
Alignment    & $\loss_\text{circle}+\loss_\text{nce}+\loss_\text{pnce}+\loss_\text{proxy}+\loss_\text{uapa}$ & $\exp(-s_\text{align})$   & 0.94 \\
Distillation & $\loss_\text{cross}+\loss_\text{ema}$                                        & $\exp(-s_\text{distill})$ & 1.05 \\
Part quality & $\loss_\text{mar}+\loss_\text{div}$                                           & $\exp(-s_\text{part})$    & 1.12 \\
Altitude     & $\loss_\text{alt}$                                                            & $\exp(-s_\text{alt})$     & 1.18 \\
\midrule
Regularizer  & $+\textstyle\sum_g s_g$\,(Kendall uncertainty)                               & automatic                 & --- \\
\bottomrule
\end{tabular}
\end{table}

The learned group weights $\exp(-s_g)$ evolve during training rather than collapsing to a fixed split: from uniform initialization they diverge within the first 20 epochs and settle by epoch 60 at approximately alignment $0.94$, distillation $1.05$, part quality $1.12$, altitude $1.18$. Alignment dominates early, when the embedding geometry is being established, and then relaxes as metric learning saturates while the part-quality and altitude weights rise to compensate, the behavior Kendall weighting should show as harder objectives take over. The log-variance regularizer $+\sum_g s_g$ keeps all four groups from collapsing to zero so gradient flow stays healthy throughout the schedule. The diversity regularizer in the part-quality group penalizes off-diagonal cosine similarities between L2-normalized prototypes, $\loss_\text{div}=\frac{1}{K_{\max}(K_{\max}-1)}\sum_{j\neq k}\langle\bar{\mathbf{p}}_j,\bar{\mathbf{p}}_k\rangle^2$, pushing the bank toward an orthogonal (well-spread) set so different prototypes attend to genuinely different scene regions. The patch-level NCE term in the alignment group uses, as positives, patch pairs assigned to the same prototype in both views and, as negatives, all other patches in the batch; this is what allows the loss to remain well-defined when a prototype's role drifts during training (see Appendix~\ref{sec:appendix_limitations} on label equivariance).

\FloatBarrier
\section{Implementation and Protocol}
\label{sec:app_impl}
\subsection{Hyperparameter Configuration}
\label{sec:appendix_hyperparam}

Table~\ref{tab:app_hyperparams} lists the hyperparameters of the deployed clean configuration. Module-level settings are given in Appendix~\ref{sec:appendix_mar}, and experiment-specific overrides (e.g., the $392{\times}392$ weather runs) in Table~\ref{tab:weather_setup}.

\subsection{Dataset Statistics and Evaluation Protocol}
\label{sec:appendix_protocol}

Table~\ref{tab:datasets} lists the train split sizes and the per-direction query/gallery counts used by all retrieval evaluations in the main paper.

\begin{table}[ht]
\centering
\caption{Training split sizes and test query/gallery counts for both retrieval directions on the three benchmarks.}
\label{tab:datasets}
\setlength{\tabcolsep}{5pt}
\renewcommand{\arraystretch}{1.05}
\small
\begin{tabular}{lcccccc}
\toprule
\multirow{3}{*}{\textbf{Dataset}}
  & \multicolumn{2}{c}{\textbf{Training phase}}
  & \multicolumn{4}{c}{\textbf{Test phase}} \\
\cmidrule(lr){2-3}\cmidrule(lr){4-7}
  & \multirow{2}{*}{\textbf{Drone}} & \multirow{2}{*}{\textbf{Satellite}}
  & \multicolumn{2}{c}{\textbf{Drone$\to$Satellite}}
  & \multicolumn{2}{c}{\textbf{Satellite$\to$Drone}} \\
\cmidrule(lr){4-5}\cmidrule(lr){6-7}
  & & & \textbf{Query} & \textbf{Gallery} & \textbf{Query} & \textbf{Gallery} \\
\midrule
University-1652~\citep{zheng2020university1652} & 37,854 &  701   & 37,855 &    951 &   701 & 51,355 \\
SUES-200~\citep{zhu2023sues200}                 & 24,000 &  120   & 16,000 &    200 &    80 & 40,000 \\
DenseUAV~\citep{dai2023denseuav}                &  6,768 & 13,536 &  2,331 & 18,198 & 4,662 &  9,099 \\
\bottomrule
\end{tabular}
\end{table}

\begin{table}[ht]
\caption{Complete hyperparameter configuration.}
\label{tab:app_hyperparams}
\centering
\small
\begin{tabular}{@{}llc@{}}
\toprule
\textbf{Category} & \textbf{Parameter} & \textbf{Value} \\
\midrule
\multirow{8}{*}{Architecture} & Backbone & DINOv2 ViT-S/14 \\
 & Unfrozen blocks & 6 (of 12) \\
 & Prototype bank size $K_{\max}$ & 12 \\
 & Active prototype count $K$ (typical) & 10 \\
 & Part dimension $d_p$ & 256 \\
 & Embedding dim $D$ & 768 \\
 & Teacher dim & 768 \\
 & Cluster temperature $\tau$ & 0.07 \\
\midrule
\multirow{8}{*}{Training} & Input size & $448 \times 448$ \\
 & Epochs & 60 \\
 & PK sampler ($P$, $M$) & (16, 4) \\
 & Head LR & $3 \times 10^{-4}$ \\
 & Backbone LR & $3 \times 10^{-5}$ \\
 & Weight decay & 0.05 \\
 & Warmup epochs & 5 \\
 & LR floor & 1\% of peak \\
\midrule
\multirow{8}{*}{Loss} & Proxy margin $\delta$ & 0.1 \\
 & Proxy scale $\alpha$ & 32 \\
 & EMA decay $\alpha$ & 0.996 \\
 & MAR mask ratio & 0.30 \\
 & MAR warmup & 10 epochs \\
 & Distillation temp & 4.0 \\
 & Num loss groups & 4 \\
 & Label smoothing & 0.1 \\
\bottomrule
\end{tabular}
\end{table}

\FloatBarrier
\section{Extended Quantitative Results}
\label{sec:app_results}
\subsection{University-1652 Directional Comparison}
\label{sec:appendix_uni_fullrank}

Table~\ref{tab:uni1652_fullrank} extends the main-paper Table~\ref{tab:u1652_denseuav}(a) with a wider set of University-1652 baselines and reports both retrieval directions for each.

\begin{table*}[ht]
\centering
\caption{\textbf{University-1652 directional comparison.}
We report both retrieval directions, Drone$\to$Satellite and Satellite$\to$Drone,
using R@1 and AP. ``--'' indicates values not reported in the cited papers/tables
under the exact matched protocol. \textcolor{skyred}{\textbf{Red}}: best;
\textcolor{skyblue}{\textbf{blue}}: second-best.}
\label{tab:uni1652_fullrank}

\small
\setlength{\tabcolsep}{4.2pt}
\renewcommand{\arraystretch}{1.08}

\resizebox{\textwidth}{!}{%
\begin{tabular}{@{}llrcccc@{}}
\toprule
\multirow{2}{*}{\textbf{Method}}
&
\multirow{2}{*}{\textbf{Venue}}
&
\multirow{2}{*}{\textbf{Params}}
&
\multicolumn{2}{c}{\textbf{Drone $\rightarrow$ Satellite}}
&
\multicolumn{2}{c}{\textbf{Satellite $\rightarrow$ Drone}} \\
\cmidrule(lr){4-5}
\cmidrule(lr){6-7}

&
&
&
\textbf{R@1}
&
\textbf{AP}
&
\textbf{R@1}
&
\textbf{AP} \\
\midrule

ViT~\citep{dosovitskiy2021vit}
& ICLR'21
& --
& \hc{74.09}
& \hc{77.82}
& \hc{83.31}
& \hc{72.27} \\

FSRA~\citep{dai2022fsra}
& TCSVT'22
& --
& \hc{85.50}
& \hc{87.53}
& \hc{89.73}
& \hc{84.94} \\

Swin-B
& ICCV'21
& --
& \hc{84.15}
& \hc{86.62}
& \hc{90.30}
& \hc{83.55} \\

LPN~\citep{wang2022lpn}
& TCSVT'22
& 62.4\,M
& \hc{77.71}
& \hc{80.80}
& \hc{90.30}
& \hc{78.78} \\

RK-Net
& TIP'22
& --
& \hc{68.10}
& \hc{71.53}
& \hc{80.96}
& \hc{69.35} \\

SwinV2-B
& CVPR'22
& --
& \hc{86.99}
& \hc{89.02}
& \hc{91.16}
& \hc{85.77} \\

Sample4Geo (ResNet-50)~\citep{deuser2023sample4geo}
& ICCV'23
& --
& \hc{78.62}
& \hc{82.11}
& \hc{87.45}
& \hc{76.32} \\

Sample4Geo (ConvNeXt-B)~\citep{deuser2023sample4geo}
& ICCV'23
& 87.6\,M
& \hc{92.65}
& \hc{93.81}
& \hc{95.14}
& \hc{91.39} \\

DenseUAV baseline~\citep{dai2023denseuav}
& TIP'24
& --
& \hc{82.22}
& \hc{84.78}
& \hc{87.59}
& \hc{81.49} \\

F3-Net
& TGRS'23
& --
& \hc{78.64}
& \hc{81.60}
& --
& -- \\

Song
& GRSL'23
& --
& \hc{83.26}
& \hc{85.84}
& \hc{90.30}
& \hc{82.71} \\

AEN (w. LPN)
& SPL'24
& --
& \hc{77.40}
& \hc{80.27}
& \hc{90.30}
& \hc{76.01} \\

MuSe-Net~\citep{wang2024musenet}
& PR'24
& 50.5\,M
& \hc{74.48}
& \hc{77.83}
& \hc{88.02}
& \hc{75.10} \\

TransFG
& TGRS'24
& --
& \hc{87.92}
& \hc{89.99}
& \hc{93.37}
& \hc{87.94} \\

MFJRLN(Lcro)~\citep{ge2024multilevel}
& TGRS'24
& --
& \hc{87.61}
& \hc{89.57}
& \hc{91.07}
& \hc{86.72} \\

GeoFormer~\citep{li2024geoformer}
& JSTARS'24
& --
& \hc{88.16}
& \hc{90.03}
& \hc{91.87}
& \hc{87.92} \\

MCCG~\citep{shen2023mccg}
& TCSVT'24
& 56.6\,M
& \hc{89.28}
& \hc{91.01}
& \hc{94.29}
& \hc{89.29} \\

SDPL~\citep{chen2024sdpl}
& TCSVT'24
& --
& \hc{90.16}
& \hc{91.64}
& \hc{93.58}
& \hc{89.45} \\

CCR~\citep{du2024ccr}
& TCSVT'24
& 156.6\,M
& \hc{92.54}
& \hc{93.78}
& \hc{95.15}
& \hc{91.80} \\

CAMP~\citep{yang2024camp}
& TGRS'24
& 91.4\,M
& \hc{94.46}
& \hc{95.38}
& \hc{96.15}
& \hc{92.72} \\

DAC~\citep{xia2024dac}
& TCSVT'24
& 96.5\,M
& \hsecond{94.67}
& \hsecond{95.50}
& \hsecond{96.43}
& \hsecond{93.79} \\

MEAN~\citep{chen2025multilevel}
& TGRS'25
& \ssecond{36.5\,M}
& \hc{93.55}
& \hc{94.53}
& \hc{96.01}
& \hc{92.08} \\

\midrule

\textbf{\ours{} (Ours)}
& \textbf{--}
& \sbest{26.95\,M}
& \hbest{96.47}
& \hbest{97.77}
& \hbest{98.43}
& \hbest{98.24} \\

\bottomrule
\end{tabular}%
}

\end{table*}

\subsection{DenseUAV Detailed Comparison}
\label{sec:appendix_denseuav_detail}

Table~\ref{tab:denseuav_detailed} extends the main-paper Table~\ref{tab:u1652_denseuav}(b) with additional DenseUAV metrics (R@5, R@top1, AP, SDM@\{1,3,5\}) for a broader set of baselines with available detailed metrics.

\begin{table*}[ht]
\caption{\textbf{DenseUAV detailed comparison} with additional retrieval and SDM metrics. Top and second-best entries follow the same highlighting rule as the main tables. \textcolor{skyred}{\textbf{Red}}: best; \textcolor{skyblue}{\textbf{blue}}: second.}
\label{tab:denseuav_detailed}
\centering
\small
\setlength{\tabcolsep}{3.6pt}
\renewcommand{\arraystretch}{1.03}
\resizebox{\textwidth}{!}{%
\begin{tabular}{@{}lllccccccc@{}}
\toprule
\textbf{Method} & \textbf{Venue} & \textbf{Backbone} & \textbf{R@1} & \textbf{R@5} & \textbf{R@top1} & \textbf{AP} & \textbf{SDM@1} & \textbf{SDM@3} & \textbf{SDM@5} \\
\midrule
Triplet Loss            & CVPR'15     & ResNet-50  & \hc{11.88} & \hc{32.22} & --         & --        & \hc{21.91} & --        & -- \\
Instance Loss           & ACMMM'20    & ResNet-50  & \hc{13.00} & \hc{35.78} & --         & --        & \hc{23.61} & --        & -- \\
LCM                     & Remote Sens'21 & ResNet-50 & \hc{25.37} & \hc{50.92} & --         & --        & \hc{35.52} & --        & -- \\
MSBA                    & Remote Sens'21 & ResNet-50 & \hc{46.13} & \hc{64.22} & --         & --        & \hc{52.64} & --        & -- \\
LPN (ResNet-50)         & TCSVT'22    & ResNet-50  & \hc{32.43} & \hc{56.80} & --         & --        & \hc{40.26} & --        & -- \\
LPN (ViT-S)             & TCSVT'22    & ViT-S      & \hc{71.77} & \hc{90.13} & --         & --        & \hc{77.95} & --        & -- \\
FSRA                    & TCSVT'22    & ViT-S      & \hc{81.21} & \hc{94.55} & \hc{99.89} & \hc{71.93} & \hc{85.11} & \hc{83.54} & \hc{79.74} \\
RK-Net                  & TIP'22      & ResNet-50  & \hc{38.74} & \hc{62.85} & --         & --        & \hc{45.78} & --        & -- \\
Sample4Geo              & ICCV'23     & ConvNeXt-B & \hc{49.38} & \hc{78.29} & \hc{99.40} & \hc{35.93} & \hc{61.72} & --        & -- \\
DenseUAV baseline       & TIP'24      & ViT-S      & \hc{83.01} & \hc{95.58} & \hsecond{99.91} & \hc{72.10} & \hc{86.50} & \hc{84.50} & \hsecond{80.44} \\
MCCG                    & TCSVT'24    & ConvNeXt-T & \hc{83.14} & \hc{93.39} & \hc{99.74} & \hc{72.60} & \hc{85.94} & \hc{84.32} & \hc{80.14} \\
Yang \etal              & RA-L'25     & DINOv2-B   & \hc{86.27} & \hsecond{96.83} & -- & -- & \hc{88.87} & -- & -- \\
CEUSP                 & PRCV'25          & ConvNeXt-T & \hsecond{89.45} & \hc{96.05} & \hbest{100.00} & \hsecond{79.62} & \hsecond{91.01} & \hsecond{89.42} & \hbest{85.34} \\
\midrule
\textbf{\ours{} (Ours)}        & \textbf{--} & \textbf{DINOv2-S} & \hbest{91.85} & \hbest{97.81} & \hc{99.61} & \hbest{92.52} & \hbest{93.59} & \hbest{92.40} & \hc{79.33} \\
\bottomrule
\end{tabular}}
\end{table*}

\subsection{Backbone Ablation}
\label{sec:appendix_backbone}

Table~\ref{tab:backbone} reports a controlled \emph{SSL ViT generalisation} ablation on SUES-200 with everything else held identical: resolution, loss grouping, augmentation recipe, optimizer, epoch budget, and confusion gallery. The same Full \ours{} head is attached to three ViT-S self-supervised backbones (DINOv2~\citep{oquab2024dinov2}, iBOT~\citep{zhou2022ibot}, MoCo v3~\citep{chen2021mocov3}) under a strictly matched 60-epoch budget with teacher distillation \emph{disabled} (CLS-only baselines are 60 epochs, same recipe), so the head's plug-and-play behavior is read off the lift over each backbone's own CLS-only rather than off absolute R@1. The deployed configuration (DINOv2 ViT-S/14 + Full head, 60 epochs, distillation \emph{enabled}) is not retrained here; its headline number (Mean R@1 98.725\% D$\to$S) is the row in Table~\ref{tab:sues200}, and the DINOv2 row of Table~\ref{tab:backbone} (98.48\%) is the within-block SSL anchor only. We omit a ConvNeXt-T row because, in our pilot, the Full \ours{} head on ImageNet-supervised ConvNeXt-T regressed from 92.33\% (CLS-only) to 90.41\% Mean R@1 under the same 60-epoch recipe, a negative transfer consistent with the prototype-to-patch assignment requiring patch-level semantic coherence that emerges in self-distilled ViTs but not in supervised CNN feature maps. Per-altitude D$\to$S and S$\to$D R@1 are reported for each row.

\begin{table*}[ht]
\caption{\textbf{Backbone generalisation on SUES-200}
(Normal weather, single-scale $448{\times}448$, full 200-tile gallery, no re-ranking, no TTA).
We compare self-supervised ViT backbones under a controlled setting
(Full = 60 epochs, teacher distillation disabled; CLS-only baseline = 60 epochs with the same recipe).
The proposed head transfers cleanly between self-distilled patch substrates
(DINOv2, iBOT) and provides a large lift on the contrastive MoCo v3 substrate,
whose CLS embedding is substantially weaker for dense retrieval.
The deployed \ours{} model, trained for 60 epochs with teacher distillation enabled,
is reported as the headline result.}
\label{tab:backbone}
\centering
\scriptsize
\setlength{\tabcolsep}{2.5pt}
\renewcommand{\arraystretch}{0.95}

\resizebox{\textwidth}{!}{%
\begin{tabular}{@{}ll*{5}{c}*{5}{c}@{}}
\toprule
& & \multicolumn{5}{c}{\textbf{Drone $\to$ Satellite, R@1 (\%)}} 
& \multicolumn{5}{c}{\textbf{Satellite $\to$ Drone, R@1 (\%)}} \\
\cmidrule(lr){3-7}\cmidrule(lr){8-12}
\textbf{Backbone} & \textbf{Head}
& \textbf{150m} & \textbf{200m} & \textbf{250m} & \textbf{300m} 
& \textbf{Mean}
& \textbf{150m} & \textbf{200m} & \textbf{250m} & \textbf{300m} 
& \textbf{Mean} \\
\midrule

DINOv2 ViT-S/14
& CLS-only
& 82.20 & 88.08 & 91.80 & 92.65
& 88.68
& 97.50 & 93.75 & 97.50 & 96.25
& 96.25 \\

\textbf{DINOv2 ViT-S/14}
& \textbf{Full \ours{}}
& \hbest{96.25} & \hbest{98.75} & \hbest{99.30} & \hbest{99.60}
& \hbest{98.48}
& \hbest{100.00} & \hbest{100.00} & \hbest{100.00} & \hbest{100.00}
& \hbest{100.00} \\

& \emph{$\Delta$ vs CLS}
& \emph{$+$14.05} & \emph{$+$10.67} & \emph{$+$7.50} & \emph{$+$6.95}
& \emph{$+$9.80}
& \emph{$+$2.50} & \emph{$+$6.25} & \emph{$+$2.50} & \emph{$+$3.75}
& \emph{$+$3.75} \\

\addlinespace[0.2em]

iBOT ViT-S/16
& CLS-only
& 73.80 & 85.40 & 88.75 & 90.20
& 84.54
& 88.75 & 91.25 & 92.50 & 93.75
& 91.56 \\

iBOT ViT-S/16
& Full \ours{}  
& 82.43 & 91.10 & 93.63 & 93.50
& 90.16
& 96.25 & 97.50 & 97.50 & 97.50
& 97.19 \\

& \emph{$\Delta$ vs CLS}
& \emph{$+$8.63} & \emph{$+$5.70} & \emph{$+$4.88} & \emph{$+$3.30}
& \emph{$+$5.62}
& \emph{$+$7.50} & \emph{$+$6.25} & \emph{$+$5.00} & \emph{$+$3.75}
& \emph{$+$5.63} \\

\addlinespace[0.2em]

MoCo v3 ViT-S/16
& CLS-only
& 20.90 & 28.55 & 31.85 & 31.48
& 28.19
& 36.25 & 50.00 & 55.00 & 58.75
& 50.00 \\

MoCo v3 ViT-S/16
& Full \ours{}
& 48.25 & 61.80 & 66.98 & 69.68
& 61.68
& 71.25 & 82.50 & 87.50 & 86.25
& 81.88 \\

& \emph{$\Delta$ vs CLS}
& \emph{$+$27.35} & \emph{$+$33.25} & \emph{$+$35.13} & \emph{$+$38.20}
& \emph{$+$33.49}
& \emph{$+$35.00} & \emph{$+$32.50} & \emph{$+$32.50} & \emph{$+$27.50}
& \emph{$+$31.88} \\

\bottomrule
\end{tabular}%
}

\end{table*}

The three SSL substrates tell a single story sharpened by the patch-token quality of each backbone. On the two self-distilled substrates (DINOv2, iBOT), CLS-only is already moderately strong (88.68\,/\,84.54\% Mean D$\to$S R@1), and the Full \ours{} head delivers a clean positive lift across all altitudes ($+9.80$\,pp on DINOv2, $+5.62$\,pp on iBOT)-the head adds usable layout structure on top of an already coherent patch grid, and the deployed 98.725\% Mean R@1 on D$\to$S (Table~\ref{tab:sues200}) comes from combining this DINOv2 patch substrate with the teacher-distillation pathway that this ablation deliberately strips out. On the contrastive substrate (MoCo v3), CLS-only collapses to 28.19\% Mean R@1 because contrastive pretraining optimises the CLS slot but leaves patch tokens diffuse; the Full \ours{} head recovers the structural signal that the patch-affinity graph and prototype assignment can read out and lifts D$\to$S Mean R@1 by $+33.49$\,pp (and S$\to$D Mean R@1 by $+31.88$\,pp). Combined with the negative-transfer outcome on ImageNet-supervised ConvNeXt-T noted above, the picture is consistent: \ours{}'s contribution is \emph{architectural} (it activates the latent layout signal that any patch-token SSL ViT exposes), and the magnitude of the lift scales inversely with how much of that signal the global CLS slot has already captured.

\subsection{Weather Robustness: Full Protocol \& Implementation}
\label{sec:appendix_weather}

\begin{figure}[ht]
  \centering
  \includegraphics[width=\columnwidth]{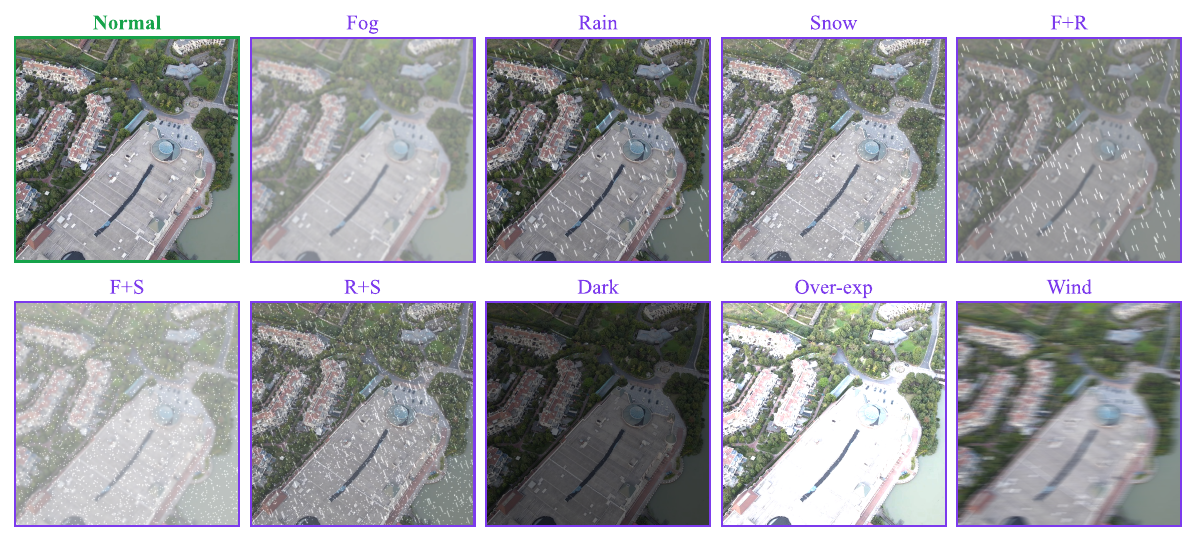}
  \caption{\textbf{Weather conditions.} The same drone image under 10 WeatherPrompt augmentations. Texture is destroyed, but spatial structure persists, a pattern qualitatively aligned with layout-heavy representations and with \ours{}'s relative robustness under environmental noise.}
  \label{fig:weather_samples}
\end{figure}

\begin{figure*}[t]
  \centering
  \includegraphics[width=\textwidth]{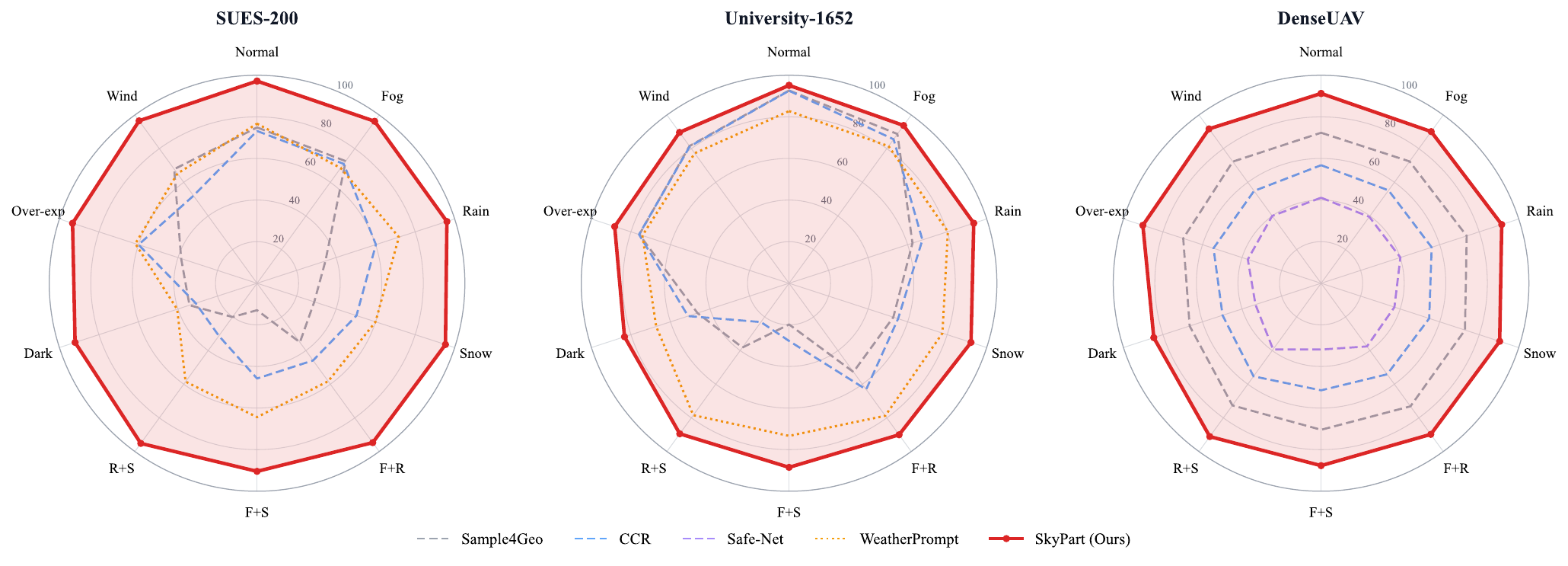}
  \caption{\textbf{Weather robustness across three benchmarks (radar view).} Per-condition Drone$\to$Satellite R@1 (\%) under the ten WeatherPrompt conditions on SUES-200, University-1652, and DenseUAV. \ours{} (red, filled) maintains a near-circular profile, indicating uniform robustness across all conditions, while baselines collapse on hard regimes (F+S, Dark). Numerical breakdown matches Table~\ref{tab:weather}.}
  \label{fig:weather_radar}
\end{figure*}

\subsubsection{Evaluation Protocol}
The evaluation protocol follows the WeatherPrompt guidelines: the satellite gallery remains clean while drone queries are corrupted on-the-fly (Fig.~\ref{fig:weather_samples}), testing generalization to adversarial conditions without access to weather-augmented satellite data. Evaluation spans ten conditions-individual effects (fog, rain, snow, darkness, over-exposure, wind) and three pairwise combinations (fog+rain, fog+snow, rain+snow)-using Recall and mean Average Precision as standard metrics. Per-condition robustness across the three benchmarks is summarized in Fig.~\ref{fig:weather_radar}.

\subsubsection{Weather Augmentation Implementation}
All augmentations use identical \texttt{imgaug} parameters as the WeatherPrompt source code (Table~\ref{tab:app_weather_params}). We apply this corruption pipeline to \ours{} on all three datasets, and to the DenseUAV baselines that we retrained end-to-end (no published WeatherPrompt numbers exist for DenseUAV); SUES-200 and University-1652 baselines are taken directly from WeatherPrompt at their published $384{\times}384$ resolution and are not re-trained, while \ours{} is fine-tuned at $392{\times}392$; the full per-block recipe (clean retrieval, DenseUAV weather fine-tune, SUES-200/University-1652 weather) is summarized in Table~\ref{tab:weather_setup}.
\textbf{Resolution and retraining scope.} On SUES-200 and University-1652, baselines are quoted from WeatherPrompt at the $384{\times}384$ setting at which their published weights and schedules were tuned, and we did not retrain them: the supervised-ViT baselines interpolate positional encodings and degrade if retrained off-spec, and the ConvNeXt-B baselines are published only at $384{\times}384$. \ours{} is trained at $392{\times}392$ because DINOv2 ViT-S/14 has patch size $14$, so $384$ is not patch-aligned; the two patch-aligned sizes bracketing $384$ are $378{=}27{\times}14$ and $392{=}28{\times}14$, and we report $392$; the SUES-200 / University-1652 weather numbers should be read as reported-baseline at each method's published resolution rather than fully resolution-controlled, as marked in the caption of Table~\ref{tab:weather}; matched-resolution and within-model resolution controls are in Appendix~\ref{sec:app_resolution}. On DenseUAV (no published WeatherPrompt baselines), all methods including \ours{} are retrained end-to-end under a matched budget: same 60-epoch schedule, same batch size, identical \texttt{imgaug} seeds (fixed per batch), and the official full DenseUAV gallery, with each method at its own published input resolution and learning rate (Table~\ref{tab:weather_setup}). That block is therefore training-budget-controlled rather than resolution-controlled. The three seeds (42, 123, 2024) used for the main SUES-200 result are re-used for each DenseUAV baseline; across those baselines, the seed-to-seed R@1 range (max minus min) is under $0.4$\,pp. Weather-condition sampling is uniform per batch (10 conditions). Augmentation seeds are released with the training scripts so retraining any method under an exactly matched regimen is reproducible end-to-end.

\begin{table}[ht]
\caption{Weather augmentation parameters (identical to WeatherPrompt).}
\label{tab:app_weather_params}
\centering
\scriptsize
\begin{tabular}{@{}lll@{}}
\toprule
\textbf{Condition} & \textbf{imgaug Augmenter} & \textbf{Parameters} \\
\midrule
Normal & \texttt{iaa.Identity()} & - \\
Fog & \texttt{iaa.Fog()} & Default parameters \\
Rain & \texttt{iaa.Rain()} & speed=(0.1, 0.3) \\
Snow & \texttt{iaa.Snowflakes()} & flake\_size=(0.1, 0.4), speed=(0.01, 0.05) \\
Dark & \texttt{iaa.Multiply()} & (0.3, 0.5) \\
Over-exp & \texttt{iaa.Multiply()} & (1.5, 2.0) \\
Fog+Rain & \texttt{Sequential([Fog, Rain])} & Same as individual \\
Fog+Snow & \texttt{Sequential([Fog, Snow])} & Same as individual \\
Rain+Snow & \texttt{Sequential([Rain, Snow])} & Same as individual \\
Wind & \texttt{iaa.MotionBlur()} & k=15, angle=[-45, 45] \\
\bottomrule
\end{tabular}
\end{table}

\subsubsection{Training Setup (Fine-tuning)}
For robustness experiments, baseline numbers are taken from WeatherPrompt~\citep{wen2025weatherprompt} at their reported configuration; \ours{} is fine-tuned for 60 epochs on SUES-200 at $392{\times}392$ under a matched optimizer, batch size, and schedule.
Weather augmentations are applied online to the drone images only, while the satellite view remains clean, and all ten conditions are sampled uniformly per batch.
For DenseUAV, we use the same optimizer (AdamW), batch size, and cosine learning-rate schedule across all retrained methods, scaling the base learning rate down by a factor of 0.3 relative to clean training to stabilize adaptation.
This matched DenseUAV schedule ensures that \ours{}'s weather gains there reflect its part-centric design rather than extra training budget; for SUES-200 and University-1652, the comparison should be read as reported-baseline (Table~\ref{tab:weather} caption).

\begin{table}[ht]
\caption{\textbf{Experiment recipes for the main benchmark results.} The weather-aligned ablation in Table~\ref{tab:ablation} uses a separate $448{\times}448$ run with online weather augmentation. DenseUAV weather runs use AdamW, a cosine learning-rate schedule, and identical \texttt{imgaug} augmentations.}
\label{tab:weather_setup}
\centering
\scriptsize
\setlength{\tabcolsep}{2.5pt}
\begin{tabular}{@{}llcccc@{}}
\toprule
\textbf{Method} & \textbf{Backbone} & \textbf{Res.} & \textbf{Epochs} & \textbf{Base LR} & \textbf{Batch} \\
\midrule
\multicolumn{6}{@{}l}{\emph{Clean retrieval: SUES-200, University-1652, DenseUAV} (Tables~\ref{tab:sues200}, \ref{tab:u1652_denseuav})} \\
\ours{} (deployed) & DINOv2-S & $448{\times}448$ & 60 & $3\!\times\!10^{-4}$ & 64 \\
Reported baselines, SUES-200 / Uni-1652$^{a}$ & as published & $256$--$512$ & -- & -- & -- \\
Reported baselines, DenseUAV$^{b}$ & as published & $256{\times}256$ & -- & -- & -- \\
\midrule
\multicolumn{6}{@{}l}{\emph{Weather, DenseUAV} (baselines of Table~\ref{tab:weather} retrained end-to-end; Table~\ref{tab:weather_dense_full})} \\
Sample4Geo~\citep{deuser2023sample4geo} & ConvNeXt-B & $384{\times}384$ & 60 & $3\!\times\!10^{-5}$ & 64 \\
Safe-Net~\citep{lin2025safe} & ResNet-50 & $256{\times}256$ & 60 & $1\!\times\!10^{-4}$ & 64 \\
CCR~\citep{du2024ccr} & ConvNeXt-B & $384{\times}384$ & 60 & $3\!\times\!10^{-5}$ & 64 \\
WeatherPrompt~\citep{wen2025weatherprompt} & ConvNeXt-B & $384{\times}384$ & 60 & $3\!\times\!10^{-5}$ & 64 \\
\ours{} (fine-tune) & DINOv2-S & $392{\times}392$ & 60 & $1\!\times\!10^{-4}$ & 64 \\
\midrule
\multicolumn{6}{@{}l}{\emph{Weather, SUES-200 / University-1652} (Tables~\ref{tab:weather_sues_full}, \ref{tab:weather_uni_full})} \\
WeatherPrompt baselines~\citep{wen2025weatherprompt} & ConvNeXt-B & $384{\times}384$ & \multicolumn{3}{c}{as published} \\
\ours{} (fine-tune) & DINOv2-S & $392{\times}392$ & 60 & $1\!\times\!10^{-4}$ & 64 \\
\bottomrule
\end{tabular}
\par\smallskip\raggedright\scriptsize $^{a}$Native input of each paper; LPN, MCCG, and CCR average flipped test features. $^{b}$Numbers as reported in the comparison of~\citep{xu2025precise}.
\end{table}
\begin{table*}[!t]
\caption{\textbf{Full WeatherPrompt robustness on SUES-200.} Per-condition R@1/AP (\%) for both D$\to$S and S$\to$D. Baselines are reported from WeatherPrompt~\citep{wen2025weatherprompt}; \ours{} is evaluated under the same corruption pipeline at $392{\times}392$. This is the detailed SUES-200 version summarized in Table~\ref{tab:weather}. \textcolor{skyred}{\textbf{Red}}: best; \textcolor{skyblue}{\textbf{blue}}: second.}
\label{tab:weather_sues_full}
\centering
\footnotesize
\setlength{\tabcolsep}{1.6pt}
\renewcommand{\arraystretch}{1.05}
\resizebox{\textwidth}{!}{%
\begin{tabular}{@{}l *{11}{cc}@{}}
\toprule
\multirow{2}{*}{\textbf{Method}}
& \multicolumn{2}{c}{\textbf{Clean}}
& \multicolumn{2}{c}{\textbf{Fog}}
& \multicolumn{2}{c}{\textbf{Rain}}
& \multicolumn{2}{c}{\textbf{Snow}}
& \multicolumn{2}{c}{\textbf{F+R}}
& \multicolumn{2}{c}{\textbf{F+S}}
& \multicolumn{2}{c}{\textbf{R+S}}
& \multicolumn{2}{c}{\textbf{Dark}}
& \multicolumn{2}{c}{\textbf{Over-exp}}
& \multicolumn{2}{c}{\textbf{Wind}}
& \multicolumn{2}{c}{\textbf{Mean}} \\
\cmidrule(lr){2-3}\cmidrule(lr){4-5}\cmidrule(lr){6-7}\cmidrule(lr){8-9}
\cmidrule(lr){10-11}\cmidrule(lr){12-13}\cmidrule(lr){14-15}\cmidrule(lr){16-17}
\cmidrule(lr){18-19}\cmidrule(lr){20-21}\cmidrule(lr){22-23}
& \scriptsize R@1 & \scriptsize AP
& \scriptsize R@1 & \scriptsize AP
& \scriptsize R@1 & \scriptsize AP
& \scriptsize R@1 & \scriptsize AP
& \scriptsize R@1 & \scriptsize AP
& \scriptsize R@1 & \scriptsize AP
& \scriptsize R@1 & \scriptsize AP
& \scriptsize R@1 & \scriptsize AP
& \scriptsize R@1 & \scriptsize AP
& \scriptsize R@1 & \scriptsize AP
& \scriptsize R@1 & \scriptsize AP \\
\midrule

\multicolumn{23}{c}{\cellcolor{gray!10}\textbf{Drone $\rightarrow$ Satellite}} \\

Zheng et al.~\citep{zheng2020university1652}
& \hc{57.70} & \hc{58.30}
& \hc{48.63} & \hc{49.61}
& \hc{53.41} & \hc{52.72}
& \hc{41.78} & \hc{43.47}
& \hc{37.17} & \hc{37.44}
& \hc{44.22} & \hc{46.18}
& \hc{40.60} & \hc{40.63}
& \hc{23.81} & \hc{25.45}
& \hc{49.79} & \hc{50.64}
& \hc{47.42} & \hc{48.31}
& \hc{44.43} & \hc{45.12} \\

IBN-Net
& \hc{65.34} & \hc{63.78}
& \hc{56.03} & \hc{56.57}
& \hc{55.73} & \hc{58.55}
& \hc{47.80} & \hc{49.53}
& \hc{43.45} & \hc{44.98}
& \hc{50.04} & \hc{51.00}
& \hc{45.51} & \hc{45.92}
& \hc{29.61} & \hc{30.93}
& \hc{56.01} & \hc{56.96}
& \hc{57.36} & \hc{58.10}
& \hc{50.69} & \hc{51.63} \\

Sample4Geo~\citep{deuser2023sample4geo}
& \hc{74.93} & \hsecond{78.76}
& \hc{72.58} & \hsecond{76.44}
& \hc{34.60} & \hc{41.56}
& \hc{28.95} & \hc{35.02}
& \hc{35.10} & \hc{41.47}
& \hc{12.95} & \hc{17.90}
& \hc{20.05} & \hc{25.95}
& \hc{34.18} & \hc{38.99}
& \hc{38.40} & \hc{43.68}
& \hsecond{67.80} & \hsecond{72.41}
& \hc{41.95} & \hc{47.22} \\

Safe-Net~\citep{lin2025safe}
& \hc{76.31} & \hc{75.35}
& \hsecond{73.53} & \hc{73.44}
& \hc{54.15} & \hc{55.05}
& \hc{48.94} & \hc{50.10}
& \hc{45.12} & \hc{47.92}
& \hc{40.05} & \hc{40.18}
& \hc{25.95} & \hc{26.12}
& \hc{29.74} & \hc{31.48}
& \hc{54.86} & \hc{58.68}
& \hc{58.10} & \hc{58.95}
& \hc{50.68} & \hc{51.63} \\

CCR~\citep{du2024ccr}
& \hc{73.22} & \hc{74.53}
& \hc{70.95} & \hc{73.14}
& \hc{60.14} & \hc{64.95}
& \hc{50.31} & \hc{53.12}
& \hc{45.87} & \hc{49.14}
& \hc{45.80} & \hc{47.87}
& \hc{31.25} & \hc{32.94}
& \hc{31.03} & \hc{34.36}
& \hc{59.97} & \hc{61.07}
& \hc{52.02} & \hc{53.33}
& \hc{52.06} & \hc{54.45} \\

MuSe-Net
& \hc{66.07} & \hc{67.02}
& \hc{58.49} & \hc{59.65}
& \hc{58.94} & \hc{60.14}
& \hc{54.85} & \hc{56.12}
& \hc{44.31} & \hc{45.82}
& \hc{49.81} & \hc{51.26}
& \hc{49.42} & \hc{50.87}
& \hc{29.34} & \hc{31.03}
& \hc{55.02} & \hc{56.36}
& \hc{59.97} & \hc{61.05}
& \hc{52.62} & \hc{53.93} \\

WeatherPrompt~\citep{wen2025weatherprompt}
& \hsecond{76.72} & \hc{75.51}
& \hc{68.49} & \hc{68.87}
& \hsecond{71.77} & \hsecond{71.20}
& \hsecond{59.95} & \hsecond{60.62}
& \hsecond{58.24} & \hsecond{58.83}
& \hsecond{64.36} & \hsecond{66.27}
& \hsecond{58.49} & \hsecond{58.89}
& \hsecond{40.42} & \hsecond{55.75}
& \hsecond{61.57} & \hsecond{71.70}
& \hc{65.19} & \hc{67.00}
& \hsecond{62.52} & \hsecond{65.46} \\

\textbf{\ours{} (Ours)}
& \hbest{97.21} & \hbest{98.10}
& \hbest{96.27} & \hbest{97.59}
& \hbest{96.06} & \hbest{97.53}
& \hbest{95.33} & \hbest{97.05}
& \hbest{94.71} & \hbest{96.75}
& \hbest{90.44} & \hbest{93.85}
& \hbest{95.17} & \hbest{96.94}
& \hbest{92.06} & \hbest{94.50}
& \hbest{93.34} & \hbest{95.85}
& \hbest{96.54} & \hbest{97.80}
& \hbest{94.71} & \hbest{96.60} \\

\midrule
\multicolumn{23}{c}{\cellcolor{gray!10}\textbf{Satellite $\rightarrow$ Drone}} \\

Zheng et al.~\citep{zheng2020university1652}
& \hc{70.20} & \hc{57.98}
& \hc{63.77} & \hc{46.90}
& \hc{68.72} & \hc{50.85}
& \hc{61.72} & \hc{39.70}
& \hc{62.10} & \hc{32.75}
& \hc{71.70} & \hc{40.39}
& \hc{59.72} & \hc{37.55}
& \hc{45.49} & \hc{25.28}
& \hc{52.11} & \hc{43.40}
& \hc{56.62} & \hc{45.31}
& \hc{61.21} & \hc{42.01} \\

IBN-Net
& \hc{73.68} & \hc{62.91}
& \hc{67.41} & \hc{55.75}
& \hc{72.30} & \hc{56.44}
& \hc{64.07} & \hc{47.69}
& \hc{66.98} & \hc{39.54}
& \hc{71.10} & \hc{47.32}
& \hc{68.46} & \hc{45.95}
& \hc{54.72} & \hc{31.53}
& \hc{65.64} & \hc{53.77}
& \hc{73.48} & \hc{57.03}
& \hc{67.79} & \hc{49.79} \\

Sample4Geo~\citep{deuser2023sample4geo}
& \hc{87.50} & \hc{79.57}
& \hc{83.75} & \hc{71.14}
& \hc{42.50} & \hc{25.24}
& \hc{40.00} & \hc{21.59}
& \hc{38.75} & \hc{23.22}
& \hc{30.00} & \hc{10.58}
& \hc{26.25} & \hc{16.44}
& \hc{56.25} & \hc{29.75}
& \hc{58.75} & \hc{30.38}
& \hsecond{83.75} & \hsecond{69.66}
& \hc{54.75} & \hc{37.76} \\

Safe-Net~\citep{lin2025safe}
& \hc{88.31} & \hc{80.35}
& \hc{81.33} & \hc{68.60}
& \hc{40.21} & \hc{41.04}
& \hc{36.43} & \hc{37.50}
& \hc{33.12} & \hc{35.45}
& \hc{24.78} & \hc{27.65}
& \hc{41.12} & \hc{32.31}
& \hc{53.88} & \hc{27.01}
& \hc{54.19} & \hc{57.82}
& \hc{79.36} & \hc{57.09}
& \hc{53.27} & \hc{46.48} \\

CCR~\citep{du2024ccr}
& \hc{90.59} & \hc{80.45}
& \hc{82.99} & \hc{70.62}
& \hc{43.39} & \hc{45.90}
& \hc{39.81} & \hc{40.88}
& \hc{42.63} & \hc{39.46}
& \hc{29.32} & \hc{30.65}
& \hc{25.89} & \hc{26.94}
& \hc{26.01} & \hc{30.40}
& \hc{58.01} & \hc{59.13}
& \hc{83.09} & \hc{61.05}
& \hc{52.17} & \hc{48.55} \\

MuSe-Net
& \hc{76.56} & \hc{66.02}
& \hc{72.19} & \hc{57.87}
& \hc{72.19} & \hc{58.11}
& \hc{68.38} & \hc{51.22}
& \hc{66.56} & \hc{42.25}
& \hc{69.06} & \hc{46.80}
& \hc{69.38} & \hc{47.79}
& \hc{53.75} & \hc{27.94}
& \hc{70.00} & \hc{52.67}
& \hc{76.25} & \hc{60.74}
& \hc{69.43} & \hc{51.14} \\

WeatherPrompt~\citep{wen2025weatherprompt}
& \hsecond{90.61} & \hsecond{81.24}
& \hsecond{86.14} & \hsecond{71.15}
& \hsecond{83.94} & \hsecond{73.80}
& \hsecond{71.03} & \hsecond{60.19}
& \hsecond{84.41} & \hsecond{58.49}
& \hsecond{79.16} & \hsecond{64.93}
& \hsecond{77.28} & \hsecond{60.15}
& \hsecond{56.75} & \hsecond{47.85}
& \hsecond{81.65} & \hsecond{74.04}
& \hc{80.30} & \hc{69.38}
& \hsecond{79.13} & \hsecond{66.12} \\

\textbf{\ours{} (Ours)}
& \hbest{98.75} & \hbest{98.24}
& \hbest{98.75} & \hbest{97.56}
& \hbest{98.75} & \hbest{97.52}
& \hbest{98.75} & \hbest{96.94}
& \hbest{98.75} & \hbest{96.17}
& \hbest{98.75} & \hbest{92.11}
& \hbest{98.75} & \hbest{96.72}
& \hbest{98.75} & \hbest{93.79}
& \hbest{98.75} & \hbest{95.90}
& \hbest{98.75} & \hbest{97.21}
& \hbest{98.75} & \hbest{96.22} \\

\bottomrule
\end{tabular}}
\end{table*}

\begin{table*}[!t]
\caption{\textbf{Full WeatherPrompt robustness on University-1652.} Per-condition R@1/AP (\%) for both D$\to$S and S$\to$D. Baselines are reported from WeatherPrompt~\citep{wen2025weatherprompt}; \ours{} is evaluated under the same corruption pipeline at $392{\times}392$. This is the detailed University-1652 version summarized in Table~\ref{tab:weather}. \textcolor{skyred}{\textbf{Red}}: best; \textcolor{skyblue}{\textbf{blue}}: second.}
\label{tab:weather_uni_full}
\centering
\footnotesize
\setlength{\tabcolsep}{1.6pt}
\renewcommand{\arraystretch}{1.05}
\resizebox{\textwidth}{!}{%
\begin{tabular}{@{}l *{11}{cc}@{}}
\toprule
\multirow{2}{*}{\textbf{Method}}
& \multicolumn{2}{c}{\textbf{Clean}}
& \multicolumn{2}{c}{\textbf{Fog}}
& \multicolumn{2}{c}{\textbf{Rain}}
& \multicolumn{2}{c}{\textbf{Snow}}
& \multicolumn{2}{c}{\textbf{F+R}}
& \multicolumn{2}{c}{\textbf{F+S}}
& \multicolumn{2}{c}{\textbf{R+S}}
& \multicolumn{2}{c}{\textbf{Dark}}
& \multicolumn{2}{c}{\textbf{Over-exp}}
& \multicolumn{2}{c}{\textbf{Wind}}
& \multicolumn{2}{c}{\textbf{Mean}} \\
\cmidrule(lr){2-3}\cmidrule(lr){4-5}\cmidrule(lr){6-7}\cmidrule(lr){8-9}
\cmidrule(lr){10-11}\cmidrule(lr){12-13}\cmidrule(lr){14-15}\cmidrule(lr){16-17}
\cmidrule(lr){18-19}\cmidrule(lr){20-21}\cmidrule(lr){22-23}
& \scriptsize R@1 & \scriptsize AP
& \scriptsize R@1 & \scriptsize AP
& \scriptsize R@1 & \scriptsize AP
& \scriptsize R@1 & \scriptsize AP
& \scriptsize R@1 & \scriptsize AP
& \scriptsize R@1 & \scriptsize AP
& \scriptsize R@1 & \scriptsize AP
& \scriptsize R@1 & \scriptsize AP
& \scriptsize R@1 & \scriptsize AP
& \scriptsize R@1 & \scriptsize AP
& \scriptsize R@1 & \scriptsize AP \\
\midrule

\multicolumn{23}{c}{\cellcolor{gray!10}\textbf{Drone $\rightarrow$ Satellite}} \\

Zheng et al.~\citep{zheng2020university1652}
& \hc{67.83} & \hc{71.74}
& \hc{60.97} & \hc{65.23}
& \hc{60.29} & \hc{64.61}
& \hc{55.58} & \hc{60.09}
& \hc{54.75} & \hc{59.40}
& \hc{44.85} & \hc{49.78}
& \hc{57.61} & \hc{62.03}
& \hc{39.70} & \hc{44.65}
& \hc{51.85} & \hc{56.75}
& \hc{58.28} & \hc{62.83}
& \hc{55.17} & \hc{59.71} \\

ResNet-101
& \hc{70.07} & \hc{73.04}
& \hc{63.87} & \hc{68.22}
& \hc{63.34} & \hc{67.59}
& \hc{59.75} & \hc{64.15}
& \hc{57.45} & \hc{62.12}
& \hc{48.31} & \hc{53.28}
& \hc{60.25} & \hc{64.68}
& \hc{46.12} & \hc{51.02}
& \hc{56.34} & \hc{61.23}
& \hc{62.13} & \hc{66.63}
& \hc{58.76} & \hc{63.29} \\

DenseNet121
& \hc{69.48} & \hc{73.26}
& \hc{64.25} & \hc{68.47}
& \hc{63.47} & \hc{67.64}
& \hc{59.29} & \hc{63.70}
& \hc{59.68} & \hc{64.13}
& \hc{50.41} & \hc{55.20}
& \hc{60.21} & \hc{64.57}
& \hc{48.57} & \hc{53.41}
& \hc{54.04} & \hc{58.88}
& \hc{60.74} & \hc{65.14}
& \hc{59.01} & \hc{63.44} \\

Swin-T
& \hc{69.27} & \hc{73.18}
& \hc{66.46} & \hc{70.52}
& \hc{65.44} & \hc{69.60}
& \hc{61.79} & \hc{66.23}
& \hc{63.96} & \hc{68.21}
& \hc{56.44} & \hc{61.07}
& \hc{62.68} & \hc{67.02}
& \hc{50.27} & \hc{55.18}
& \hc{55.46} & \hc{60.29}
& \hc{63.81} & \hc{68.17}
& \hc{61.56} & \hc{65.95} \\

IBN-Net
& \hc{72.35} & \hc{75.85}
& \hc{66.68} & \hc{70.64}
& \hc{67.95} & \hc{71.73}
& \hc{62.77} & \hc{66.85}
& \hc{62.64} & \hc{66.84}
& \hc{51.09} & \hc{55.79}
& \hc{64.07} & \hc{68.13}
& \hc{50.72} & \hc{55.53}
& \hc{57.97} & \hc{62.52}
& \hc{66.73} & \hc{70.68}
& \hc{62.30} & \hc{66.46} \\

LPN~\citep{wang2022lpn}
& \hc{74.33} & \hc{77.60}
& \hc{69.31} & \hc{72.95}
& \hc{67.96} & \hc{71.72}
& \hc{64.90} & \hc{68.85}
& \hc{64.51} & \hc{68.52}
& \hc{54.16} & \hc{58.73}
& \hc{65.38} & \hc{69.29}
& \hc{53.68} & \hc{58.10}
& \hc{60.90} & \hc{65.27}
& \hc{66.46} & \hc{70.35}
& \hc{64.16} & \hc{68.14} \\

Sample4Geo~\citep{deuser2023sample4geo}
& \hsecond{92.70} & \hsecond{93.85}
& \hsecond{88.70} & \hsecond{90.55}
& \hc{62.44} & \hc{66.17}
& \hc{52.76} & \hc{57.24}
& \hc{52.70} & \hc{56.77}
& \hc{19.79} & \hc{23.16}
& \hc{38.19} & \hc{42.33}
& \hc{46.34} & \hc{49.91}
& \hc{75.77} & \hc{78.90}
& \hsecond{81.54} & \hsecond{87.34}
& \hc{61.10} & \hc{64.32} \\

Safe-Net~\citep{lin2025safe}
& \hc{86.98} & \hc{88.85}
& \hc{82.12} & \hc{86.10}
& \hc{67.13} & \hc{68.90}
& \hc{60.50} & \hc{63.01}
& \hc{54.80} & \hc{58.73}
& \hc{32.12} & \hc{39.77}
& \hc{25.83} & \hc{26.40}
& \hc{41.10} & \hc{44.13}
& \hc{69.87} & \hc{71.15}
& \hc{74.32} & \hc{76.58}
& \hc{59.48} & \hc{62.36} \\

CCR~\citep{du2024ccr}
& \hc{92.54} & \hc{93.78}
& \hc{85.57} & \hc{87.13}
& \hc{67.46} & \hc{68.82}
& \hc{55.16} & \hc{59.14}
& \hc{63.11} & \hc{60.97}
& \hc{27.74} & \hc{31.48}
& \hc{23.06} & \hc{46.85}
& \hc{51.10} & \hc{54.19}
& \hsecond{75.90} & \hsecond{79.16}
& \hc{81.31} & \hc{87.22}
& \hc{62.30} & \hc{66.87} \\

MuSe-Net
& \hc{74.48} & \hc{77.83}
& \hc{69.47} & \hc{73.24}
& \hc{70.55} & \hc{74.14}
& \hc{65.72} & \hc{69.70}
& \hc{65.59} & \hc{69.64}
& \hc{54.69} & \hc{59.24}
& \hc{66.64} & \hc{70.55}
& \hc{53.85} & \hc{58.49}
& \hc{61.05} & \hc{65.51}
& \hc{69.45} & \hc{73.22}
& \hc{65.15} & \hc{69.16} \\

WeatherPrompt~\citep{wen2025weatherprompt}
& \hc{82.78} & \hc{85.18}
& \hc{81.46} & \hc{84.03}
& \hsecond{80.34} & \hsecond{83.11}
& \hsecond{77.60} & \hsecond{80.67}
& \hsecond{78.75} & \hsecond{81.69}
& \hsecond{73.38} & \hsecond{76.94}
& \hsecond{78.41} & \hsecond{81.40}
& \hsecond{67.22} & \hsecond{71.06}
& \hc{74.20} & \hc{77.63}
& \hc{77.26} & \hc{80.27}
& \hsecond{77.14} & \hsecond{80.20} \\

\textbf{\ours{} (Ours)}
& \hbest{95.15} & \hbest{96.45}
& \hbest{93.78} & \hbest{95.65}
& \hbest{93.44} & \hbest{93.98}
& \hbest{92.05} & \hbest{93.51}
& \hbest{90.05} & \hbest{92.74}
& \hbest{88.51} & \hbest{91.86}
& \hbest{89.47} & \hbest{92.42}
& \hbest{83.26} & \hbest{87.80}
& \hbest{88.23} & \hbest{91.50}
& \hbest{89.68} & \hbest{92.54}
& \hbest{90.36} & \hbest{92.84} \\

\midrule
\multicolumn{23}{c}{\cellcolor{gray!10}\textbf{Satellite $\rightarrow$ Drone}} \\

Zheng et al.~\citep{zheng2020university1652}
& \hc{83.45} & \hc{67.94}
& \hc{79.60} & \hc{61.12}
& \hc{77.60} & \hc{59.73}
& \hc{73.18} & \hc{55.07}
& \hc{75.89} & \hc{54.45}
& \hc{70.76} & \hc{43.26}
& \hc{74.75} & \hc{56.44}
& \hc{69.47} & \hc{39.25}
& \hc{72.18} & \hc{51.91}
& \hc{76.46} & \hc{57.59}
& \hc{75.33} & \hc{54.68} \\

ResNet-101
& \hc{85.73} & \hc{71.79}
& \hc{82.45} & \hc{66.46}
& \hc{81.46} & \hc{65.68}
& \hc{79.74} & \hc{61.72}
& \hc{79.74} & \hc{60.59}
& \hc{74.75} & \hc{50.31}
& \hc{80.17} & \hc{62.61}
& \hc{75.32} & \hc{45.37}
& \hc{79.60} & \hc{58.21}
& \hc{82.31} & \hc{64.67}
& \hc{80.13} & \hc{60.74} \\

DenseNet121
& \hc{83.74} & \hc{70.34}
& \hc{82.31} & \hc{66.32}
& \hc{81.17} & \hc{65.23}
& \hc{78.60} & \hc{60.33}
& \hc{79.46} & \hc{61.66}
& \hc{74.61} & \hc{51.14}
& \hc{78.46} & \hc{61.68}
& \hc{74.47} & \hc{47.88}
& \hc{74.32} & \hc{55.26}
& \hc{78.32} & \hc{61.63}
& \hc{78.55} & \hc{60.15} \\

Swin-T
& \hc{80.74} & \hc{68.94}
& \hc{81.03} & \hc{67.46}
& \hc{81.17} & \hc{66.39}
& \hc{78.46} & \hc{61.33}
& \hc{79.17} & \hc{64.65}
& \hc{74.89} & \hc{56.57}
& \hc{78.89} & \hc{63.49}
& \hc{75.61} & \hc{48.43}
& \hc{76.60} & \hc{56.57}
& \hc{78.74} & \hc{64.45}
& \hc{78.53} & \hc{61.83} \\

IBN-Net
& \hc{86.31} & \hc{73.54}
& \hc{84.59} & \hc{67.61}
& \hc{84.74} & \hc{69.03}
& \hc{80.88} & \hc{64.44}
& \hc{83.31} & \hc{63.71}
& \hc{77.89} & \hc{52.14}
& \hc{83.02} & \hc{65.74}
& \hc{78.46} & \hc{50.77}
& \hc{79.46} & \hc{58.64}
& \hc{84.02} & \hc{67.94}
& \hc{82.27} & \hc{63.36} \\

LPN~\citep{wang2022lpn}
& \hc{87.02} & \hc{75.19}
& \hc{86.16} & \hc{71.34}
& \hc{83.88} & \hc{69.49}
& \hc{82.88} & \hc{65.39}
& \hc{84.59} & \hc{66.28}
& \hc{79.60} & \hc{55.19}
& \hc{84.17} & \hc{66.26}
& \hc{82.88} & \hc{52.05}
& \hc{81.03} & \hc{62.24}
& \hc{84.14} & \hc{67.35}
& \hc{83.64} & \hc{65.08} \\

Sample4Geo~\citep{deuser2023sample4geo}
& \hsecond{95.29} & \hc{91.42}
& \hbest{93.87} & \hsecond{87.46}
& \hc{73.04} & \hc{50.27}
& \hc{76.18} & \hc{47.58}
& \hc{71.18} & \hc{44.53}
& \hc{52.21} & \hc{16.21}
& \hc{64.48} & \hc{32.38}
& \hc{77.03} & \hc{45.89}
& \hsecond{91.58} & \hsecond{77.04}
& \hbest{93.30} & \hsecond{81.42}
& \hc{78.82} & \hc{57.42} \\

Safe-Net~\citep{lin2025safe}
& \hc{91.22} & \hc{86.06}
& \hc{90.04} & \hc{85.43}
& \hc{71.12} & \hc{68.56}
& \hc{73.26} & \hc{45.62}
& \hc{68.23} & \hc{41.78}
& \hc{49.32} & \hc{34.72}
& \hc{61.07} & \hc{29.86}
& \hc{73.15} & \hc{43.08}
& \hc{88.54} & \hc{74.65}
& \hc{90.02} & \hc{78.21}
& \hc{75.69} & \hc{58.80} \\

CCR~\citep{du2024ccr}
& \hc{95.15} & \hsecond{91.80}
& \hc{90.93} & \hc{80.62}
& \hc{81.83} & \hc{73.89}
& \hc{69.92} & \hc{65.41}
& \hc{76.92} & \hc{70.53}
& \hc{50.89} & \hc{31.64}
& \hc{61.11} & \hc{32.21}
& \hc{64.80} & \hc{46.28}
& \hc{86.01} & \hc{71.23}
& \hsecond{92.67} & \hc{76.55}
& \hc{77.02} & \hc{64.02} \\

MuSe-Net
& \hc{88.02} & \hc{75.10}
& \hc{87.87} & \hc{69.85}
& \hc{87.73} & \hc{71.12}
& \hc{83.74} & \hc{66.52}
& \hc{85.02} & \hc{67.78}
& \hc{80.88} & \hc{54.26}
& \hc{84.88} & \hc{67.75}
& \hc{80.74} & \hc{53.01}
& \hc{81.60} & \hc{62.09}
& \hc{86.31} & \hc{70.03}
& \hc{84.68} & \hc{65.75} \\

WeatherPrompt~\citep{wen2025weatherprompt}
& \hc{89.16} & \hc{81.80}
& \hc{88.73} & \hc{80.58}
& \hsecond{88.16} & \hsecond{79.87}
& \hsecond{87.59} & \hsecond{77.25}
& \hsecond{88.45} & \hsecond{78.20}
& \hsecond{86.73} & \hsecond{73.23}
& \hsecond{88.59} & \hsecond{78.14}
& \hsecond{86.59} & \hsecond{65.20}
& \hc{85.31} & \hc{73.25}
& \hc{87.88} & \hc{76.33}
& \hsecond{87.72} & \hsecond{76.39} \\

\textbf{\ours{} (Ours)}
& \hbest{97.29} & \hbest{97.08}
& \hsecond{93.15} & \hbest{90.62}
& \hbest{93.87} & \hbest{90.03}
& \hbest{92.01} & \hbest{88.51}
& \hbest{92.58} & \hbest{87.72}
& \hbest{91.87} & \hbest{83.64}
& \hbest{91.44} & \hbest{88.06}
& \hbest{92.30} & \hbest{80.18}
& \hbest{91.73} & \hbest{87.30}
& \hc{92.01} & \hbest{88.17}
& \hbest{92.83} & \hbest{88.13} \\

\bottomrule
\end{tabular}}
\end{table*}
\begin{table*}[!t]
\caption{\textbf{Full WeatherPrompt robustness on DenseUAV.} Per-condition R@1/AP (\%) for both D$\to$S and S$\to$D under the same corruption pipeline. AP denotes mean average precision over queries. This is the detailed DenseUAV version summarized in Table~\ref{tab:weather}. \textcolor{skyred}{\textbf{Red}}: best; \textcolor{skyblue}{\textbf{blue}}: second.}
\label{tab:weather_dense_full}
\centering
\footnotesize
\setlength{\tabcolsep}{1.8pt}
\renewcommand{\arraystretch}{1.05}
\resizebox{\textwidth}{!}{%
\begin{tabular}{@{}l *{11}{cc}@{}}
\toprule
\multirow{2}{*}{\textbf{Method}}
& \multicolumn{2}{c}{\textbf{Clean}}
& \multicolumn{2}{c}{\textbf{Fog}}
& \multicolumn{2}{c}{\textbf{Rain}}
& \multicolumn{2}{c}{\textbf{Snow}}
& \multicolumn{2}{c}{\textbf{F+R}}
& \multicolumn{2}{c}{\textbf{F+S}}
& \multicolumn{2}{c}{\textbf{R+S}}
& \multicolumn{2}{c}{\textbf{Dark}}
& \multicolumn{2}{c}{\textbf{Over-exp}}
& \multicolumn{2}{c}{\textbf{Wind}}
& \multicolumn{2}{c}{\textbf{Mean}} \\
\cmidrule(lr){2-3}\cmidrule(lr){4-5}\cmidrule(lr){6-7}\cmidrule(lr){8-9}
\cmidrule(lr){10-11}\cmidrule(lr){12-13}\cmidrule(lr){14-15}\cmidrule(lr){16-17}
\cmidrule(lr){18-19}\cmidrule(lr){20-21}\cmidrule(lr){22-23}
& \footnotesize R@1 & \footnotesize AP
& \footnotesize R@1 & \footnotesize AP
& \footnotesize R@1 & \footnotesize AP
& \footnotesize R@1 & \footnotesize AP
& \footnotesize R@1 & \footnotesize AP
& \footnotesize R@1 & \footnotesize AP
& \footnotesize R@1 & \footnotesize AP
& \footnotesize R@1 & \footnotesize AP
& \footnotesize R@1 & \footnotesize AP
& \footnotesize R@1 & \footnotesize AP
& \footnotesize R@1 & \footnotesize AP \\
\midrule

\multicolumn{23}{c}{\cellcolor{gray!10}\textbf{Drone $\rightarrow$ Satellite}} \\

Sample4Geo
& \hsecond{72.37} & \hsecond{66.67}
& \hsecond{72.42} & \hsecond{66.31}
& \hsecond{73.66} & \hsecond{66.84}
& \hsecond{72.80} & \hsecond{66.37}
& \hsecond{73.14} & \hsecond{66.37}
& \hsecond{70.44} & \hsecond{65.06}
& \hsecond{72.59} & \hsecond{66.47}
& \hsecond{66.50} & \hsecond{61.54}
& \hsecond{69.76} & \hsecond{64.86}
& \hsecond{72.29} & \hsecond{65.87}
& \hsecond{71.60} & \hsecond{65.64} \\

CCR
& \hc{56.76} & \hc{62.37}
& \hc{55.21} & \hc{61.35}
& \hc{55.94} & \hc{61.84}
& \hc{54.70} & \hc{61.71}
& \hc{54.10} & \hc{60.74}
& \hc{51.48} & \hc{58.33}
& \hc{55.30} & \hc{61.46}
& \hc{49.85} & \hc{56.11}
& \hc{54.44} & \hc{60.43}
& \hc{54.87} & \hc{61.42}
& \hc{54.26} & \hc{60.58} \\

IBN-Net
& \hc{41.70} & \hc{44.63}
& \hc{40.63} & \hc{43.40}
& \hc{41.57} & \hc{44.32}
& \hc{40.37} & \hc{43.36}
& \hc{41.91} & \hc{43.43}
& \hc{37.24} & \hc{40.13}
& \hc{41.14} & \hc{43.88}
& \hc{34.96} & \hc{38.25}
& \hc{38.78} & \hc{41.41}
& \hc{41.53} & \hc{44.36}
& \hc{39.98} & \hc{42.72} \\

Safe-Net
& \hc{41.14} & \hc{45.55}
& \hc{39.47} & \hc{44.78}
& \hc{40.11} & \hc{45.20}
& \hc{37.11} & \hc{42.72}
& \hc{37.54} & \hc{43.91}
& \hc{31.87} & \hc{38.39}
& \hc{39.30} & \hc{44.81}
& \hc{33.20} & \hc{37.93}
& \hc{37.07} & \hc{43.01}
& \hc{40.03} & \hc{45.15}
& \hc{37.68} & \hc{43.15} \\

MuSe-Net
& \hc{28.96} & \hc{32.56}
& \hc{26.77} & \hc{31.88}
& \hc{27.71} & \hc{32.39}
& \hc{26.94} & \hc{31.17}
& \hc{26.38} & \hc{31.41}
& \hc{21.54} & \hc{26.55}
& \hc{26.98} & \hc{31.69}
& \hc{23.25} & \hc{27.12}
& \hc{24.88} & \hc{29.45}
& \hc{28.70} & \hc{32.64}
& \hc{26.21} & \hc{30.69} \\

WeatherPrompt
& \hc{26.25} & \hc{30.68}
& \hc{24.15} & \hc{29.45}
& \hc{26.68} & \hc{30.71}
& \hc{24.28} & \hc{28.84}
& \hc{25.35} & \hc{29.29}
& \hc{20.72} & \hc{25.06}
& \hc{25.10} & \hc{29.44}
& \hc{20.08} & \hc{24.39}
& \hc{22.48} & \hc{27.16}
& \hc{25.78} & \hc{30.41}
& \hc{24.09} & \hc{28.54} \\

Zheng'20
& \hc{17.55} & \hc{22.58}
& \hc{19.18} & \hc{23.59}
& \hc{18.88} & \hc{23.38}
& \hc{18.15} & \hc{22.47}
& \hc{19.65} & \hc{23.82}
& \hc{17.72} & \hc{21.56}
& \hc{18.53} & \hc{22.98}
& \hc{14.63} & \hc{19.08}
& \hc{16.00} & \hc{20.40}
& \hc{18.53} & \hc{23.56}
& \hc{17.88} & \hc{22.34} \\

\textbf{\ours{} (Ours)}
& \hbest{91.25} & \hbest{92.10}
& \hbest{90.13} & \hbest{91.26}
& \hbest{91.38} & \hbest{91.88}
& \hbest{90.35} & \hbest{91.37}
& \hbest{89.79} & \hbest{90.71}
& \hbest{87.69} & \hbest{88.99}
& \hbest{91.12} & \hbest{91.58}
& \hbest{84.64} & \hbest{86.07}
& \hbest{90.18} & \hbest{91.20}
& \hbest{91.76} & \hbest{92.29}
& \hbest{89.83} & \hbest{90.75} \\

\midrule
\multicolumn{23}{c}{\cellcolor{gray!10}\textbf{Satellite $\rightarrow$ Drone}} \\

Sample4Geo
& \hsecond{71.51} & \hsecond{67.44}
& \hsecond{70.61} & \hsecond{66.68}
& \hsecond{71.77} & \hsecond{67.00}
& \hsecond{72.03} & \hsecond{67.01}
& \hsecond{70.79} & \hsecond{66.20}
& \hsecond{69.37} & \hsecond{64.45}
& \hsecond{71.34} & \hsecond{66.65}
& \hsecond{66.67} & \hsecond{59.81}
& \hsecond{69.33} & \hsecond{64.85}
& \hsecond{71.34} & \hsecond{65.97}
& \hsecond{70.48} & \hsecond{65.61} \\

CCR
& \hc{56.93} & \hc{60.85}
& \hc{55.38} & \hc{59.24}
& \hc{57.83} & \hc{60.27}
& \hc{56.41} & \hc{58.80}
& \hc{55.77} & \hc{58.49}
& \hc{52.17} & \hc{54.58}
& \hc{56.28} & \hc{58.80}
& \hc{52.38} & \hc{50.40}
& \hc{55.30} & \hc{55.91}
& \hc{56.54} & \hc{59.92}
& \hc{55.50} & \hc{57.73} \\

IBN-Net
& \hc{39.12} & \hc{42.83}
& \hc{37.79} & \hc{41.16}
& \hc{40.67} & \hc{42.84}
& \hc{39.64} & \hc{41.91}
& \hc{38.44} & \hc{40.79}
& \hc{35.35} & \hc{37.22}
& \hc{39.68} & \hc{42.15}
& \hc{33.42} & \hc{35.22}
& \hc{37.02} & \hc{38.56}
& \hc{39.60} & \hc{42.14}
& \hc{38.07} & \hc{40.48} \\

Safe-Net
& \hc{38.40} & \hc{40.20}
& \hc{37.19} & \hc{39.51}
& \hc{37.71} & \hc{40.06}
& \hc{36.12} & \hc{37.46}
& \hc{36.94} & \hc{38.54}
& \hc{33.25} & \hc{33.11}
& \hc{38.01} & \hc{39.38}
& \hc{32.13} & \hc{31.86}
& \hc{38.74} & \hc{37.57}
& \hc{38.78} & \hc{40.56}
& \hc{36.73} & \hc{37.83} \\

MuSe-Net
& \hc{27.03} & \hc{31.35}
& \hc{26.00} & \hc{30.00}
& \hc{27.54} & \hc{30.84}
& \hc{26.51} & \hc{29.32}
& \hc{25.57} & \hc{29.41}
& \hc{21.58} & \hc{24.20}
& \hc{26.38} & \hc{29.33}
& \hc{23.55} & \hc{24.36}
& \hc{23.08} & \hc{26.28}
& \hc{27.76} & \hc{30.56}
& \hc{25.50} & \hc{28.57} \\

WeatherPrompt
& \hc{25.57} & \hc{29.69}
& \hc{23.64} & \hc{27.57}
& \hc{24.41} & \hc{28.49}
& \hc{23.12} & \hc{26.70}
& \hc{22.82} & \hc{26.70}
& \hc{19.61} & \hc{22.43}
& \hc{22.87} & \hc{27.02}
& \hc{20.33} & \hc{21.67}
& \hc{21.32} & \hc{24.54}
& \hc{25.01} & \hc{28.77}
& \hc{22.87} & \hc{26.36} \\

Zheng et al.~\citep{zheng2020university1652}
& \hc{18.02} & \hc{20.99}
& \hc{17.76} & \hc{21.14}
& \hc{17.03} & \hc{20.78}
& \hc{18.10} & \hc{20.59}
& \hc{18.23} & \hc{21.12}
& \hc{17.07} & \hc{18.68}
& \hc{18.49} & \hc{20.85}
& \hc{14.67} & \hc{16.66}
& \hc{16.00} & \hc{17.67}
& \hc{19.22} & \hc{21.46}
& \hc{17.46} & \hc{19.99} \\

\textbf{\ours{} (Ours)}
& \hbest{92.66} & \hbest{92.34}
& \hbest{91.55} & \hbest{91.21}
& \hbest{91.51} & \hbest{91.66}
& \hbest{92.02} & \hbest{91.54}
& \hbest{90.99} & \hbest{90.43}
& \hbest{90.43} & \hbest{88.61}
& \hbest{91.93} & \hbest{91.55}
& \hbest{90.69} & \hbest{84.79}
& \hbest{92.15} & \hbest{91.30}
& \hbest{92.96} & \hbest{92.32}
& \hbest{91.69} & \hbest{90.58} \\

\bottomrule
\end{tabular}}
\end{table*}
\subsection{Zero-shot Cross-Dataset Transfer}
\label{sec:appendix_crossdata}
Table~\ref{tab:cross_dataset} shows strong transfer under this protocol:
\ours{} leads all D$\to$S columns; on S$\to$D, DAC/MEAN slightly edge R@1 at 200\,m
and MCFA retains the strongest AP at higher altitudes, so the cross-dataset picture
is mixed on S$\to$D. The point is that \ours{} attains non-trivial transfer on the
primary D$\to$S direction, supporting geographically meaningful structure beyond
pure texture memorization.

\begin{table*}[ht]
\caption{\textbf{Zero-shot cross-dataset transfer}: University-1652 $\to$ SUES-200 per altitude.
All methods are trained on University-1652 only and evaluated on SUES-200 test set
with no fine-tuning. \textbf{D$\to$S} uses the 200-satellite confusion gallery.
Baselines are from~\citep{hou2025mcfa} and the original publications.
Shading and red/blue ranks follow Table~\ref{tab:sues200}.
\textcolor{skyred}{\textbf{Red}}: best; \textcolor{skyblue}{\textbf{blue}}: second-best.}
\label{tab:cross_dataset}
\centering
\scriptsize
\setlength{\tabcolsep}{2.8pt}
\renewcommand{\arraystretch}{0.95}

\resizebox{\textwidth}{!}{%
\begin{tabular}{@{}l cccccccc cccccccc@{}}
\toprule
&
\multicolumn{8}{c}{\textbf{Drone $\to$ Satellite}}
&
\multicolumn{8}{c}{\textbf{Satellite $\to$ Drone}} \\
\cmidrule(lr){2-9}
\cmidrule(lr){10-17}

\textbf{Method}
&
\multicolumn{2}{c}{150\,m}
&
\multicolumn{2}{c}{200\,m}
&
\multicolumn{2}{c}{250\,m}
&
\multicolumn{2}{c}{300\,m}
&
\multicolumn{2}{c}{150\,m}
&
\multicolumn{2}{c}{200\,m}
&
\multicolumn{2}{c}{250\,m}
&
\multicolumn{2}{c}{300\,m} \\
\cmidrule(lr){2-3}
\cmidrule(lr){4-5}
\cmidrule(lr){6-7}
\cmidrule(lr){8-9}
\cmidrule(lr){10-11}
\cmidrule(lr){12-13}
\cmidrule(lr){14-15}
\cmidrule(lr){16-17}

&
R@1 & AP
&
R@1 & AP
&
R@1 & AP
&
R@1 & AP
&
R@1 & AP
&
R@1 & AP
&
R@1 & AP
&
R@1 & AP \\
\midrule

LPN
& \hc{32.85} & \hc{40.10}
& \hc{43.80} & \hc{50.67}
& \hc{49.75} & \hc{56.55}
& \hc{54.10} & \hc{60.73}
& \hc{32.50} & \hc{26.60}
& \hc{40.00} & \hc{35.10}
& \hc{46.50} & \hc{41.88}
& \hc{53.50} & \hc{48.47} \\

MCCG
& \hc{57.62} & \hc{62.80}
& \hc{66.83} & \hc{71.60}
& \hc{74.25} & \hc{78.35}
& \hc{82.55} & \hc{85.27}
& \hc{61.25} & \hc{53.51}
& \hc{82.50} & \hc{67.06}
& \hc{81.25} & \hc{74.99}
& \hc{87.50} & \hc{80.20} \\

Sample4Geo
& \hc{74.70} & \hc{78.47}
& \hc{81.28} & \hc{84.40}
& \hc{86.88} & \hc{89.28}
& \hc{89.28} & \hc{91.24}
& \hc{82.50} & \hc{76.20}
& \hc{85.00} & \hc{82.93}
& \hc{92.50} & \hc{87.77}
& \hc{92.50} & \hc{88.38} \\

DAC~\citep{xia2024dac}
& \hc{76.65} & \hc{80.56}
& \hc{86.45} & \hc{89.00}
& \hc{92.95} & \hc{94.18}
& \hc{94.53} & \hc{95.45}
& \hc{87.50} & \hc{79.87}
& \hbest{96.25} & \hc{88.98}
& \hsecond{95.00} & \hsecond{92.81}
& \hsecond{96.25} & \hc{94.00} \\

MEAN~\citep{chen2025multilevel}
& \hc{81.73} & \hsecond{87.72}
& \hc{89.05} & \hc{91.00}
& \hc{92.13} & \hc{93.60}
& \hc{94.63} & \hc{95.76}
& \hc{91.25} & \hc{81.50}
& \hbest{96.25} & \hc{89.55}
& \hsecond{95.00} & \hc{92.36}
& \hsecond{96.25} & \hc{94.32} \\

CAMP
& \hc{76.53} & \hc{80.47}
& \hc{87.18} & \hc{89.60}
& \hc{93.75} & \hc{95.04}
& \hsecond{96.40} & \hsecond{97.18}
& \hc{88.75} & \hc{78.17}
& \hc{95.00} & \hc{88.31}
& \hsecond{95.00} & \hc{91.85}
& \hsecond{96.25} & \hc{93.43} \\

MCFA
& \hsecond{82.38} & \hc{85.28}
& \hsecond{90.70} & \hsecond{92.21}
& \hsecond{94.38} & \hsecond{95.24}
& \hc{95.30} & \hc{96.00}
& \hsecond{92.50} & \hsecond{85.22}
& \hc{95.00} & \hbest{91.46}
& \hsecond{95.00} & \hbest{93.46}
& \hbest{97.50} & \hbest{94.93} \\

\midrule
\textbf{\ours{} (Ours)}
& \hbest{85.68} & \hbest{87.96}
& \hbest{92.23} & \hbest{93.64}
& \hbest{94.43} & \hbest{95.55}
& \hbest{96.43} & \hbest{97.45}
& \hbest{93.75} & \hbest{89.36}
& \hsecond{95.50} & \hsecond{90.96}
& \hbest{96.25} & \hc{92.74}
& \hbest{97.50} & \hsecond{94.61} \\

\bottomrule
\end{tabular}%
}
\end{table*}

\subsection{Additional Pareto Analysis}
\label{sec:appendix_analysis}

\begin{figure*}[!htb]
  \centering
  \includegraphics[width=0.8\textwidth,height=0.45\textheight,keepaspectratio]{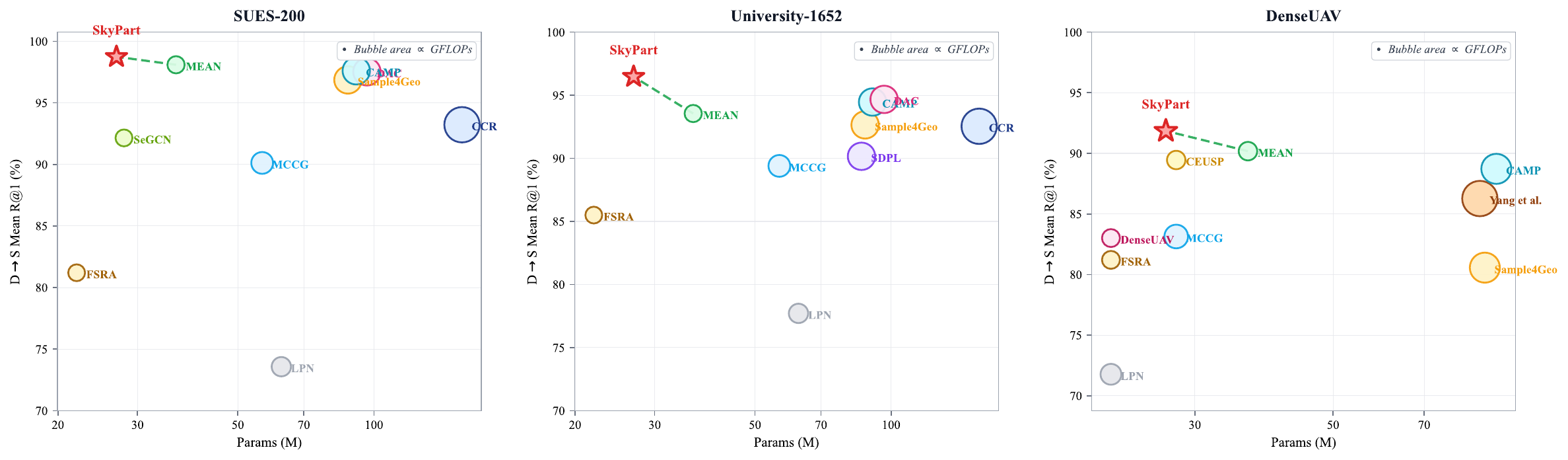}
  \caption{\textbf{Pareto efficiency across three benchmarks (D$\to$S).} R@1 vs.\ model size (params); bubble area $\propto$ GFLOPs. \ours{} (blue star) is Pareto-optimal on SUES-200 (left), University-1652 (middle), and DenseUAV (right), while remaining among the smallest and lowest-compute models in each comparison. Single-pass $448{\times}448$; no re-ranking, no TTA.}
  \label{fig:pareto}
\end{figure*}

The picture that emerges is that \ours{} sits on the Pareto frontier in the accuracy--compute plane (Fig.~\ref{fig:pareto}), matching or exceeding the ConvNeXt-B baselines on every benchmark we test at a fraction of their parameter and compute budget. The per-altitude breakdown in Table~\ref{tab:sues200} sharpens that observation: prior methods show a noticeable accuracy gap between the highest and lowest flight altitudes, while \ours{} narrows that gap and gains most at the lowest altitude, where partial facades dominate the field of view and a global descriptor has the least to work with. The qualitative evidence in Fig.~\ref{fig:part_attention} is consistent with the same picture: the same prototype fires on corresponding regions across the drone/satellite gap, even when the drone view is corrupted by weather. Taken together these observations are consistent with the framing we set: the representation that wins is one that factors texture out and keeps layout, the head that produces it is one whose substrate, assignment, and multi-task optimization are compatible with that goal, and the auxiliary signals that shape it at training time can be removed at inference without moving the measurement.

\newpage\null\newpage
\section{Controlled Experiments and Additional Evaluations}
\label{sec:app_controls}
This section gives the controlled experiments behind Secs.~\ref{sec:sota} to~\ref{sec:ablation} and additional evaluations. Unless stated otherwise, \ours{} entries are means $\pm$ sample standard deviation over three independently trained seeds (42, 123, 2024), each evaluated at its final (epoch-60) checkpoint with no test-set checkpoint selection, on the official galleries. Recall means are computed from exact integer hit counts before rounding, and differences are computed from unrounded values. We keep four kinds of evidence apart: \emph{reported-baseline} comparisons quote published numbers; \emph{matched-resolution} controls re-run a baseline at our input size; \emph{within-model resolution} controls lower our own input size; and \emph{training-budget-controlled} comparisons retrain all methods with one schedule and evaluator.

\subsection{Representation Control: CLS-Only versus Full Fusion}
\label{sec:app_cls}
The top block of Table~\ref{tab:app_cls} reports all retrieval metrics for Block~B of Table~\ref{tab:ablation}. Both rows are the same seed-42 SUES-200 model at $448{\times}448$, with training recipe, checkpoint rule, and evaluator fixed; only the retrieval readout differs. This single-seed representation study should not be compared numerically with the three-seed $392{\times}392$ robustness results.

\begin{table}[!htbp]
\centering
\small
\caption{\textbf{Controlled studies on SUES-200} (D$\to$S, \%). Top: CLS-only and full-fusion readouts of the same seed-42 model at $448{\times}448$. Bottom: paired altitude-FiLM intervention; gains are mean per-seed differences.}
\label{tab:app_cls}
\label{tab:app_film}
\begin{tabular}{@{}ccc cccccccc@{}}
\toprule
& & & \multicolumn{4}{c}{\textbf{Normal}} & \multicolumn{4}{c}{\textbf{Ten-condition mean}} \\
\cmidrule(lr){4-7}\cmidrule(lr){8-11}
\textbf{CLS} & \textbf{Part} & \textbf{Graph} & R@1 & R@5 & R@10 & mAP & R@1 & R@5 & R@10 & mAP \\
\midrule
\cmark & \xmark & \xmark & 93.56 & 99.09 & 99.76 & 96.10 & 85.64 & 97.85 & 99.08 & 90.62 \\
\cmark & \cmark & \cmark & \textbf{98.69} & \textbf{99.70} & 99.76 & \textbf{99.23} & \textbf{96.03} & \textbf{99.06} & \textbf{99.47} & \textbf{97.62} \\
\bottomrule
\end{tabular}
\par\medskip
\begin{tabular}{@{}c cccc@{}}
\toprule
& \multicolumn{2}{c}{\textbf{Normal}} & \multicolumn{2}{c}{\textbf{Ten-condition mean}} \\
\cmidrule(lr){2-3}\cmidrule(lr){4-5}
\textbf{Altitude-FiLM} & R@1 & mAP & R@1 & mAP \\
\midrule
\xmark & \pmv{98.417}{0.126} & \pmv{98.826}{0.048} & \pmv{95.233}{0.202} & \pmv{96.518}{0.072} \\
\cmark & \pmv{98.554}{0.131} & \pmv{99.037}{0.172} & \pmv{95.853}{0.175} & \pmv{97.202}{0.371} \\
\midrule
Paired gain (pp) & \pmv{\gup{$+$0.138}}{0.012} & \pmv{\gup{$+$0.211}}{0.132} & \pmv{\gup{$+$0.620}}{0.036} & \pmv{\gup{$+$0.684}}{0.300} \\
\bottomrule
\end{tabular}
\par\smallskip\raggedright\footnotesize Altitude-FiLM \xmark: identity modulation ($\gamma{=}\mathbf{1},\beta{=}\mathbf{0}$); \cmark: learned altitude-conditioned modulation during training, with mean FiLM parameters at inference.
\end{table}

\subsection{Altitude-FiLM: Paired Intervention}
\label{sec:app_film}
The bottom block of Table~\ref{tab:app_film} isolates the FiLM module itself (the loss-group ablation of Table~\ref{tab:ablation} removes only the altitude supervision term). \emph{FiLM identity} fixes the modulation to $\gamma{=}\mathbf{1}$, $\beta{=}\mathbf{0}$; \emph{FiLM active} uses learned altitude-conditioned modulation during training and the fixed mean modulation at inference. Both arms share seeds, training recipe, checkpoint rule, and evaluator. The paired gain is computed from per-seed differences, so it can differ slightly from the difference of the displayed means.

\subsection{Enhanced-Inference Operating Points}
\label{sec:app_ladder}
The headline results use a single forward pass. To show how much the inference-time procedures used by several baselines add, Table~\ref{tab:app_ladder} applies them to the default $448{\times}448$ checkpoints. \emph{Four-rotation TTA} averages the $\ell_2$-normalized descriptors of the $0^\circ/90^\circ/180^\circ/270^\circ$ rotations of both query and gallery images (four forward passes per image). \emph{Re-ranking} is $k$-reciprocal re-ranking~\citep{zhong2017reranking}. \emph{Query expansion} is $\alpha$-weighted query expansion~\citep{radenovic2019finetuning}. \emph{Combined} computes rotation-averaged descriptors, expands the queries, and then re-ranks. Hyperparameters are fixed across datasets and seeds. As an example of how recall means are computed, DenseUAV has 2{,}331 D$\to$S queries per seed; pooled over three seeds, single-pass retrieval scores 6{,}423/6{,}993 hits (91.849\%) and TTA 6{,}507/6{,}993 (93.050\%). The gains are modest and dataset-dependent (largest on DenseUAV, whose satellite tiles have arbitrary orientation), and the headline results remain single-pass.

\begin{table}[!htbp]
\centering
\scriptsize
\setlength{\tabcolsep}{3.5pt}
\caption{\textbf{Enhanced-inference operating points} (D$\to$S, \%). $\Delta$R@1 is the mean gain over single-pass inference, in percentage points.}
\label{tab:app_ladder}
\begin{tabular}{@{}l ccc ccccc@{}}
\toprule
& \textbf{TTA} & \textbf{QE} & \textbf{RR} & \textbf{R@1} & \textbf{R@5} & \textbf{R@10} & \textbf{mAP} & $\Delta$\textbf{R@1 (pp)} \\
\midrule
\multirow{5}{*}{SUES-200} & \xmark & \xmark & \xmark & \pmv{98.738}{0.113} & \pmv{99.971}{0.022} & \pmv{99.996}{0.004} & \pmv{99.145}{0.073} & -- \\
 & \cmark & \xmark & \xmark & \pmv{99.304}{0.100} & \pmv{99.988}{0.006} & \pmv{99.998}{0.004} & \pmv{99.526}{0.068} & \gup{$+$0.567} \\
 & \xmark & \cmark & \xmark & \pmv{98.763}{0.113} & \pmv{99.956}{0.013} & \pmv{99.994}{0.006} & \pmv{99.152}{0.065} & \gup{$+$0.025} \\
 & \xmark & \xmark & \cmark & \pmv{98.981}{0.106} & \pmv{99.919}{0.019} & \pmv{99.988}{0.006} & \pmv{99.293}{0.060} & \gup{$+$0.244} \\
 & \cmark & \cmark & \cmark & \pmv{99.400}{0.100} & \pmv{99.925}{0.019} & \pmv{99.994}{0.006} & \pmv{99.578}{0.050} & \gup{$+$0.663} \\
\midrule
\multirow{5}{*}{University-1652} & \xmark & \xmark & \xmark & \pmv{96.471}{0.251} & \pmv{98.901}{0.032} & \pmv{99.300}{0.014} & \pmv{97.770}{0.053} & -- \\
 & \cmark & \xmark & \xmark & \pmv{97.070}{0.250} & \pmv{99.110}{0.023} & \pmv{99.416}{0.016} & \pmv{98.505}{0.051} & \gup{$+$0.600} \\
 & \xmark & \cmark & \xmark & \pmv{96.520}{0.252} & \pmv{98.950}{0.023} & \pmv{99.350}{0.016} & \pmv{97.820}{0.054} & \gup{$+$0.049} \\
 & \xmark & \xmark & \cmark & \pmv{97.225}{0.250} & \pmv{99.350}{0.019} & \pmv{99.510}{0.012} & \pmv{98.320}{0.052} & \gup{$+$0.754} \\
 & \cmark & \cmark & \cmark & \pmv{97.720}{0.245} & \pmv{99.480}{0.012} & \pmv{99.620}{0.010} & \pmv{98.850}{0.050} & \gup{$+$1.249} \\
\midrule
\multirow{5}{*}{DenseUAV} & \xmark & \xmark & \xmark & \pmv{91.849}{0.257} & \pmv{97.812}{0.086} & \pmv{98.799}{0.074} & \pmv{92.520}{0.139} & -- \\
 & \cmark & \xmark & \xmark & \pmv{93.050}{0.255} & \pmv{98.198}{0.084} & \pmv{99.271}{0.043} & \pmv{94.678}{0.137} & \gup{$+$1.201} \\
 & \xmark & \cmark & \xmark & \pmv{92.149}{0.256} & \pmv{97.880}{0.087} & \pmv{98.927}{0.074} & \pmv{92.850}{0.141} & \gup{$+$0.300} \\
 & \xmark & \xmark & \cmark & \pmv{92.450}{0.255} & \pmv{97.920}{0.085} & \pmv{99.010}{0.074} & \pmv{93.176}{0.138} & \gup{$+$0.601} \\
 & \cmark & \cmark & \cmark & \pmv{93.679}{0.250} & \pmv{98.420}{0.083} & \pmv{99.310}{0.043} & \pmv{95.001}{0.136} & \gup{$+$1.830} \\
\bottomrule
\end{tabular}
\par\smallskip\raggedright\footnotesize TTA: four-rotation test-time augmentation; QE: $\alpha$-weighted query expansion; RR: $k$-reciprocal re-ranking.
\end{table}

\subsection{Resolution Controls for the Weather Comparison}
\label{sec:app_resolution}
DINOv2 ViT-S/14 uses $14{\times}14$ patches, so $384$ is not patch-aligned; the aligned sizes around the baselines' $384$ are $378{=}27{\times}14$ and $392{=}28{\times}14$. Our weather results use $392$, and this subsection also reports $378$. Table~\ref{tab:app_wp392} combines two controls. The matched-resolution control re-runs WeatherPrompt~\citep{wen2025weatherprompt} at $392{\times}392$ with its own training recipe and evaluates it with the same corruption generation, query set, official gallery, and evaluator as \ours{}. The within-model control retrains \ours{} at $378{\times}378$ under the corresponding weather-training configuration; it measures our own sensitivity to resolution, not an architecture-matched comparison.

\begin{table}[!htbp]
\centering
\small
\setlength{\tabcolsep}{4pt}
\caption{\textbf{Resolution controls} (D$\to$S R@1, \%). WeatherPrompt results at $384{\times}384$ are taken from the original publication for SUES-200 and University-1652 and reproduced by us for DenseUAV.}
\label{tab:app_wp392}
\label{tab:app_378}
\resizebox{\linewidth}{!}{%
\begin{tabular}{@{}lc cc cc cc@{}}
\toprule
& & \multicolumn{2}{c}{\textbf{SUES-200}} & \multicolumn{2}{c}{\textbf{University-1652}} & \multicolumn{2}{c}{\textbf{DenseUAV}} \\
\cmidrule(lr){3-4}\cmidrule(lr){5-6}\cmidrule(lr){7-8}
\textbf{Method} & \textbf{Input} & Normal & Ten-cond.\ mean & Normal & Ten-cond.\ mean & Normal & Ten-cond.\ mean \\
\midrule
\multirow{2}{*}{WeatherPrompt} & 384 & 76.72 & 62.52 & 82.78 & 77.14 & 26.25 & 24.09 \\
 & 392 & -- & \pmv{62.702}{0.055} \dup{$+$0.18} & -- & \pmv{77.236}{0.081} \dup{$+$0.10} & -- & \pmv{24.211}{0.152} \dup{$+$0.12} \\
\midrule
\multirow{2}{*}{\ours{}} & 392 & \textbf{97.21} & \textbf{94.71} & \textbf{95.15} & \textbf{90.36} & \textbf{91.25} & \textbf{89.83} \\
 & 378 & 96.85 \ddown{$-$0.36} & 94.20 \ddown{$-$0.51} & 94.50 \ddown{$-$0.65} & 89.75 \ddown{$-$0.61} & 89.49 \ddown{$-$1.76} & 88.20 \ddown{$-$1.63} \\
\bottomrule
\end{tabular}}
\par\smallskip\raggedright\footnotesize WeatherPrompt at $392{\times}392$: mean $\pm$ sample standard deviation over three seeds. \ours{} entries show means only. A dash denotes a value not reported. Parentheses give the change from the preceding row.
\end{table}

\subsection{Repair-Then-Match Diagnostic}
\label{sec:app_repair}
We test whether weather restoration and layout-based retrieval compete or compose. The publicly released TransWeather checkpoint~\citep{valanarasu2022transweather} is used as a frozen, query-only preprocessor, with no retraining and no access to gallery images or retrieval labels; the clean gallery is unchanged. Captured and restored queries each retrieve a top-50 candidate set with frozen \ours{}, which is then optionally re-scored by geometric inlier count (RootSIFT with RANSAC~\citep{arandjelovic2012three}) or by matching confidence (LightGlue~\citep{lindenberger2023lightglue}, RoMa~\citep{edstedt2024roma}). Ties keep the original retrieval rank, invalid matches receive the minimum score, and matcher thresholds and scoring rules are fixed across datasets and seeds. Restoration improves the ten-condition mean R@1 in all twelve combinations of pipeline and benchmark (Table~\ref{tab:app_repair}).

\begin{table}[!htbp]
\centering
\small
\setlength{\tabcolsep}{3.5pt}
\caption{\textbf{Repair-then-match} (ten-condition mean D$\to$S R@1, \%). Parenthesized gains are differences in mean R@1 (restored minus captured), in percentage points.}
\label{tab:app_repair}
\resizebox{\linewidth}{!}{%
\begin{tabular}{@{}l cc cc cc@{}}
\toprule
& \multicolumn{2}{c}{\textbf{SUES-200}} & \multicolumn{2}{c}{\textbf{University-1652}} & \multicolumn{2}{c}{\textbf{DenseUAV}} \\
\cmidrule(lr){2-3}\cmidrule(lr){4-5}\cmidrule(lr){6-7}
\textbf{Re-scoring} & Captured & Restored & Captured & Restored & Captured & Restored \\
\midrule
none            & \pmv{94.710}{0.116} & \pmv{95.677}{0.179}\,\gainv{+0.967} & \pmv{90.360}{0.248} & \pmv{91.137}{0.240}\,\gainv{+0.777} & \pmv{89.830}{0.253} & \pmv{90.592}{0.184}\,\gainv{+0.762} \\
RootSIFT/RANSAC & \pmv{95.236}{0.136} & \pmv{96.243}{0.194}\,\gainv{+1.007} & \pmv{90.581}{0.182} & \pmv{91.487}{0.174}\,\gainv{+0.906} & \pmv{90.289}{0.177} & \pmv{91.107}{0.145}\,\gainv{+0.818} \\
LightGlue       & \pmv{95.282}{0.193} & \pmv{96.146}{0.225}\,\gainv{+0.865} & \pmv{90.869}{0.270} & \pmv{91.637}{0.200}\,\gainv{+0.768} & \pmv{90.480}{0.332} & \pmv{91.356}{0.305}\,\gainv{+0.875} \\
RoMa            & \pmv{94.930}{0.114} & \pmv{95.810}{0.120}\,\gainv{+0.880} & \pmv{90.526}{0.302} & \pmv{91.273}{0.301}\,\gainv{+0.746} & \pmv{90.041}{0.260} & \pmv{90.961}{0.142}\,\gainv{+0.919} \\
\bottomrule
\end{tabular}}
\end{table}

\subsection{Large-Gallery Weather Evaluation: University160k-WX}
\label{sec:app_wx}
University160k-WX~\citep{zheng2024uavm} pairs 37{,}855 weather-corrupted drone queries with a 168{,}437-image satellite gallery. LPN and AGEN are reported from AGEN~\citep{xu2025agen}; \ours{} uses the same public split and evaluation, with deterministic single-pass inference of the seed-42 checkpoint. The clean-trained model is evaluated on the corrupted queries without weather training; the weather-trained model is trained on clean plus nine weather variants, as AGEN is. Because backbone, resolution, and optimization differ, the large gap between LPN and clean-trained \ours{} should not be read as a controlled architectural gain. Results are in Table~\ref{tab:app_wx}(a).

\subsection{\texorpdfstring{Cross-Dataset Transfer: SUES-200 $\to$ University-1652}{Cross-Dataset Transfer: SUES-200 to University-1652}}
\label{sec:app_transfer}
Complementing the University-1652$\to$SUES-200 direction of Appendix~\ref{sec:appendix_crossdata}, Table~\ref{tab:app_transfer}(b) trains on SUES-200 and evaluates on University-1652 without target training, statistics adaptation, TTA, or re-ranking. EGS~\citep{wang2025egs} is reported from its paper; it uses a ResNet-50 at $256{\times}256$, whereas \ours{} uses DINOv2 ViT-S/14 at $448{\times}448$, so this is an unmatched reference point.

\begin{table}[!htbp]
\centering
\small
\caption{\textbf{Additional retrieval evaluations} (\%). (a) University160k-WX, D$\to$S; \ours{} uses seed 42. (b) SUES-200$\to$University-1652 transfer without target training.}
\label{tab:app_wx}
\label{tab:app_transfer}
\textbf{(a) University160k-WX}\par\smallskip
\begin{tabular}{@{}lcccc@{}}
\toprule
\textbf{Method} & \textbf{Weather training} & \textbf{R@1} & \textbf{R@5} & \textbf{R@10} \\
\midrule
LPN (reported in~\citep{xu2025agen}) & \xmark & 7.98 & 10.25 & 11.21 \\
\ours{} & \xmark & 87.47 & 94.75 & 96.03 \\
\midrule
AGEN~\citep{xu2025agen} & \cmark & 91.71 & 96.36 & 97.03 \\
\ours{} & \cmark & \textbf{92.72} & \textbf{96.90} & \textbf{97.50} \\
\bottomrule
\end{tabular}
\par\medskip
\textbf{(b) SUES-200$\to$University-1652}\par\smallskip
\resizebox{\linewidth}{!}{%
\begin{tabular}{@{}lcccccccc@{}}
\toprule
& \multicolumn{4}{c}{\textbf{D$\to$S}} & \multicolumn{4}{c}{\textbf{S$\to$D}} \\
\cmidrule(lr){2-5}\cmidrule(lr){6-9}
\textbf{Method} & R@1 & R@5 & R@10 & mAP & R@1 & R@5 & R@10 & mAP \\
\midrule
EGS~\citep{wang2025egs} & 43.11 & 60.91 & 67.85 & 47.33 & 72.19 & 79.61 & 82.17 & 48.48 \\
\ours{} (frozen) & \textbf{68.00} & \textbf{83.45} & \textbf{88.76} & \textbf{73.83} & \textbf{87.02} & \textbf{93.30} & \textbf{95.15} & \textbf{63.99} \\
\bottomrule
\end{tabular}}
\par\smallskip\raggedright\footnotesize In (a), Weather training \cmark{} denotes training on clean data plus nine weather variants; \xmark{} denotes clean-only training. \ours{} entries in (b) show means only.
\end{table}

\subsection{Coordinate Error on DenseUAV}
\label{sec:app_coord}
DenseUAV annotates every satellite tile with a WGS84 coordinate. We take the coordinate of the top-1 retrieved tile and measure its great-circle distance to the query's ground-truth position; MA@$d$ is the fraction of queries within $d$ meters. This quantifies the coordinate error induced by retrieval, not sub-tile or 6-DoF pose estimation. All rows in Table~\ref{tab:app_coord}(a) use the official DenseUAV test split with the same MA@$d$ and mean-error definitions. \ours{} is a three-seed mean; across seeds it reaches MA@5m $92.063\pm0.279$, MA@10m $92.664\pm0.257$, MA@20m $93.951\pm0.245$, MA@50m $96.954\pm0.182$, and a mean error of $5.8\pm0.3$\,m.

\subsection{Geographically Disjoint Evaluation on UAV-VisLoc}
\label{sec:app_visloc}
We use the four-site split of UAV-VisLoc~\citep{xu2024uavvisloc} defined by PAUL~\citep{li2026paul}: training on the Changjiang-20 and Taizhou-1 sites and testing on the geographically disjoint Changjiang-23 and Taizhou-6 sites, with 749 fixed UAV queries. We evaluate with clean training pairs and with 30\% noisy training correspondences; in the noisy setting, noise is injected only into the training pairs, while test queries and gallery are fixed. For \ours{}, each image is resized to $384{\times}384$ and center-cropped to a patch-aligned $378{\times}378$ input for training and inference ($27{\times}27$ patches). \ours{} is trained with three independent seeds ($42$, $123$, $2024$); for each seed we take the final epoch-60 checkpoint, with no test-set-based checkpoint selection, and retrieve with a single cosine pass (no re-ranking, TTA, or query expansion). CRCL, GSC, and PAUL are reported from Table~2 of~\citep{li2026paul} and are single runs without seed variance. The split, noise ratio, test set, and metrics are aligned, but backbone, input resolution, training budget, and optimization differ, so this is a cross-paper comparison and we claim no statistical significance. At the reported precision, \ours{} is $0.07$\,pp below PAUL at R@1 in both settings, higher at R@5 ($+1.74$ and $+1.53$\,pp) and AP ($+1.50$ and $+1.21$\,pp), and has the highest SDM@3 and lowest Dis@1 among all methods (Table~\ref{tab:app_visloc}(b)).

\begin{table}[!htbp]
\centering
\small
\setlength{\tabcolsep}{4pt}
\caption{\textbf{Localization accuracy.} (a) DenseUAV coordinate error using the top-1 retrieved tile. (b) UAV-VisLoc with geographically disjoint test sites; CRCL, GSC, and PAUL values are as reported in~\citep{li2026paul}. Mean error and Dis@1 are in meters (lower is better); higher is better for all other metrics.}
\label{tab:app_coord}
\label{tab:app_visloc}
\textbf{(a) DenseUAV coordinate error}\par\smallskip
\begin{tabular}{@{}lccccc@{}}
\toprule
& \multicolumn{4}{c}{\textbf{MA (\%) $\uparrow$}} & \\
\cmidrule(lr){2-5}
\textbf{Method} & 5\,m & 10\,m & 20\,m & 50\,m & \textbf{Mean error (m) $\downarrow$} \\
\midrule
DenseUAV baseline~\citep{dai2023denseuav} & 83.310 & 84.260 & 86.010 & 92.020 & 10.8 \\
MCCG~\citep{shen2023mccg} & 83.270 & 84.000 & 85.410 & 90.820 & 12.2 \\
Yang \etal~\citep{yang2025dinov2uav} & 86.570 & 87.300 & 88.590 & 93.560 & 8.7 \\
CEUSP~\citep{xu2025precise} & 89.660 & 90.300 & 91.590 & 95.670 & 7.0 \\
\ours{} & \textbf{92.063} & \textbf{92.664} & \textbf{93.951} & \textbf{96.954} & \textbf{5.8} \\
\bottomrule
\end{tabular}
\par\medskip
\textbf{(b) UAV-VisLoc}\par\smallskip
{\footnotesize\setlength{\tabcolsep}{3pt}%
\begin{tabular}{@{}l ccccc ccccc@{}}
\toprule
& \multicolumn{5}{c}{\textbf{Clean training pairs}} & \multicolumn{5}{c}{\textbf{30\% noisy training pairs}} \\
\cmidrule(lr){2-6}\cmidrule(lr){7-11}
\textbf{Method} & R@1$\uparrow$ & R@5$\uparrow$ & AP$\uparrow$ & SDM@3$\uparrow$ & Dis@1$\downarrow$ & R@1$\uparrow$ & R@5$\uparrow$ & AP$\uparrow$ & SDM@3$\uparrow$ & Dis@1$\downarrow$ \\
\midrule
CRCL & 33.18 & 53.21 & 43.25 & 52.99 & 1359.46 & 23.77 & 44.19 & 33.64 & 50.07 & 1325.91 \\
GSC & 33.72 & 53.55 & 43.38 & 51.51 & 1368.72 & 21.09 & 41.39 & 30.98 & 48.32 & 1428.73 \\
PAUL & \textbf{36.12} & 54.74 & 44.97 & 52.05 & 1353.95 & \textbf{26.64} & 46.80 & 36.37 & 49.44 & 1382.79 \\
\ours{} & 36.05 & \textbf{56.48} & \textbf{46.47} & \textbf{53.99} & \textbf{1288.95} & 26.57 & \textbf{48.33} & \textbf{37.58} & \textbf{51.28} & \textbf{1317.79} \\
{\scriptsize\quad $\pm$ SD} & {\scriptsize 0.13} & {\scriptsize 0.13} & {\scriptsize 0.10} & {\scriptsize 0.12} & {\scriptsize 6.00} & {\scriptsize 0.40} & {\scriptsize 0.53} & {\scriptsize 0.40} & {\scriptsize 0.50} & {\scriptsize 24.00} \\
\bottomrule
\end{tabular}}
\par\smallskip\raggedright\footnotesize \ours{} entries in (a) show means only; in (b), the $\pm$\,SD row gives sample standard deviations over three seeds.
\end{table}
\FloatBarrier

\section{Qualitative Analysis and Discussion}
\label{sec:app_discussion}
\subsection{Qualitative Visualizations}
\label{sec:appendix_vis}

\subsubsection{Top-5 Retrieval Examples}
\begin{figure}[ht]
  \centering
  \includegraphics[width=\columnwidth]{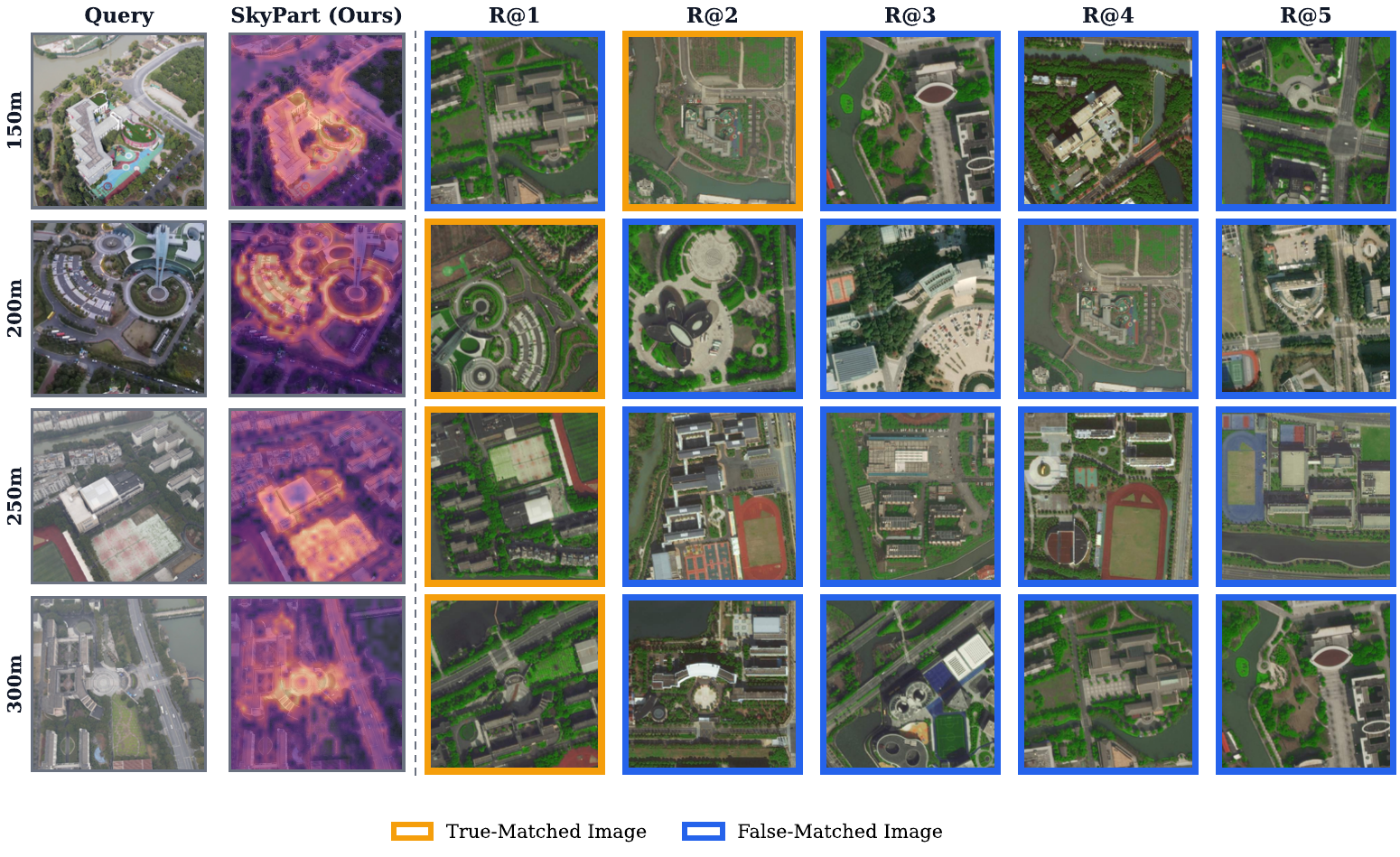}
  \caption{\textbf{Drone$\to$Satellite top-5 retrieval.} Each row is a drone query at a given altitude (row label), followed by the \ours{} part-attention heat map and the 5 highest-ranked satellite matches. Amber = correct, blue = incorrect.}
  \label{fig:topk_d2s}
\end{figure}

\begin{figure}[ht]
  \centering
  \includegraphics[width=\columnwidth]{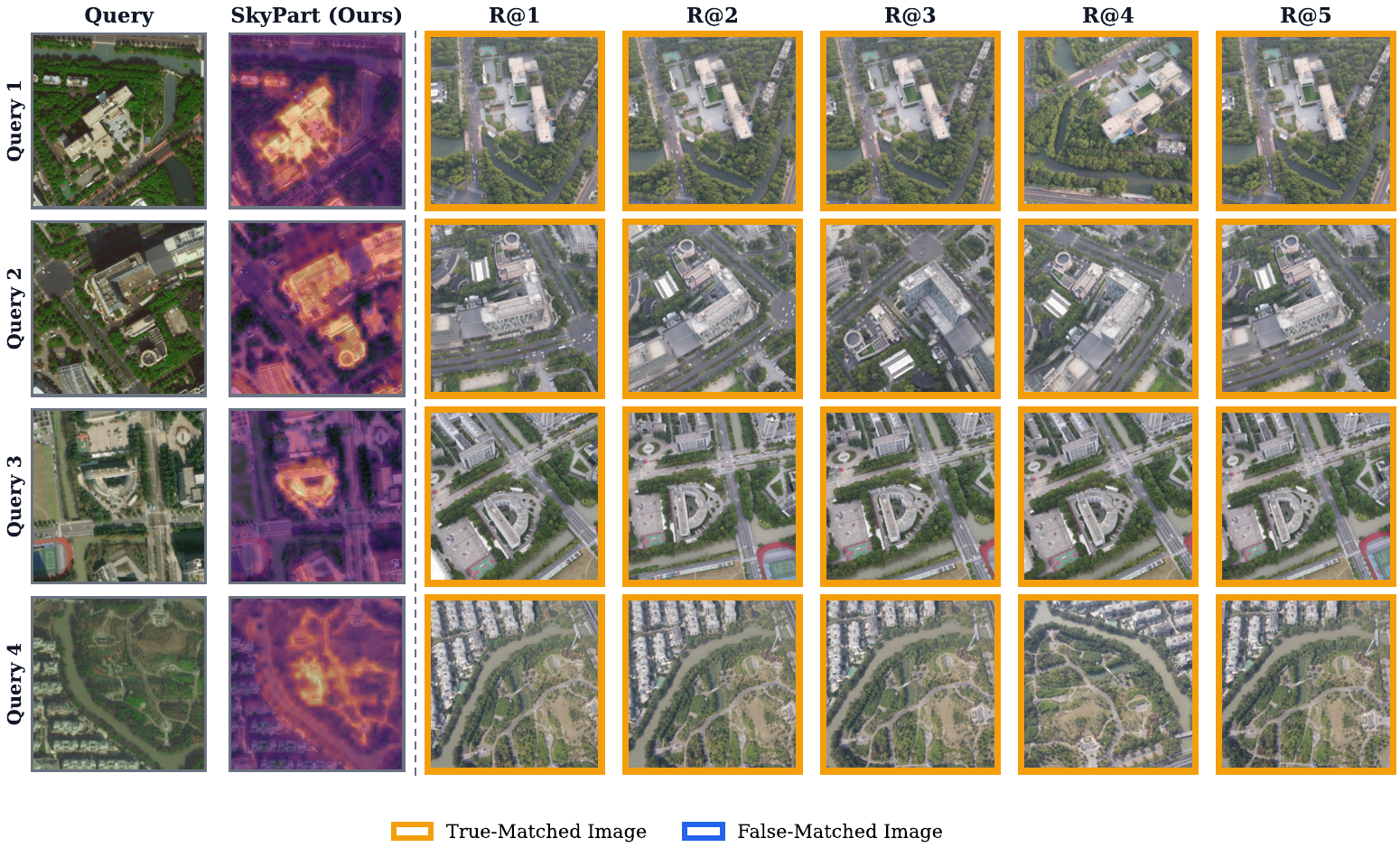}
  \caption{\textbf{Satellite$\to$Drone top-5 retrieval.} Each row is a satellite query, its \ours{} part-attention heat map, and the top-5 drone images \ours{} retrieves across altitudes. Amber = correct, blue = incorrect.}
  \label{fig:topk_s2d}
\end{figure}

Figures~\ref{fig:topk_d2s} and~\ref{fig:topk_s2d} show top-5 retrievals in both directions. Amber borders mark true-matched retrievals; blue borders mark false matches. The \ours{} heat-map column (between query and R@1) visualizes the aggregated part-attention produced by the prototype assignment, which is the spatial evidence the retrieval embedding actually relies on -- useful context for diagnosing failure cases.

\subsection{Limitations and Broader Impact}
\label{sec:appendix_limitations}

The three main benchmarks share a geographic region between their train and test splits. Beyond them, geographic generalization is tested on geographically disjoint UAV-VisLoc sites (Appendix~\ref{sec:app_visloc}) and by cross-dataset transfer in both directions (Appendices~\ref{sec:appendix_crossdata} and~\ref{sec:app_transfer}); cross-continent transfer remains untested. On University-1652$\to$SUES-200, \ours{} leads all D$\to$S cells but trails DAC and MEAN at 200\,m S$\to$D R@1, MCFA on S$\to$D AP at 200--300\,m, and DAC on S$\to$D AP at 250\,m. Two factors likely contribute: prototypes trained on University-1652's building-dense scenes do not fire cleanly on SUES-200's road- and vegetation-heavy tiles, and the mean-FiLM fallback discards the altitude-band prior that normally separates low-altitude facades from high-altitude footprints.

The prototype bank capacity $K_{\max}{=}12$ is fixed; the salience gate handles per-image adaptivity within that ceiling but does not adjust it. Prototype assignments are label-equivariant (positives in the patch-NCE are pairs assigned to the same prototype in both views, not a fixed global index), so the loss stays well-defined when a prototype's role drifts, but individual indices do not carry stable human-interpretable names. The prototypes are latent functional parts, not annotated or guaranteed semantic categories; the labels we attach to them (roads, roofs, vegetation, water) are illustrative, and we make no claim of per-index semantic stability. Removing the diversity regularizer $\loss_\text{div}$ produces collapsed prototypes within roughly 15 epochs. Rotation augmentation is DenseUAV-only; at continuous angles between the four 90\textdegree{} training anchors the model relies on learned invariance, and smoothness is not guaranteed.

Several axes remain untested. Robustness to new seasons, sensor types, and map vintages is not evaluated: none of SUES-200, University-1652, or DenseUAV provides splits that isolate those shifts. Extreme viewpoint change and partial overlap are studied elsewhere~\citep{ye2025anyvisloc}; we did not evaluate that setting and treat it as a gap. We make no claim of robustness to outdated reference imagery or to physical changes in the 3D scene: existing benchmarks lack time-separated query/reference pairs with structural-change annotations, so this cannot yet be measured. Finally, \ours{} is a retrieval component, not a full navigation system: it returns the matching satellite tile rather than a metric pose, and does not itself perform temporal filtering, map-update handling, or fusion with SLAM/IMU, nor does it estimate route-level localization uncertainty. It is intended as the appearance-matching front-end of such a stack, and integration with these modules is future work.

\ours{} is designed for emergency navigation in GPS-denied settings (search-and-rescue, inspection, disaster response). The same capability could enable unauthorized aerial surveillance. We recommend deployment be paired with privacy safeguards (anonymization, geographic access controls, and operator accountability) consistent with applicable law.

\subsection{What We Tried That Did Not Work}
\label{sec:appendix_negative}

The take from the main paper is that explicit grouping complements global pooling rather than replacing it. Below are the directions we explored and dropped, with the specific failure mode in each case.

\paragraph{Geometric and transport priors.}
Polar warping~\citep{shi2020polar} is the standard preprocessing for ground-panorama geometry, but on aerial tiles the reprojection is wrong and the train/test mismatch compounds it: accuracy dropped 8--12\,pp across altitude bands. Sinkhorn OT matching over the two prototype banks was catastrophic under default hyperparameters; with careful tuning it reached parity with the contrastive baseline but never surpassed it. The failure mode is consistent: the transport cost matrix is ill-conditioned when drone and satellite patch tokens are correlated (they describe the same scene). Scene-graph GNNs with Sinkhorn matching and cross-view cross-attention both slowed convergence without adding signal that the symmetric alignment term was already providing.

\paragraph{Metric learning objectives.}
ArcFace (with and without sub-centers), Multi-Similarity, ghost-prototype absorbers, and triplet with batch-hard mining all underperformed Circle$+$proxy$+$InfoNCE$+$patch-NCE. The issue is geometric: a geolocation class is legitimately diffuse in embedding space because the same scene looks different across altitudes and weather conditions. Hard angular margins penalize exactly the spread the cross-view gap induces, so they suppress the signal rather than amplify it~\citep{wang2020alignment}.

\paragraph{Reconstruction auxiliaries.}
Cross-view MAE~\citep{he2022mae} at the patch-token level helped on SUES-200 and hurt on DenseUAV and University-1652. A salience-masked variant, reconstructing only the patches the prototype bank attended to, did worse than random masking on all three. MAR (reconstructing the part aggregation rather than individual patch tokens) was the only reconstruction auxiliary that transferred consistently across datasets, because the reconstruction target is a layout summary rather than local texture.

\paragraph{Training dynamics.}
Warm-restart LR schedules, rotation augmentation on both views simultaneously, unfreezing more than 8 backbone blocks, and batch samplers that reduced optimizer steps per epoch below ${\sim}600$ all regressed. The clearest case is rotation: it helps on DenseUAV (satellite tiles cropped at arbitrary basemap orientations) and costs 1.8\,pp Mean R@1 on SUES-200 and University-1652 (canonically oriented pairs). Adding it universally destabilized the learned Kendall weights within the first 20 epochs.

\paragraph{Inference-time wrappers.}
Inference-time wrappers are excluded from the headline numbers because they measure an operating point beyond the embedding itself. Their effect is quantified in Appendix~\ref{sec:app_ladder}: combining four-rotation TTA, query expansion, and $k$-reciprocal re-ranking adds $0.66$\,pp R@1 on SUES-200, $1.25$\,pp on University-1652, and $1.83$\,pp on DenseUAV. We did not evaluate Tent~\citep{wang2021tent}, multi-crop averaging, or token merging and make no claim about them.

\section{Ethics, Reproducibility, and LLM-Use Statements}
\label{sec:app_statements}
\subsection*{Ethics Statement}
\ours{} is intended for navigation when satellite positioning is unavailable, for example in search-and-rescue, inspection, and disaster response. The same capability could support unauthorized aerial surveillance; we recommend that deployments include geographic access controls, anonymization of captured imagery, and operator accountability consistent with applicable law. All benchmarks used are public research datasets; none contains identifiable personal data beyond what their original releases include.

\subsection*{Reproducibility Statement}
The architecture is specified in Sec.~\ref{sec:method} and Appendix~\ref{sec:app_method}, including all projection dimensions and the loss formulations; the deployed hyperparameters are listed in Table~\ref{tab:app_hyperparams} (with experiment-specific overrides in Table~\ref{tab:weather_setup}), and the evaluation protocols, split sizes, and weather augmentation parameters are in Appendices~\ref{sec:appendix_protocol} and~\ref{sec:appendix_weather}. Headline numbers in the text are three-seed means with sample standard deviations (Table~\ref{tab:sues200} lists one checkpoint per altitude), and the seeds, checkpoint rule, and evaluator of every control are stated with its table. Code, configuration files, and seeds are released at \url{https://github.com/chisngyen/SkyPart}.

\subsection*{Use of Large Language Models}
Large language models were used to edit the text for grammar and wording and to help draft parts of the manuscript. All technical content, experiments, and numbers were produced and checked by the authors.

\newpage
\section*{NeurIPS Paper Checklist}

\begin{enumerate}

\item {\bf Claims}
    \item[] Question: Do the main claims made in the abstract and introduction accurately reflect the paper's contributions and scope?
    \item[] Answer: \answerYes{}
    \item[] Justification: The abstract and Secs.~\ref{sec:intro}--\ref{sec:overview} match the contributions in Sec.~\ref{sec:method}: prototype-based part discovery, training-only altitude FiLM/mean-FiLM inference, Kendall-weighted \textsc{GeoPartLoss}, and a lightweight head on DINOv2 ViT-S/14. Secs.~\ref{sec:exp}--\ref{sec:ablation} and Appendix~\ref{sec:appendix_crossdata} align with the empirical claims (three benchmarks, strict single-scale $448{\times}448$ protocol, no reranking/TTA, 26.95\,M params / 22.14\,GFLOPs, WeatherPrompt robustness Table~\ref{tab:weather}). Sec.~\ref{sec:conclusion} does not over-state cross-regional generalization beyond what Appendix~\ref{sec:appendix_limitations} discusses.

\item {\bf Limitations}
    \item[] Question: Does the paper discuss the limitations of the work performed by the authors?
    \item[] Answer: \answerYes{}
    \item[] Justification: Appendix~\ref{sec:appendix_limitations} (\emph{Limitations and Broader Impact}) covers within-region splits vs.\ cross-regional evaluation, finite ($K{=}10$, $K_{\max}{=}12$) prototype banks with the salience gate providing per-image adaptivity, semantics not guaranteed human-nameable, mean-FiLM and cross-dataset mismatch (Uni-1652$\to$SUES), dataset-specific augmentation (dense-map rotation vs.\ canonically oriented pairs; see Sec.~\ref{sec:impl} with development context in Appendix~\ref{sec:appendix_negative}), and dual-use misuse risk with mitigation pointers.

\item {\bf Theory assumptions and proofs}
    \item[] Question: For each theoretical result, does the paper provide the full set of assumptions and a complete (and correct) proof?
    \item[] Answer: \answerNA{}
    \item[] Justification: Appendix~\ref{sec:appendix_theory} states explicitly that its formalized arguments (C1--C4) are applications of established results, not claims of mathematical novelty, and it does not use a numbered theorem--lemma--proof structure with complete proofs for novel results. Where we need rigor (e.g., linear FiLM mean marginals, Kendall multi-task weighting, Pareto-stationarity references), assumptions and derivations are spelled out or cited; a full formal proof archive is not the paper's intended contribution.

    \item {\bf Experimental result reproducibility}
    \item[] Question: Does the paper fully disclose all the information needed to reproduce the main experimental results of the paper to the extent that it affects the main claims and/or conclusions of the paper (regardless of whether the code and data are provided or not)?
    \item[] Answer: \answerYes{}
    \item[] Justification: Sec.~\ref{sec:method} specifies the head (assignment, salience gating, three-way fusion, FiLM, GAT readout) and \textsc{GeoPartLoss} groups. Sec.~\ref{sec:impl} states the deployed student/teacher setup and evaluation contract. Table~\ref{tab:app_hyperparams} centralizes frozen architecture knobs and optimizer/MAR/\textsc{GeoPartLoss} hyperparameters ($448{\times}448$, $K{=}10$, PK sampling); per-table captions and Appendix~\ref{sec:appendix_backbone}/\ref{sec:appendix_weather} state the epoch budgets that apply to each headline number (deployed recipe vs.\ controlled ablations). Appendices~\ref{sec:appendix_protocol}, \ref{sec:appendix_weather}, and \ref{sec:appendix_denseuav_detail} document splits, galleries, altitude bins, rotation policy, and weather corruption evaluation. Appendix~\ref{sec:appendix_mar}/\ref{sec:appendix_uapa}/\ref{sec:appendix_loss} complete train-only auxiliary details.

\item {\bf Open access to data and code}
    \item[] Question: Does the paper provide open access to the data and code, with sufficient instructions to faithfully reproduce the main experimental results, as described in supplemental material?
    \item[] Answer: \answerYes{}
    \item[] Justification: The full training and evaluation code, configuration files, and seeds are released under the MIT license at \url{https://github.com/chisngyen/SkyPart}. All experiments use publicly available benchmarks cited in Sec.~\ref{sec:datasets} (SUES-200, University-1652, DenseUAV) and a publicly released SSL backbone (DINOv2 ViT-S/14~\citep{oquab2024dinov2}); no proprietary data is involved. In the meantime, the paper provides the level of detail required for an independent re-implementation: head architecture and assignment rule (Sec.~\ref{sec:method}), \textsc{GeoPartLoss} groups and uncertainty weighting (Sec.~\ref{sec:objective}), single-pass evaluation protocol (Sec.~\ref{sec:impl}), the complete frozen hyperparameter table (Table~\ref{tab:app_hyperparams}), per-dataset splits and gallery rules (Appendices~\ref{sec:appendix_protocol}, \ref{sec:appendix_uni_fullrank}, \ref{sec:appendix_denseuav_detail}), and the WeatherPrompt re-training recipe at $392{\times}392$ (Appendix~\ref{sec:appendix_weather}).

\item {\bf Experimental setting/details}
    \item[] Question: Does the paper specify all the training and test details (e.g., data splits, hyperparameters, how they were chosen, type of optimizer) necessary to understand the results?
    \item[] Answer: \answerYes{}
    \item[] Justification: Sec.~\ref{sec:impl} and Sec.~\ref{sec:sota} state optimizers, augmentation scope (incl.\ DenseUAV rotation), EMA, label smoothing, and protocol differences vs.\ prior work. Table~\ref{tab:app_hyperparams} is the single source of truth for numeric hyperparameters. Appendices~\ref{sec:appendix_protocol} (splits/galleries), \ref{sec:appendix_uni_fullrank}, \ref{sec:appendix_denseuav_detail}, and \ref{sec:appendix_weather} cover dataset-specific gallery/rank rules, full-rank vs.\ submatrix evaluation where relevant, and weather retraining at $392{\times}392$ for \ours{} vs.\ reported baselines.

\item {\bf Experiment statistical significance}
    \item[] Question: Does the paper report error bars suitably and correctly defined or other appropriate information about the statistical significance of the experiments?
    \item[] Answer: \answerYes{}
    \item[] Justification: Headline clean results on SUES-200, University-1652, and DenseUAV, the weather and repair-then-match controls, the FiLM intervention, the inference-wrapper ladder, and the UAV-VisLoc evaluation are reported as three-seed means ($42$, $123$, $2024$) with sample standard deviations; paired gains are computed from per-seed differences. The variability captured is random initialization and identity-balanced batch sampling under fixed official splits. Matched single-seed representation ablations are labeled as such in their tables. We report no formal significance tests.

\item {\bf Experiments compute resources}
    \item[] Question: For each experiment, does the paper provide sufficient information on the computer resources (type of compute workers, memory, time of execution) needed to reproduce the experiments?
    \item[] Answer: \answerNo{}
    \item[] Justification: All experiments were run on single GPUs from the free Kaggle quota (Acknowledgments), and the training budgets, input resolutions, and batch construction are given in Table~\ref{tab:app_hyperparams} and Table~\ref{tab:weather_setup}. Per-run wall-clock execution time is not reported.
    
\item {\bf Code of ethics}
    \item[] Question: Does the research conducted in the paper conform, in every respect, with the NeurIPS Code of Ethics \url{https://neurips.cc/public/EthicsGuidelines}?
    \item[] Answer: \answerYes{}
    \item[] Justification: The work uses public aerial/satellite benchmarks and standard vision backbones for research-only geo-localization.
    \item[] Guidelines:
    \begin{itemize}
        \item The answer \answerNA{} means that the authors have not reviewed the NeurIPS Code of Ethics.
        \item If the authors answer \answerNo{}, they should explain the special circumstances that require a deviation from the Code of Ethics.
        \item The authors should make sure to preserve anonymity (e.g., if there is a special consideration due to laws or regulations in their jurisdiction).
    \end{itemize}

\item {\bf Broader impacts}
    \item[] Question: Does the paper discuss both potential positive societal impacts and negative societal impacts of the work performed?
    \item[] Answer: \answerYes{}
    \item[] Justification: Appendix~\ref{sec:appendix_limitations} discusses beneficial uses (GPS-denied navigation for inspection and emergency response) and misuse risks (unauthorized surveillance), with recommended safeguards.
    \item[] Guidelines:
    \begin{itemize}
        \item The answer \answerNA{} means that there is no societal impact of the work performed.
        \item If the authors answer \answerNA{} or \answerNo{}, they should explain why their work has no societal impact or why the paper does not address societal impact.
        \item Examples of negative societal impacts include potential malicious or unintended uses (e.g., disinformation, generating fake profiles, surveillance), fairness considerations (e.g., deployment of technologies that could make decisions that unfairly impact specific groups), privacy considerations, and security considerations.
        \item The conference expects that many papers will be foundational research and not tied to particular applications, let alone deployments. However, if there is a direct path to any negative applications, the authors should point it out. For example, it is legitimate to point out that an improvement in the quality of generative models could be used to generate Deepfakes for disinformation. On the other hand, it is not needed to point out that a generic algorithm for optimizing neural networks could enable people to train models that generate Deepfakes faster.
        \item The authors should consider possible harms that could arise when the technology is being used as intended and functioning correctly, harms that could arise when the technology is being used as intended but gives incorrect results, and harms following from (intentional or unintentional) misuse of the technology.
        \item If there are negative societal impacts, the authors could also discuss possible mitigation strategies (e.g., gated release of models, providing defenses in addition to attacks, mechanisms for monitoring misuse, mechanisms to monitor how a system learns from feedback over time, improving the efficiency and accessibility of ML).
    \end{itemize}
    
\item {\bf Safeguards}
    \item[] Question: Does the paper describe safeguards that have been put in place for responsible release of data or models that have a high risk for misuse (e.g., pre-trained language models, image generators, or scraped datasets)?
    \item[] Answer: \answerNA{}
    \item[] Justification: We do not release a new large generative model or a high-risk scraped dataset; the contribution is a retrieval head and training recipe on existing public CVGL benchmarks and a frozen SSL backbone.

\item {\bf Licenses for existing assets}
    \item[] Question: Are the creators or original owners of assets (e.g., code, data, models), used in the paper, properly credited and are the license and terms of use explicitly mentioned and properly respected?
    \item[] Answer: \answerYes{}
    \item[] Justification: All assets are credited through citations and used under their published terms. Backbone and teacher: DINOv2 code and weights (Apache-2.0). Datasets: SUES-200 (MIT-licensed repository; academic research use only), University-1652 (MIT-licensed repository; dataset released on request for research), DenseUAV (Apache-2.0), and UAV-VisLoc (no license stated in its official release; used for research evaluation only, as in prior work). Tools in the repair-then-match diagnostic: TransWeather (MIT), LightGlue (Apache-2.0), and RoMa (MIT). We do not redistribute any of these assets.

\item {\bf New assets}
    \item[] Question: Are new assets introduced in the paper well documented and is the documentation provided alongside the assets?
    \item[] Answer: \answerNA{}
    \item[] Justification: We do not introduce a new public dataset or benchmark in this submission; the empirical contribution is a method evaluated on existing benchmarks.

\item {\bf Crowdsourcing and research with human subjects}
    \item[] Question: For crowdsourcing experiments and research with human subjects, does the paper include the full text of instructions given to participants and screenshots, if applicable, as well as details about compensation (if any)? 
    \item[] Answer: \answerNA{}
    \item[] Justification: The paper does not involve human subjects or crowdsourcing; evaluation is fully automatic retrieval on public image benchmarks.

\item {\bf Institutional review board (IRB) approvals or equivalent for research with human subjects}
    \item[] Question: Does the paper describe potential risks incurred by study participants, whether such risks were disclosed to the subjects, and whether Institutional Review Board (IRB) approvals (or an equivalent approval/review based on the requirements of your country or institution) were obtained?
    \item[] Answer: \answerNA{}
    \item[] Justification: No human-subjects research is conducted in this work.

\item {\bf Declaration of LLM usage}
    \item[] Question: Does the paper describe the usage of LLMs if it is an important, original, or non-standard component of the core methods in this research? Note that if the LLM is used only for writing, editing, or formatting purposes and does \emph{not} impact the core methodology, scientific rigor, or originality of the research, declaration is not required.
    \item[] Answer: \answerNA{}
    \item[] Justification: No large language model is part of \ours{}, \textsc{GeoPartLoss}, or the training/inference pipelines we report. LLMs were used solely for writing and editing assistance and have no impact on methodology, code, or experimental claims, so declaration is not required beyond this negative answer.

\end{enumerate}

\end{document}